\documentclass[preprint,11.99pt]{elsarticle}

\usepackage{graphicx}
\usepackage{multirow}
\usepackage{amsmath,amssymb,amsfonts}
\usepackage{amsthm}
\usepackage{mathrsfs}
\usepackage[title]{appendix}
\usepackage{textcomp}
\usepackage{manyfoot}
\usepackage{booktabs}
\usepackage{listings}
\usepackage[dvipsnames]{xcolor}
\usepackage{booktabs}   
\usepackage{adjustbox}
\usepackage[table]{xcolor}
\usepackage{longtable}
\usepackage{afterpage}
\usepackage{pifont}
\usepackage{tikz}
\usetikzlibrary{positioning}
\usepackage{enumitem}

\usepackage{algorithm}
\usepackage{algpseudocode}

\newcommand{\cmark}{\ding{51}} 
\newcommand{\xmark}{\ding{55}} 

\def\BibTeX{{\rm B\kern-.05em{\sc i\kern-.025em b}\kern-.08em
    T\kern-.1667em\lower.7ex\hbox{E}\kern-.125emX}}

\theoremstyle{thmstyleone}%
\theoremstyle{thmstyletwo}%

\theoremstyle{thmstylethree}%

\usepackage{graphicx}%
\usepackage{multirow}%
\usepackage{amsmath,amssymb,amsfonts}%
\usepackage{amsthm}%
\usepackage{mathrsfs}%
\usepackage[title]{appendix}%
\usepackage{xcolor}%
\usepackage{textcomp}%
\usepackage{manyfoot}%
\usepackage{booktabs}%
\usepackage{algorithm}%
\usepackage{algorithmicx}%
\usepackage{algpseudocode}%
\usepackage{listings}%

\begin{document}

\begin{frontmatter}



\title{Deep Learning for Anomaly Detection in Railway Systems: A Structured Survey}


\author[aff1,aff3]{Ammar Bouketta}
\author[aff2]{Smail Niar}
\author[aff2,aff4]{Hamza Ouarnoughi}

\affiliation[aff1]{
    organization={Université Polytechnique Hauts-de-France, LAMIH CNRS UMR 8201},
    city={Valenciennes},
    country={France}
}

\affiliation[aff2]{
    organization={Université Polytechnique Hauts-de-France, LAMIH CNRS UMR 8201, INSA Hauts-de-France},
    city={Valenciennes},
    country={France}
}

\affiliation[aff3]{
    organization={Alstom},
    city={Crespin},
    country={France}
}

\affiliation[aff4]{
    organization={Computer Science Department, College of Computing and Informatics, University of Sharjah},
    city={Sharjah},
    country={UAE}
}

\begin{abstract}
Ensuring safe and reliable operation of modern railway systems increasingly relies on data-driven monitoring and intelligent fault detection solutions. In recent years, deep learning has emerged as an effective paradigm for anomaly detection in railway applications, driven by the growing availability of heterogeneous sensor data collected from rolling stock and infrastructure. This paper presents a comprehensive survey of deep learning–based anomaly detection approaches developed for railway systems. The surveyed methods are systematically analyzed and organized using a unified taxonomy that captures the location of anomaly occurrence, data representation and manifestation, sensing modality, and temporal characteristics. Existing approaches, spanning convolutional neural networks, recurrent and attention-based architectures, autoencoders, generative adversarial networks, and transformer-based models, are further structured into major learning paradigms, including classification-based, prediction-based, reconstruction-based, and hybrid strategies. Beyond algorithmic design, this survey discusses key data-centric challenges and reviews commonly adopted evaluation practices and performance metrics. Practical deployment aspects are also addressed, including edge–cloud architectures, computational constraints, and hardware-aware optimization strategies for embedded railway applications. Building on this analysis, the survey introduces a decision-oriented framework that links anomaly characteristics, data properties, and operational constraints to appropriate detection paradigms and deployment configurations. By jointly considering learning models, data characteristics, evaluation protocols, and deployment constraints, this work provides engineers and researchers with a structured and application-oriented reference for selecting and deploying deep learning solutions for railway anomaly detection, while highlighting open challenges and future research directions toward reliable and scalable intelligent monitoring systems.
\end{abstract}



\end{frontmatter}



\section{Introduction}

Railway systems constitute a fundamental component of modern transportation infrastructures, supporting large-scale passenger mobility and freight logistics worldwide. With nearly 930,000 kilometers of railway networks in operation and trillions of passenger- and tonne-kilometers transported annually, railways play a central role in sustainable, energy-efficient, and high-capacity transportation systems~\cite{UIC2025,bhadouria2025advances}. As railway networks continue to expand and aging assets are operated under increasing performance and safety requirements, ensuring reliability, availability, and operational continuity has become a major engineering challenge.

Anomaly detection has emerged as a core function in modern railway monitoring and maintenance systems, as undetected faults can rapidly propagate across interconnected subsystems and lead to service disruptions, costly downtime, or safety-critical incidents. Typical anomalies include track degradation, rolling-stock malfunctions, electrical faults, and abnormal environmental interactions, all of which may impact system performance and safety if not identified at an early stage~\cite{wrobel2024fault}. To address these challenges, the railway industry is progressively adopting intelligent condition monitoring and predictive maintenance strategies based on embedded sensing technologies and artificial intelligence (AI). Such systems are now deployed across a wide range of subsystems, including traction units, passenger access doors, braking systems, HVAC (Heating, Ventilation, and Air Conditioning) units, pantograph-catenary interfaces, and track infrastructure~\cite{cheng2020review}.

\begin{figure}[ht]
    \centering
    \includegraphics[width=\textwidth]{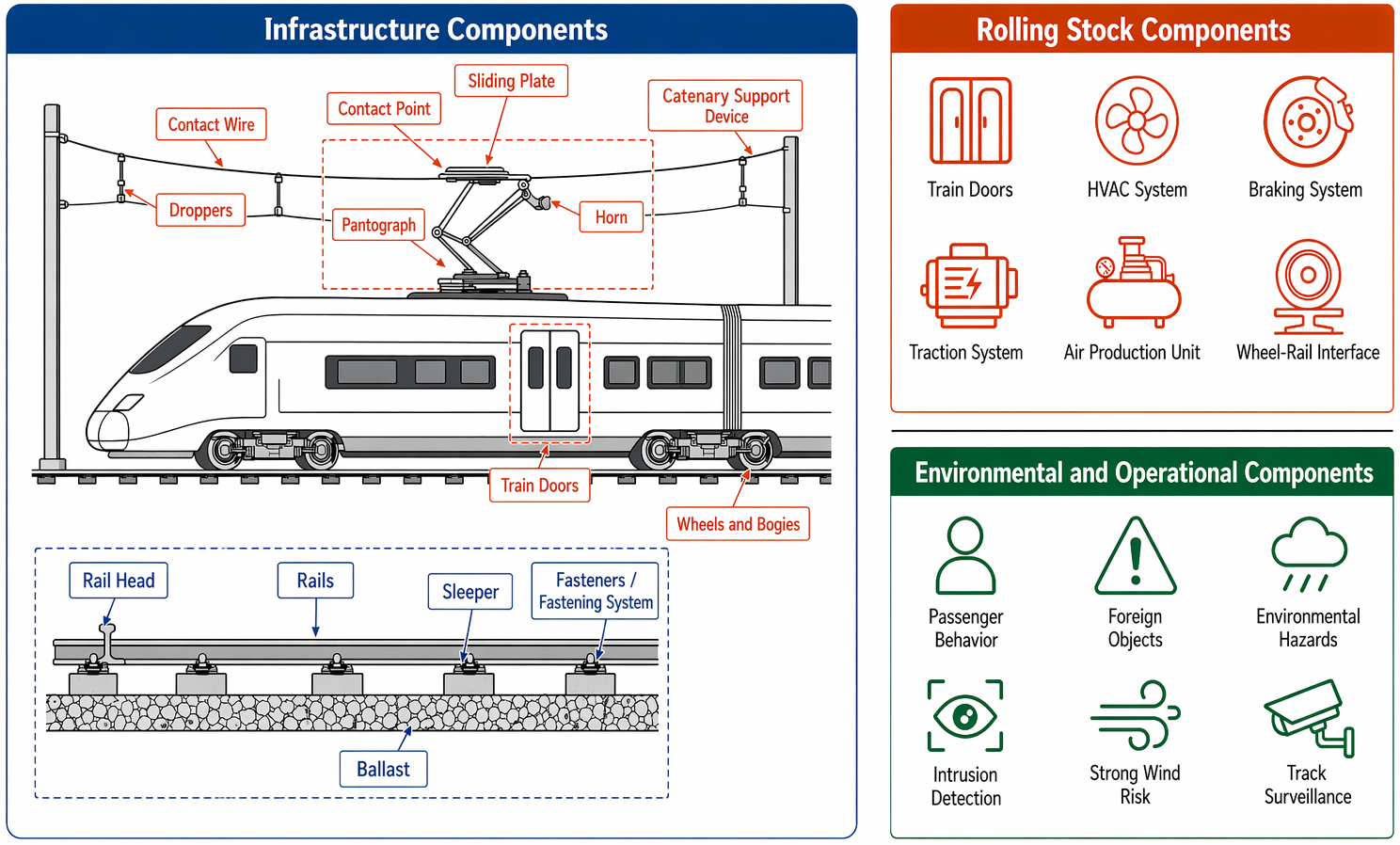}
    \caption{Overview of railway subsystems and 
anomaly detection domains addressed in this 
survey, organized into three levels: 
infrastructure, rolling stock, and 
environmental or behavioral contexts.}
    \label{fig:smart_train}
\end{figure}

As illustrated in Figure~\ref{fig:smart_train}, railway anomaly detection spans multiple operational domains, including onboard rolling-stock subsystems (e.g., doors, HVAC, traction), wayside and infrastructure components (e.g., rails, fasteners, catenary), as well as environmental and behavioral contexts (e.g., foreign objects, wind effects, passenger-related events). These domains differ substantially in sensing modalities, temporal dynamics, fault mechanisms, and data availability, which prevents the use of a single unified detection strategy. Railway monitoring data are often high-dimensional, strongly imbalanced, and temporally structured, posing significant challenges to traditional rule-based or threshold-based detection approaches ~\cite{shim2022anomaly,lourencco2024time}. Such methods typically struggle to generalize across nonlinear operating regimes, multisensor interactions, and evolving environmental conditions, resulting in brittle performance when deployed in real operational settings~\cite{song2024fault}.

Historically, anomaly detection in railway systems has relied on expert-defined rules, statistical thresholds, and handcrafted features derived from classical signal processing techniques, such as spectral analysis or principal component analysis (PCA)~\cite{butakova2017complex}. While these approaches remain effective in well-understood and stationary scenarios, they exhibit important limitations in modern railway environments. Fixed thresholds lack adaptability to variations in operating conditions, component aging, and environmental influences, often leading to high false-alarm rates or missed detections. Furthermore, manual feature engineering is labor-intensive and frequently fails to capture complex nonlinear and cross-modal relationships observed in contemporary railway subsystems, such as traction motors, door actuation mechanisms, or pantograph-catenary interfaces. These limitations are exacerbated by the rarity of safety-critical fault events, which results in severe class imbalance and further degrades detection reliability~\cite{shim2022anomaly, baldyga2024anomaly, fernandez2025gan}.

Deep learning (DL) has emerged as a powerful alternative by enabling data-driven modeling of complex spatial and temporal patterns directly from raw or minimally processed sensor data~\cite{sun2021survey,wang2020deep}. A wide range of architectures, including convolutional neural networks (CNN), recurrent and attention-based models, autoencoders, and transformer-based frameworks, have demonstrated strong potential for railway anomaly detection~\cite{he2025defect, phaphuangwittayakul2024railtrack,larese2025transformer}. Representative applications include wheel-flat detection from vibration 
signals~\cite{komorski2021advanced, shaikh2025contribution}, 
electrical arc detection~\cite{gao2020automatic}, 
door obstruction detection~\cite{1}, 
and rail-surface inspection using visual or 
Light Detection and Ranging (LiDAR) 
data~\cite{chalapathy2019deep}. Despite this rapid methodological progress, the practical adoption of deep learning–based anomaly detection in operational railway systems remains limited.

A key challenge lies in the lack of system-level alignment between anomaly characteristics, sensing modalities, learning paradigms, evaluation practices, and deployment constraints~\cite{de2022survey,binder2023predictive}. Many reported approaches achieve high performance under controlled experimental conditions but fail to translate into robust, comparable, and deployable solutions when confronted with real-world railway data and hardware limitations~\cite{bris2026survey,cao2024railway}. This gap highlights the need for a structured and engineering-oriented analysis that jointly considers data characteristics, model design, evaluation criteria, and deployment feasibility. Motivated by this observation, this survey provides a comprehensive system-level review of deep learning–based anomaly detection approaches for railway systems, with particular emphasis on practical applicability, reproducibility, and deployment-oriented decision support.

\subsection{Research Questions}
\label{sec:research-questions}

This survey is structured around six research 
questions that collectively address the full 
pipeline from anomaly characterization to 
deployment in railway systems:

\begin{itemize}[noitemsep, topsep=2pt]

\item \textbf{RQ1:} How can railway anomalies 
be systematically characterized to support 
consistent method selection across heterogeneous 
subsystems and sensing modalities?
(Section~\ref{sec:Railway Anomalies 
Categorization})

\item \textbf{RQ2:} What data-centric challenges 
recur across the surveyed literature, and which 
datasets are available for railway anomaly 
detection in terms of accessibility, modality, 
and subsystem coverage?
(Section~\ref{sec:1})

\item \textbf{RQ3:} Which deep learning detection 
paradigms are applied to railway anomaly detection, 
and how do their properties align with the 
characteristics of railway data and safety 
requirements?
(Section~\ref{sec:3})

\item \textbf{RQ4:} How is performance evaluated 
in the surveyed studies, and do existing evaluation 
practices adequately reflect the safety-critical 
nature of railway anomaly detection?
(Section~\ref{sec:4})

\item \textbf{RQ5:} How are deep learning models 
deployed in railway applications, and what hardware 
constraints, optimization strategies, and 
real-time requirements govern these deployments?
(Section~\ref{sec:deployment})

\item \textbf{RQ6:} What decision-oriented 
framework can guide practitioners in selecting 
detection paradigms and deployment configurations 
for specific railway anomaly scenarios?
(Section~\ref{sec:decision_framework})

\end{itemize}

\subsection{Survey Methodology}
\label{subsec:methodology}

This survey follows a structured and reproducible literature review protocol inspired by the PRISMA guidelines~\cite{page2021prisma}, adapted to the requirements of an engineering-oriented analysis.

\subsubsection{Scope and Time Window}

The review focuses on peer-reviewed journal articles and conference papers published between January~2019 and 2026. This period corresponds to the widespread adoption of deep learning techniques in railway monitoring and condition-based maintenance.

\subsubsection{Search Strategy}

The literature search was conducted using IEEE Xplore and Google Scholar. IEEE Xplore was used as the primary structured database to ensure reliable and reproducible retrieval of peer-reviewed engineering studies. Google Scholar was used as a complementary source to broaden the search and capture relevant publications across major publishers, including Elsevier, Springer, and ACM.

The search query was applied to titles, abstracts, and keywords (when supported), and adapted to each database:

\begin{quote}
\small
(\textbf{"railway" OR "rail transit" OR "train" OR "metro" OR "tram"}) 
AND (\textbf{"anomaly detection" OR "fault detection" OR "fault diagnosis" OR "condition monitoring"}) 
AND (\textbf{"deep learning" OR "neural network" OR "CNN" OR "LSTM" OR "autoencoder" OR "transformer"})
\end{quote}

Results were filtered by publication period (2019--2026), document type (journal articles and conference papers), and language (English).

\subsubsection{Inclusion and Exclusion Criteria}

\begingroup

Studies were included if they:
\begin{itemize}[noitemsep, topsep=2pt]
\item apply at least one deep learning method,
\item focus on railway systems,
\item address anomaly detection, fault detection, or condition monitoring,
\item report quantitative experimental results,
\item provide sufficient methodological detail, including model design, training procedure, and evaluation protocol,
\item are published in peer-reviewed venues.
\end{itemize}

To ensure the quality and relevance of the survey, additional quality-oriented criteria were considered:
\begin{itemize}[noitemsep, topsep=2pt]
\item studies lacking methodological transparency or reproducibility were excluded,
\item studies published in peer-reviewed and established venues were prioritized to improve the reliability of the reviewed evidence.
\end{itemize}

Studies were excluded if they:
\begin{itemize}[noitemsep, topsep=2pt]
\item rely only on classical machine learning methods,
\item target non-railway domains,
\item lack quantitative evaluation,
\item are survey or review papers,
\item are not accessible in full text.
\end{itemize}

\endgroup

\subsubsection{Selection Process}

\begin{figure}[H]
    \centering
    \includegraphics[width=1\linewidth]{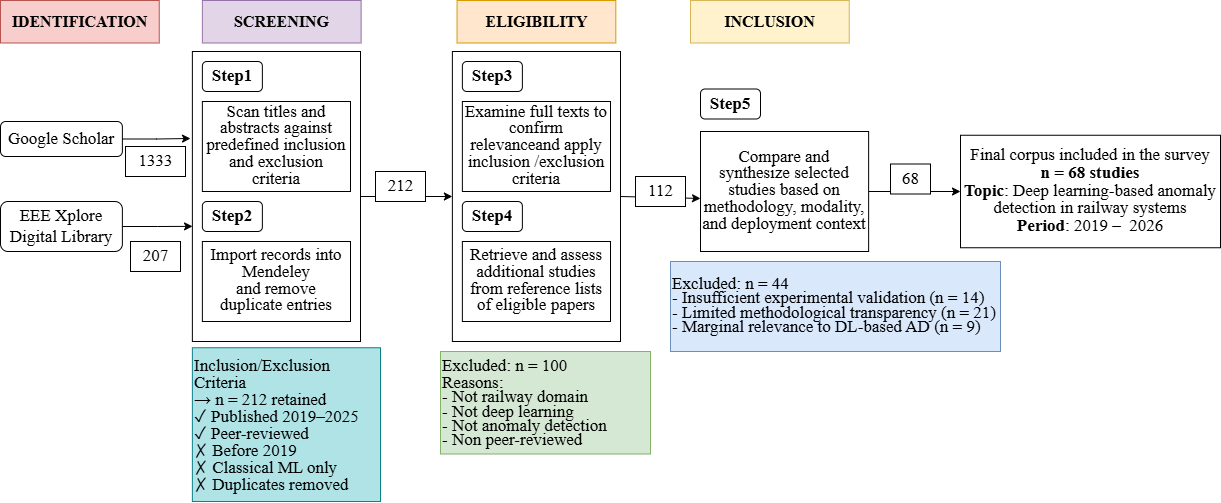}
    \caption{PRISMA-inspired flow diagram of the literature identification, screening, eligibility, and inclusion process adopted in this survey.}
    \label{fig:prisma}
\end{figure}

\begingroup

The study selection followed a PRISMA-inspired workflow (Figure~\ref{fig:prisma}).

Records were first retrieved from the selected databases and filtered based on publication period, document type, and language, followed by duplicate removal.

Titles and abstracts were then screened to retain only studies relevant to railway systems, deep learning, and anomaly or fault detection tasks.

The remaining studies were assessed through full-text analysis, where papers lacking sufficient experimental validation, methodological transparency, or relevance were excluded.

Finally, additional studies were identified through snowballing to improve coverage.

Overall, the process follows a \textbf{depth-over-breadth strategy}, retaining only studies with sufficient technical detail and experimental rigor for meaningful analysis. This strategy was adopted to avoid a purely descriptive accumulation of loosely related studies and to ensure that each included work could be analyzed consistently in terms of model design, data characteristics, evaluation protocol, and relevance to deep learning-based railway anomaly detection.

The exclusion counts and primary reasons are summarized in Figure~\ref{fig:prisma}: 100 records were excluded during screening because they were not related to railway systems, deep learning, anomaly detection, or peer-reviewed research. A further 44 records were excluded during full-text eligibility assessment due to insufficient experimental validation ($n=14$), limited methodological transparency ($n=21$), or marginal relevance to deep learning-based railway anomaly detection ($n=9$), yielding a final corpus of 68 studies.

\endgroup

\section{Related Work}
\label{sec:related-work}

Deep learning has rapidly emerged as a transformative approach for anomaly detection in railway systems, prompting several surveys that analyze progress and challenges in this field. While these reviews have advanced understanding, they remain largely fragmented~\cite{pang2021deep, chalapathy2019deep,you2026gated}, often limited to specific subsystems, single data modalities, or algorithmic benchmarks, without addressing DL methods categorization and deployment considerations such as hardware constraints, latency, and system integration. This section critically synthesizes existing surveys to contextualize and motivate the present work.

Several reviews provide broad overviews of artificial intelligence (AI) applications in railways. Oh et al.~\cite{oh2022review} cover infrastructure, rolling stock, and station safety, but their analysis is primarily descriptive and lacks discussion of deployment feasibility or modality-specific challenges. Baldyga et al.~\cite{baldyga2024anomaly} benchmark classical and deep-learning algorithms on time-series sensor data, revealing performance differences in operational environments, yet they do not address multimodal fusion or embedded constraints. Louren\c{c}o et al.~\cite{lourencco2024time} focus on predictive maintenance for wheel–track systems, emphasizing data sparsity and non-stationarity, but their scope is confined to a narrow subsystem. Tang et al.~\cite{tang2022literature} survey AI applications across seven railway domains, including safety and autonomous driving, but provide only a high-level overview without in-depth analysis of learning strategies or deployment considerations.

Other surveys concentrate on vision-based or modality-specific techniques. Kumar et al.~\cite{kumar2024systematic} and Di Summa et al.~\cite{di2023review} review computer-vision approaches for defect detection, reporting high accuracy in controlled environments but limited generalization to multimodal data or operational settings. De Donato et al.~\cite{de2022survey} extend the analysis to audio and video modalities for nonintrusive predictive maintenance, yet do not discuss system-level integration or inference under hardware constraints. Binder et al.~\cite{binder2023predictive} provide a large-scale review of predictive maintenance methods across twenty railway components, classifying them by failure mode and algorithm type, but offering limited insights into real-time feasibility, interpretability, or embedded implementation.
More recently, Phusakulkajorn et al.~\cite{bris2026survey} 
provide a structured review of AI techniques for 
railway infrastructure maintenance, covering track 
geometry, vibration monitoring, and imaging-based 
inspection across multiple subsystems. While 
comprehensive in scope, their analysis focuses on 
maintenance strategies rather than anomaly detection 
paradigms, and does not address deployment constraints 
or hardware-aware model design. Similarly, 
Hadj-Mabrouk et al. ~\cite{hadj2026contributions} 
surveys AI and machine learning applications in
railway operations and safety, but focuses primarily 
on classical AI techniques and does not address
deep learning-based anomaly detection or 
embedded deployment.

Table~\ref{tab:survey-comparison} summarizes the coverage of these surveys across major dimensions, including subsystem scope, deep-learning strategies, data challenges, and deployment considerations.

\begin{table}[htbp]
\centering
\caption{Comparative Overview of Survey Coverage 
in Railway AI. Infra: infrastructure, Env: 
environmental, DL Strat: DL Strategy, 
Data Chal: Data Challenges, Depl: Deployment.}
\label{tab:survey-comparison}
\scriptsize
\renewcommand{\arraystretch}{1.3}
\begin{tabular}{@{}|>{\centering\arraybackslash}
p{3.2cm}|p{0.8cm}|p{0.8cm}|p{0.8cm}|p{0.8cm}|
p{0.8cm}|p{0.8cm}|p{0.7cm}|@{}}
\cline{1-8}
\textbf{Survey} & \textbf{Year} & 
\textbf{Infra.} & \textbf{Train} & 
\textbf{Env.} & \textbf{DL Strat.} & 
\textbf{Data Chal.} & \textbf{Depl.} \\
\cline{1-8}
Oh et al.~\cite{oh2022review}
    & 2022 & \cmark & \cmark & \cmark 
    & \cmark & \xmark & NT \\
\cline{1-8}
Louren\c{c}o et al.~\cite{lourencco2024time}
    & 2024 & \cmark & \cmark & \xmark 
    & \cmark & \cmark & NT \\
\cline{1-8}
Tang et al.~\cite{tang2022literature}
    & 2022 & \cmark & \cmark & \cmark 
    & \cmark & \xmark & NT \\
\cline{1-8}
Kumar et al.~\cite{kumar2024systematic}
    & 2024 & \cmark & \cmark & \xmark 
    & \xmark & \xmark & NT \\
\cline{1-8}
De Donato et al.~\cite{de2022survey}
    & 2022 & \cmark & \xmark & \xmark 
    & \cmark & \xmark & T \\
\cline{1-8}
Di Summa et al.~\cite{di2023review}
    & 2023 & \cmark & \xmark & \xmark 
    & \xmark & \xmark & NT \\
\cline{1-8}
Ba\l{}dyga et al.~\cite{baldyga2024anomaly}
    & 2024 & \cmark & \cmark & \xmark 
    & \xmark & \cmark & NT \\
\cline{1-8}
Binder et al.~\cite{binder2023predictive}
    & 2023 & \cmark & \cmark & \cmark 
    & \xmark & \xmark & NT \\
\cline{1-8}
Phusakulkajorn et al.~\cite{bris2026survey}
    & 2025 & \cmark & \xmark & \xmark 
    & \cmark & \cmark & NT \\
\cline{1-8}
Hadj-Mabrouk~\cite{hadj2026contributions}
    & 2025 & \cmark & \cmark & \xmark 
    & \xmark & \xmark & NT \\
\cline{1-8}

\textbf{Our Survey}
    & \textbf{2026} & \cmark & \cmark & \cmark 
    & \cmark & \cmark & T \\
\cline{1-8}
\end{tabular}
\vspace{1ex}
\footnotesize
Note: \cmark~=~covered; \xmark~=~not covered; 
T~=~treated; NT~=~not treated.
\end{table}

\noindent
\textbf{Column descriptions:}
\begin{itemize}[noitemsep, topsep=2pt]
    \item \textbf{Infra.}, \textbf{Train.}, \textbf{Env.}: Coverage of anomaly detection in infrastructure (e.g., rails, joints), train subsystems (e.g., brakes, doors), and environmental scenarios (e.g., passenger behavior, environmental hazards).
    \item \textbf{DL Strat.}: Whether the survey categorizes deep-learning methods by detection paradigm (classification, reconstruction, prediction, hybrid).
    \item \textbf{Data Ch.}: Whether data-centric challenges, such as class imbalance, noise, or drift, are analyzed.
    \item \textbf{Deploy.}: Whether deployment aspects, including edge inference, hardware constraints, or model compression, are treated.
\end{itemize}

\textbf{Synthesis and Positioning:}
A close inspection of prior railway anomaly-detection surveys reveals several recurring limitations that reduce their usefulness for system-level engineering.
First, the literature remains \textbf{fragmented by subsystem and modality}: many surveys focus on isolated domains (e.g., vision-based track inspection, wheel-rail dynamics, or vibration monitoring) and therefore do not provide a unified perspective spanning infrastructure, rolling stock, and operational or environmental anomalies.
Second, deep-learning methods are frequently reported in a \textbf{catalog-style} manner, listing architectures or applications without sufficient abstraction into \textbf{actionable detection paradigms}. As a result, readers receive limited guidance on which learning strategy (classification, reconstruction, prediction, or hybrid) is appropriate under specific anomaly characteristics and data conditions.
Third, \textbf{deployment and decision constraints} are often treated as secondary aspects. In practice, however, railway feasibility is governed by latency, memory and power budgets, reliability requirements, and operational risk, which are rarely integrated into the methodological discussion.

\textbf{Contributions and Novelty of This Survey:}
Rather than a descriptive aggregation of prior work, this survey provides a design- and decision-oriented synthesis of deep learning-based railway anomaly detection. Its main contributions are four complementary elements that support systematic method selection and deployment planning:
\begin{itemize}[noitemsep, topsep=2pt]
    \item \textbf{Four-axis anomaly taxonomy:} A unified taxonomy based on location, data manifestation, sensing modality, and temporal behavior, enabling consistent characterization of anomalies across heterogeneous railway subsystems and data sources.
    \item \textbf{Paradigm-level organization of deep learning:} A structured classification of existing methods into classification-, reconstruction-, prediction-, and hybrid-based paradigms, with explicit links between paradigm choice, anomaly properties, and data regimes.
\item \textbf{Evaluation metrics and assessment practices:}
We review the evaluation metrics and experimental practices used in existing railway anomaly detection studies, and we propose an extended evaluation framework that explicitly defines how anomaly detection performance should be assessed for railway systems beyond standard accuracy reporting.
\item \textbf{Deployment analysis and decision-oriented framework:}
We review how deep learning models are deployed in railway applications by analyzing deployment architectures, hardware platforms, and optimization strategies, and we propose a decision-oriented framework that guides the selection of detection approaches and deployment configurations under real-world railway constraints.

\end{itemize}

\section{Railway Anomalies Categorization}
\label{sec:Railway Anomalies Categorization}

\begin{figure}[h]
  \centering
     \includegraphics[width=\linewidth]{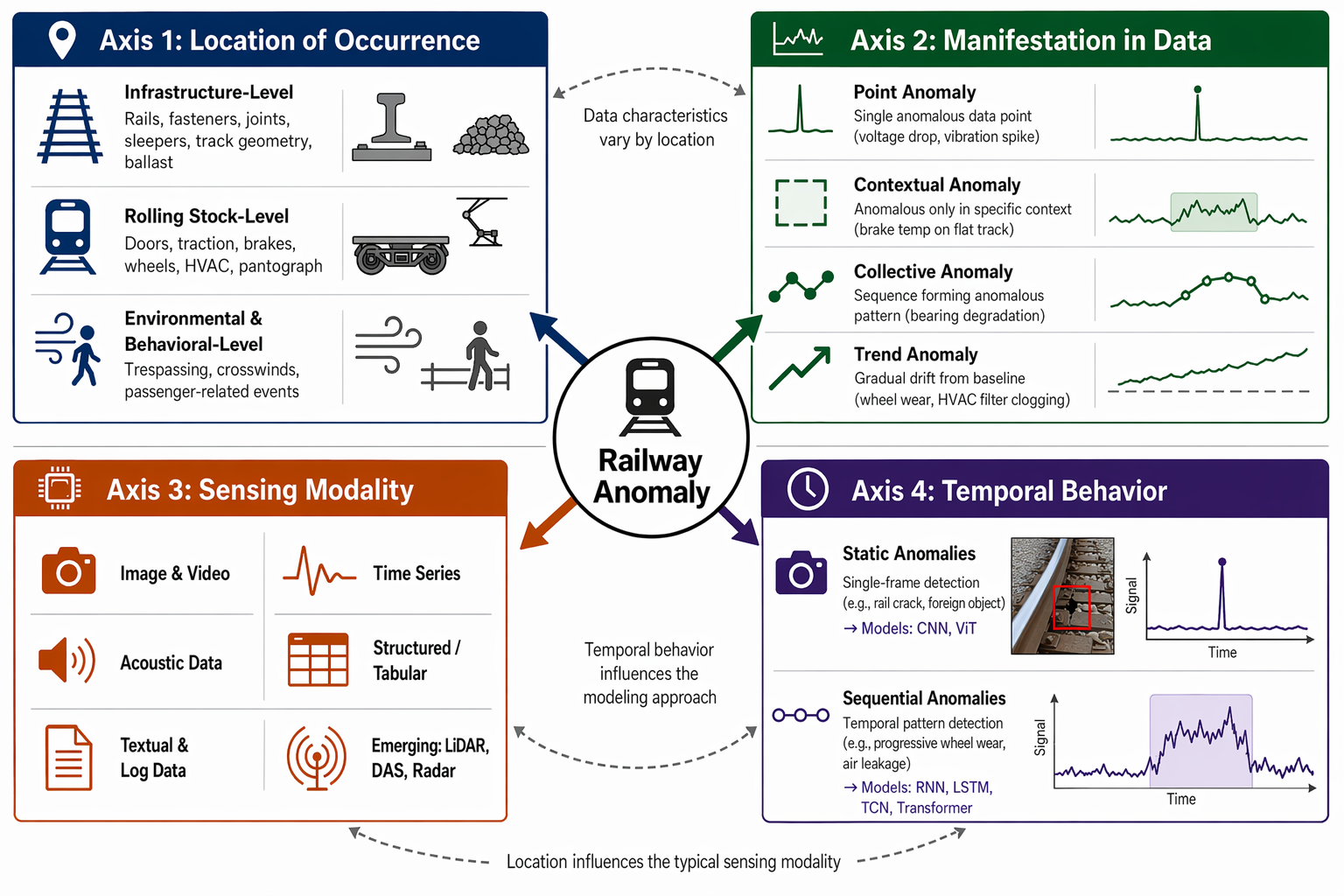}
     \caption{Four-axis taxonomy for railway anomaly characterization: location of occurrence, manifestation in data, sensing modality, and temporal behavior.}
     \label{fig:taxonomy}
 \end{figure}

Railway anomalies exhibit diverse characteristics depending on their physical origin, data manifestation, sensing modality, and temporal dynamics~\cite{baldyga2024anomaly, zamanzadeh2024deep}. To support systematic analysis and model selection, we propose a four-axis taxonomy that links anomaly types to appropriate detection strategies and deployment considerations. This framework facilitates reproducibility, multimodal integration, and real-world applicability. Figure~\ref{fig:taxonomy} provides an overview of the four axes and their categories.

\subsection{Categorization by Location of Occurrence}

Anomalies can be first classified by the physical domain in which they arise~\cite{baldyga2024anomaly, de2022survey}:

\begin{itemize}[noitemsep, topsep=2pt]

    \item \textbf{Infrastructure-Level}: Includes faults in static assets such as rails (e.g., cracks, corrugation), insulated joints (spark erosion), ballast and fasteners (fouling, loosening), sleepers (cracking, misalignment), and track geometry (e.g., curvature or super-elevation  errors)~\cite{bris2026survey, lourencco2024time}.

    \item \textbf{Rolling Stock-Level}: Covers dynamic subsystems such as train doors, 
    traction and braking systems, wheels and bogies, pantographs, HVAC units, and structural components of the train body~\cite{cheng2020review, binder2023predictive}.

    \item \textbf{Environmental and Behavioral-Level}: 
    Encompasses anomalies arising from environmental interaction or user behavior, including trespassing, extreme weather (e.g., crosswinds), and unsafe passenger actions~\cite{liu2021short, tang2022literature,du2024study}.

\end{itemize}

\subsection{Categorization by Manifestation in Data}

Anomalies can also be distinguished by how they deviate from expected data patterns~\cite{chandola2009anomaly, zamanzadeh2024deep}:

\begin{itemize}[noitemsep, topsep=2pt]

    \item \textbf{Point Anomalies}: Single data points that significantly deviate from the norm, such as a sudden voltage drop or vibration spike~\cite{chandola2009anomaly}.

    \item \textbf{Contextual Anomalies}: Values that are anomalous only within a specific context (temporal, spatial, or operational). For example, elevated brake temperature may be normal during descent but anomalous on a level track~\cite{chandola2009anomaly, malhotra2015long}.

    \item \textbf{Collective Anomalies}: Sequences of individually normal observations that form an anomalous pattern, such as gradual bearing degradation or pantograph instability~\cite{chandola2009anomaly, zamanzadeh2024deep}.

    \item \textbf{Trend Anomalies}: Gradual and persistent shifts in the statistical distribution of sensor data that deviate from long-term baseline behavior. Representative examples include progressive wheel wear, rail surface degradation, or HVAC filter clogging, where no single observation is anomalous but the cumulative drift signals a developing fault~\cite{zamanzadeh2024deep, correia2024online}.

\end{itemize}

\subsection{Categorization by Sensing Modality}

The type of sensor and data captured directly influences the detection strategy and deployment feasibility~\cite{de2022survey, baldyga2024anomaly}:

\begin{itemize}[noitemsep, topsep=2pt]

    \item \textbf{Image and Video}: Used for visual inspection of structural defects and dynamic events. Widely applied to rail surface inspection, fastener detection, and foreign object recognition~\cite{kumar2024systematic, di2023review}.

    \item \textbf{Time Series}: Includes vibration, temperature, current, and pressure signals acquired from onboard and wayside sensors. These signals are often non-stationary and exhibit strong temporal dependencies, requiring sequential modeling approaches~\cite{baldyga2024anomaly, lourencco2024time}.

    \item \textbf{Acoustic Data}: Captures contact or ambient sounds, often transformed into spectrograms for convolutional neural network (CNN) or attention-based analysis~\cite{javed2020anomaly, siddiqui2022iot}.

    \item \textbf{Structured and Tabular Data}: Includes numerical logs, operational alarms, and event timestamps, commonly used for condition monitoring and fault diagnosis~\cite{binder2023predictive}.

    \item \textbf{Textual and Log Data}: Comprises maintenance records and incident reports, processed using Natural Language Processing (NLP) techniques such as transformers or embedding models~\cite{ferrari2018detecting}.

    \item \textbf{Emerging Modalities}: Light Detection and Ranging (LiDAR), radar, multispectral sensors, and Distributed Acoustic Sensing (DAS) are increasingly investigated for enhanced spatial perception and environmental awareness~\cite{nateghinia2025development, chen2024lidar, yurekli2025das}.

\end{itemize}

\subsection{Categorization by Temporal Behavior}

Temporal dynamics further differentiate anomaly types and directly guide model selection~\cite{gasparini2021anomaly, zamanzadeh2024deep}:

\begin{itemize}[noitemsep, topsep=2pt]

    \item \textbf{Static Anomalies}: Detectable from a single snapshot, such as rail cracks or  foreign objects on tracks. Typically addressed using CNN-based visual inspection~\cite{choi2024deep, kumar2024systematic}.

    \item \textbf{Sequential Anomalies}: Require temporal modeling to capture evolving patterns, such as progressive wheel wear or air leakage. Recurrent Neural Networks (RNNs), Long Short-Term Memory networks (LSTMs), Temporal Convolutional Networks (TCNs), and Transformer architectures are commonly employed~\cite{wang2025spatial, bai2018empirical, ding2026integrated}.

\end{itemize}

Temporal categorization guides the selection of models with appropriate memory and sequence learning capabilities.

\section{Data Challenges, Preprocessing, and Datasets  for Railway Anomaly Detection}
\label{sec:1}

\subsection{Data-Centric Challenges and Preprocessing Strategies}

The performance of deep learning-based anomaly detection systems in railway applications is strongly influenced by the quality, structure, and variability of sensor data collected from onboard and wayside monitoring systems~\cite{bernal2018onboard, castillo2020critical}. The following challenges are particularly critical and recur across the surveyed literature.

\begin{itemize}[noitemsep, topsep=2pt]
\item\textbf{Severe class imbalance:}
Safety-critical faults such as wheel flats, pantograph arc events, and door obstructions occur far less frequently than normal operation, causing deep models to overfit the majority class. Oversampling techniques such as the Synthetic Minority Oversampling Technique~\cite{chawla2002smote} and Adaptive Synthetic Sampling~\cite{he2008adasyn}, and Generative Adversarial Network-based augmentation~\cite{liu2023anomaly} can partially mitigate this imbalance, but must preserve temporal integrity, particularly for cycle-based systems such as door actuation sequences.
    \item \textbf{Contextual and operational variability:}
Normal behavior in railway systems depends on train speed, weather conditions, vehicle type, passenger load, and track geometry~\cite{xu2020train}. Models trained under specific operational conditions often fail to generalize across different fleets or routes. Context-aware normalization, route-specific calibration, and domain adaptation methods~\cite{su2020adapting} are essential to improve cross-fleet and cross-network generalization.

\item \textbf{Sensor noise and missing data:}
Vibration signals are affected by wheel--rail roughness; current measurements suffer from power-electronic interference; and wireless sensor streams frequently contain missing or corrupted segments~\cite{shah2025recent}. Signal filtering, interpolation, and denoising autoencoders are commonly applied to improve signal quality for downstream deep learning models~\cite{shang2021vibration}.

\item \textbf{Multimodal misalignment:}
Integrating heterogeneous data streams from vibration sensors, cameras, acoustic sensors, and current logs is challenging due to different sampling rates and asynchronous time bases. Cross-modal representation learning~\cite{baltruvsaitis2018multimodal}, dynamic time-warping alignment, and late-fusion architectures enable more reliable multimodal anomaly detection.

\item \textbf{Limited availability of public datasets:}
Safety and confidentiality constraints severely restrict access to operational railway data. A systematic review by Pappaterra et al.~\cite{pappaterra2021systematic} revealed that the number of publicly available railway datasets remains critically low relative to the volume of published research. The present survey confirms that this situation has only partially improved between 2019 and 2025: of the 32 datasets identified, fewer than one third are fully publicly accessible. Simulated datasets derived from digital twins~\cite{openrail2025} and transfer learning strategies~\cite{huang2025ai} represent promising directions to address this barrier.

\end{itemize}

\subsection{Datasets for Railway Anomaly Detection}
\label{sec:2}

This section provides a structured overview of the datasets used across the surveyed literature. A total of \textbf{32 datasets} were identified, covering infrastructure, rolling stock, and environmental monitoring contexts. These datasets exhibit substantial heterogeneity in terms of origin, accessibility, and sensing modality~\cite{gasparini2021anomaly, chalapathy2019deep}, which directly affects model design choices, evaluation protocols, and reproducibility of reported results.

\textbf{Origin of the data:}
Four categories are distinguished: 
\begin{itemize}[noitemsep, topsep=2pt]
    \item \textbf{Real-world} datasets, collected during 
in-service operations using accelerometers, cameras, 
current sensors~\cite{liu2021short, shimizu2022real}.
\item\textbf{Simulated} datasets, generated through 
digital twins or physics-based simulators~\cite{liu2020real}.
\item\textbf{Synthetic} datasets, produced via 
data augmentation or generative models to address 
rare-event scarcity~\cite{wang2022deep}.
\item\textbf{Laboratory} datasets, recorded on 
test rigs or scaled-down prototypes, offering 
clean annotations but limited operational 
representativeness~\cite{ham2019comparative}.
\end{itemize}

\textbf{Accessibility of the data:}
Accessibility describes the level of public availability of a dataset, independently of its origin. Railway datasets range from fully public benchmarks to partially public datasets shared upon request, and to restricted datasets owned by railway operators or industrial partners. Limited accessibility remains a major barrier to fair benchmarking and reproducibility~\cite{chalapathy2019deep, gasparini2021anomaly}. Among the surveyed datasets, notable fully public examples include MetroPT-3 for air production unit monitoring~\cite{veloso2022metropt}, Rail Surface Defect Datasets~\cite{niu2020unsupervised}, Rail-5k~\cite{zhang2021rail5}, FaultSeg for wheel defect detection~\cite{shaikh2025faultseg}, and the pantograph arc datasets collected from European railway networks~\cite{signorino2020dataset, mariscotti2020datasets}.

To ensure consistency with the anomaly taxonomy introduced in Section~\ref{sec:Railway Anomalies Categorization}, the surveyed datasets are organized by their location of occurrence into three categories: infrastructure-level, rolling-stock-level, and environmental or behavioral datasets. Dataset scale is reported as the exact sample count when disclosed in the original study, and marked as NR (not reported) otherwise.

The \textbf{Labeled} column indicates whether the dataset provides ground-truth annotations: \textbf{Yes} means fully annotated samples; \textbf{Partial} means annotations are incomplete or derived from maintenance reports only; and \textbf{No} means no annotations are provided, restricting usage to unsupervised approaches.

\noindent\textbf{1) Infrastructure-Level Datasets:}
These datasets focus on track geometry, fastener condition, rail surface defects, weld integrity, and catenary inspection (Table~\ref{tab:infra}).

\begin{table}[H]
\centering
\caption{Infrastructure-Level Railway Datasets.
Accessibility(Acc.): Public~(P), Restricted~(R).
NR = not reported.}
\label{tab:infra}
\renewcommand{\arraystretch}{1.3}
\scriptsize
\begin{tabular}{@{}|p{0.5cm}|p{1.2cm}|p{0.4cm}
|p{1.4cm}|p{1.4cm}|p{1.4cm}|p{0.9cm}|p{2cm}|@{}}
\cline{1-8}
\textbf{Ref.} & \textbf{Dataset name} & 
\textbf{Acc.} & \textbf{System} &
\textbf{Modality} & 
\textbf{Scale (samples)} & \textbf{Labeled}&\textbf{Fault Types} \\
\cline{1-8}

\cite{sresakoolchai2022railway} & 
Track-Geom-Defect & R & 
Railway track &
Track geometry & 
172 436 & Yes&
Rail, switch, crossing defects \\
\cline{1-8}
\cite{siddiqui2022iot} & 
Rail-Acoustic & R & 
Track, fasteners &
Audio signals & 
1625& Yes&
Wheel burn, loose nut-bolt \\
\cline{1-8}
\cite{kaya2025detection} & 
IRJ-SparkVision & R & 
Rail joints (IRJ) &
RGB images & 
$\sim$28 150 &Yes& 
Spark erosion, rail damage \\
\cline{1-8}
\cite{yilmazer2022mask} & 
Drone Fastener & R & 
Rail fasteners &
RGB images & 
500 & Yes&
Missing fastener \\
\cline{1-8}
\cite{rakshit2022railway} & 
Track Fault & P & 
Rails, fasteners &
RGB images & 
293 & Partial&
Cracked rails, rusted bolts \\
\cline{1-8}

\cite{gao2023railway} & 
Fastener Vision & R & 
Rail fasteners &
RGB + depth & 
3 300  & Yes&
Elastic strip, Nut, Baffle miss \\
\cline{1-8}
\cite{yuan2021unsupervised} & 
RailSquat-ABA & P & 
Rail surface &
Vibration signals & 
NR & Partial&
Rail squat, surface defect \\
\cline{1-8}
\cite{cui2019real} & 
Fastener 3D & R & 
Rail fasteners &
3D point cloud & 
$\sim$5 500 & Yes&
Skewed spring bar \\
\cline{1-8}
\cite{gao2020automatic} & 
6C System & R & 
Pantograph &
Video + signals & 
NR & Partial&
Pantograph arc, catenary defects \\
\cline{1-8}
\cite{niu2020unsupervised} & 
RSDDs-113 & P & 
Rail surface &
RGB images + Depth & 
113  & Yes&
Rolling scar, corrosion  \\
\cline{1-8}
\cite{zhang2021rail5} & 
Rail-5k & P & 
Rail surface &
RGB images & 
5000& Partial & Crack, spalling  \\
\cline{1-8}
\cite{arain2024railway} & 
RailSurface-Faults & P & Rail surface & RGB images & 5050  & Yes & Cracks, squats  \\
\cline{1-8}
\cite{zendel2019railsem19} & 
RailSem19 & P & Railway environment & Video frames & 8500  & Yes & Obstacles \\
\cline{1-8}
\cite{tagiew2023osdar23} & 
OSDaR23 & P & 
Track, obstacles &
LiDAR +RGB +IR 
& 1534  & Yes & Obstacles, foreign objects\\
\cline{1-8}
\cite{signorino2020dataset} & 
Pantograph Arc DC & P & 
Pantograph catenary &
Voltage + current & 
NR &No& 
Electric arc \\
\cline{1-8}
\cite{mariscotti2020datasets} & 
Pantograph Arc AC& P & 
Pantograph catenary&
Voltage + current & 
21 033  & No & Harmonic distortion, power disturbance \\
\cline{1-8}
\cite{liu2018drtrain} & 
DR-Train & P & 
Light rail &
Acceleration + GPS & 
NR & No&
 Track condition changes \\
\cline{1-8}
\end{tabular}
\end{table}

\noindent\textbf{2) Rolling Stock-Level Datasets:} These datasets capture onboard subsystems including doors, traction systems, braking systems, and heating, ventilation, and air conditioning units (Table~\ref{tab:rol}). They include vibration, acoustic, current, and multivariate signals, often recorded during real train operations.

\begin{table}[H]
\centering
\caption{Rolling Stock-Level Railway Datasets.
Accessibility(Acc.): Public~(P), Restricted~(R).
NR = not reported.}
\label{tab:rol}
\renewcommand{\arraystretch}{1.3}
\scriptsize
\begin{tabular}{@{}|p{0.5cm}|p{1.2cm}|p{0.4cm}
|p{1.4cm}|p{1.4cm}|p{1.4cm}|p{0.9cm}|p{2cm}|@{}}
\cline{1-8}
\textbf{Ref.} & \textbf{Dataset name} & 
\textbf{Acc.} & \textbf{System} &
\textbf{Modality} & 
\textbf{Scale (samples)} & \textbf{Labeled}&\textbf{Fault Types} \\
\cline{1-8}

\cite{liu2020real} & 
 HSR-Sim & R & Railway electrification system & Multivariate signals  & 103 200 & Yes & Short-circuit fault \\
\cline{1-8}
\cite{ham2019comparative} & 
 Train-DoorRig & R & Train doors & Current & 626 & Yes & Bearing wear, obstruction \\
\cline{1-8}
\cite{galvez2021fault} & 
RailHVAC-RUL & R & 
HVAC system &
Multivariate signals & 
NR & Yes & Air filter clogging \\
\cline{1-8}
\cite{sun2020fault} & 
PlugDoor Sound & R & 
Sliding doors &
Acoustic signals & 
164 & Yes & Door faults \\
\cline{1-8}
\cite{wang2021efficient} & 
Braking System & R & 
Braking system &
Multivariate signals & 
28996 & Yes & Pressure, braking force deviation \\
\cline{1-8}
\cite{shim2023methodology} & 
Wheel Flat & P & 
Wheel-rail &
Spectrogram images & 
4977  & Yes  & Wheel flat detection \\
\cline{1-8}
\cite{wang2022improved} & 
Rail Transit IoT & R & 
Multiple systems &
Multivariate streams & 
NR & No & Device anomalies (door, motor, HVAC)\\
\cline{1-8}
\cite{galvez2022hybrid} & 
HVAC Filter & R & 
HVAC air filter &
Multivariate signals & 
NR & Yes & Air filter degradation, clogging  \\
\cline{1-8}
\cite{veloso2022metropt} & 
MetroPT-3 & P & 
Air production &
Pressure, current & 
10 979 547  & No & Air leak, oil leak \\
\cline{1-8}
\cite{laurendin2021hazardous} & 
FRailTRI20-DOD & R & 
Passenger door  &
RGB video & 
44 099  & 
Yes & Passenger stuck, bag stuck \\
\cline{1-8}
\cite{shimizu2022real} & 
Railway Door & R & 
Door actuation &
Current signals & 
NR &No & Door actuation anomaly \\
\cline{1-8}
\cite{shaikh2025faultseg} & 
FaultSeg & P & 
Wheel defects &
RGB images & 
1872 & 
Yes & Cracks/Scratches, Shelling, Discoloration  \\
\cline{1-8}

\end{tabular}
\end{table}

\noindent\textbf{3) Environmental and Behavioral-Level Datasets:}
These datasets address hazards arising from environmental interactions and human behavior, including foreign object detection on tracks, trespassing events, obstacle intrusion, and wind-induced operational risks (Table~\ref{tab:oper}). 

\begin{table}[H]
\centering
\caption{Environmental and Behavioral-Level Railway 
Datasets for Anomaly Detection.
Accessibility(Acc.): Public~(P), Restricted~(R). NR = not reported.}
\label{tab:oper}
\renewcommand{\arraystretch}{1.3}
\scriptsize
\begin{tabular}{@{}|p{0.5cm}|p{1.2cm}|p{0.4cm}
|p{1.4cm}|p{1.4cm}|p{1.4cm}|p{0.9cm}|p{2cm}|@{}}
\cline{1-8}
\textbf{Ref.} & \textbf{Dataset name} & 
\textbf{Acc.} & \textbf{System} &
\textbf{Modality} & 
\textbf{Scale (samples)} & \textbf{Labeled}&\textbf{Fault Types} \\
\cline{1-8}

\cite{wang2022deep} & 
RFOD Track & R & 
Track surroundings &
RGB images & 
7235 & 
Partial & Foreign objects, obstacles \\
\cline{1-8}
\cite{ning2024research} & 
Foreign Object & R & 
Track area &
RGB images & 2534 & Yes & Pedestrians, falling rocks\\
\cline{1-8}
\cite{liu2021short} & 
HSR Strong-Wind & R & 
Wind-exposed track &
Meteoro-logical signals & 
NR & Yes  & Wind-induced speed restriction events \\
\cline{1-8}

\end{tabular}
\end{table}

\textbf{Critical perspective on dataset comparability and reported performance:} Despite the diversity of datasets identified in this survey, the absence of standardized benchmarks significantly limits the comparability and generalizability of reported results~\cite{pappaterra2021systematic,chen2025large}. Three structural limitations recur across the literature.

First, the majority of studies evaluate their models on proprietary or task-specific datasets with distinct fault definitions, sensing configurations, and operating conditions, making cross-study comparison unreliable. Vision-based infrastructure inspection datasets frequently report high accuracy or mean Average Precision scores under controlled conditions~\cite{rakshit2022railway, yilmazer2022mask}, whereas rolling stock monitoring datasets typically yield lower performance due to operational noise and variability~\cite{wang2022improved, galvez2022hybrid}.

Second, models validated on laboratory-scale or simulated datasets frequently demonstrate optimistic results that do not transfer to in-service conditions~\cite{liu2020real}. These discrepancies indicate that reported performance improvements are often influenced as much by dataset characteristics as by model architecture.

Third, documentation quality remains poor. Among the 32 datasets surveyed, several do not report their exact size, collection protocol, or known limitations-a shortcoming already identified by Pappaterra et al.~\cite{pappaterra2021systematic} in 2021 and not yet resolved in the recent literature. The absence of unified anomaly definitions, consistent train-test splits, and cross-dataset evaluation protocols further weakens reproducibility and hinders fair assessment of algorithmic progress. Establishing open, annotated, and subsystem-specific benchmarks remains a critical open challenge for the railway anomaly detection community.

\section{Deep Learning Paradigms for Railway Anomaly Detection}
\label{sec:3}

Railway anomaly detection encompasses a diverse range of tasks, including binary anomaly detection, multiclass fault classification, spatial localization, early fault prediction, and remaining useful life estimation~\cite{chalapathy2019deep, pang2021deep}. These tasks differ in their supervision requirements, temporal modeling complexity, and operational objectives. Binary detection distinguishes normal from abnormal states, multiclass classification targets specific fault types such as broken clips or valve malfunctions, localization extends detection to spatial mapping via segmentation or object detection, and temporal tasks such as early fault prediction leverage sequence models for prognostics~\cite{zamanzadeh2024deep}.

Regardless of the target task, the choice of deep learning strategy fundamentally determines how a model learns to distinguish normal from anomalous behavior. Based on our analysis of the 68 surveyed studies, we organize existing approaches into four paradigms according to their core detection principle. Each paradigm can support one or more of the above tasks depending on data availability, labeling conditions, and operational constraints:

\begin{itemize}[noitemsep, topsep=2pt]
    \item \textbf{Classification-based} approaches learn discriminative boundaries between normal and fault classes from labeled data.
    
    \item \textbf{Reconstruction-based} approaches detect anomalies as deviations from a learned normal reconstruction.
    
    \item \textbf{Prediction-based} approaches flag anomalies as significant departures from expected temporal patterns.
    
    \item \textbf{Hybrid} approaches combine two or more of the above paradigms to address complex scenarios where a single strategy is insufficient.
\end{itemize}

This taxonomy serves as the structural backbone of the survey and guides the analysis of methods across railway subsystems, sensing modalities, and deployment contexts.

\subsection{Classification-Based Approaches}
\label{sec:classification}

Classification-based anomaly detection assigns each input sample to a predefined class (either a binary label (normal or abnormal) or a specific fault type through supervised learning~\cite{baldyga2024anomaly, pang2021deep}. This paradigm is particularly effective when fault categories are defined and labeled data are available, covering scenarios such as broken fasteners, door malfunctions, wheel surface defects, or foreign object intrusion. Unlike reconstruction- or prediction-based methods, classification models learn explicit decision boundaries between known classes, making them well-suited for recurring and distinguishable anomalies.

\begin{figure}[h]
\centering
\includegraphics[width=0.9\linewidth]
{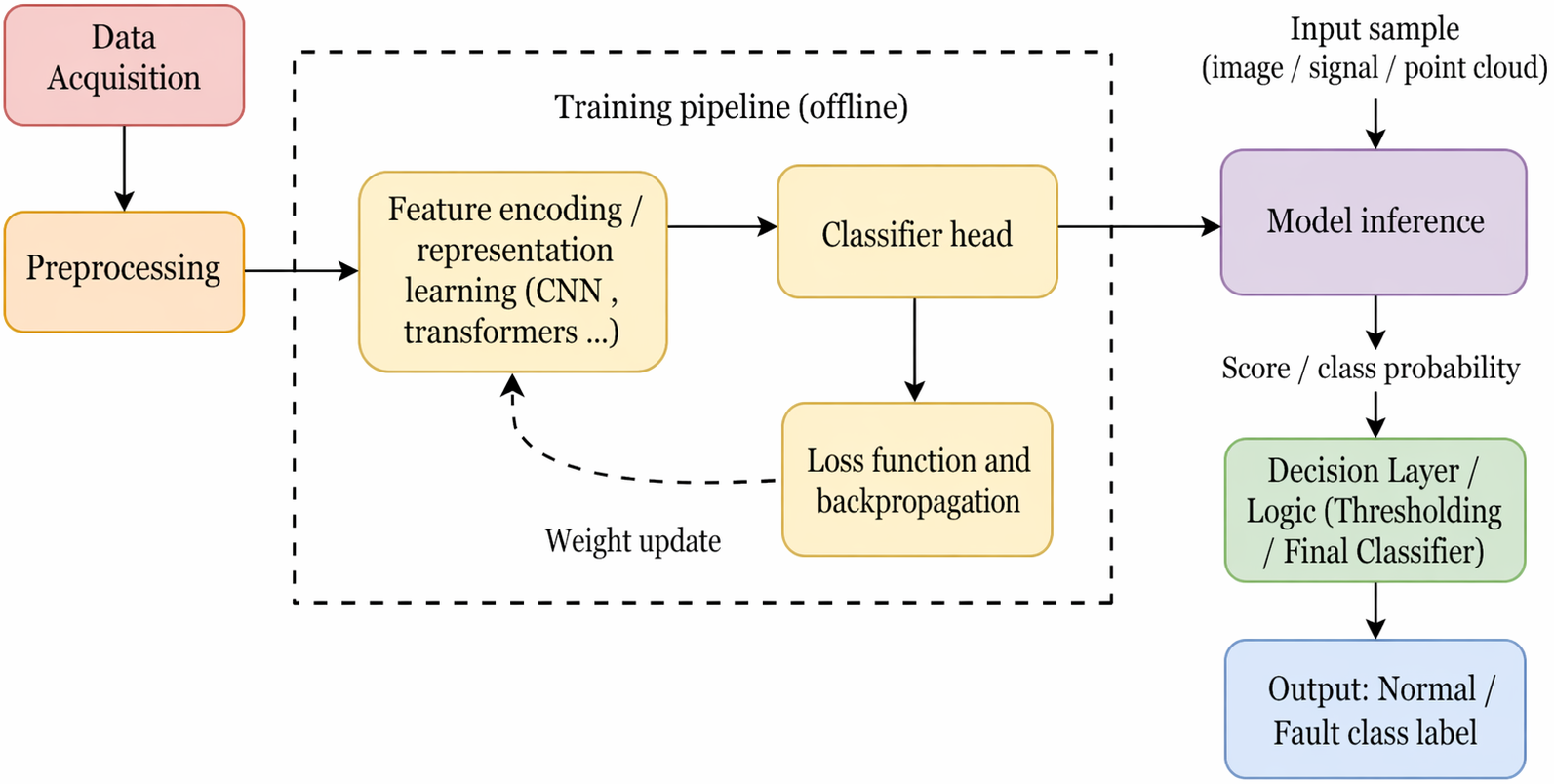}
\caption{Typical pipeline for 
classification-based anomaly detection 
in railway systems.}
\label{fig:classification_pipeline}
\end{figure}

Figure~\ref{fig:classification_pipeline} illustrates the typical pipeline: raw sensor inputs are acquired and preprocessed, then fed into a representation learning module (e.g., CNN~\cite{sresakoolchai2022railway}, Vision Transformer~\cite{mohammadi2025rail}, or PointNet++~\cite{cui2019real}) that extracts discriminative features. A classifier head produces class scores, and a decision layer applies thresholding or a final softmax to yield the fault label.

Based on our analysis of the surveyed studies, classification-based methods are organized into three groups according to their input modality. For each group, a summary table lists the surveyed works with their targeted subsystem, architecture, sensing modality, and reported metric. These tables are intended as a reference map to help identify which approaches address their specific railway component and data type and not as a direct performance comparison, since each study uses a different dataset and evaluation protocol.

\subsubsection{Classical Machine Learning (ML) Classifiers}

Classical ML pipelines combining handcrafted features with shallow classifiers remain relevant when labeled data are scarce or computation is constrained~\cite{chalapathy2019deep}. SVM (Support Vector Machine), Random Forest, and MLP (Multilayer Perceptron) have been applied to acoustic and vibration signals for door, HVAC, and track fault classification~\cite{shafique2021acoustic, sun2020fault, galvez2021fault, siddiqui2022iot, ham2019comparative}, achieving competitive accuracy on small datasets (Table~\ref{tab:cl_a}). However, their reliance on manual feature design limits scalability to complex or high-dimensional inputs, motivating the shift to deep learning.
\begin{table}[htbp]
\centering
\caption{Classical Machine Learning 
classifiers for railway anomaly detection.
All methods are supervised. 
Acc.: Accuracy.}
\label{tab:cl_a}
\renewcommand{\arraystretch}{1.25}
\scriptsize
\setlength{\tabcolsep}{3pt}
\begin{tabular}{@{}|p{0.55cm}|p{0.6cm}|
p{1.2cm}|p{1.8cm}|p{1.6cm}|
p{1.4cm}|p{3.5cm}|@{}}
\cline{1-7}
\textbf{Ref.} & \textbf{Year} & 
\textbf{System} & \textbf{Architecture} & 
\textbf{Modality} & \textbf{Metric} & 
\textbf{AD Task} \\
\cline{1-7}

\cite{shafique2021acoustic} & 2021 & 
Rail track & SVM, RF, MLP & 
Acoustic & Acc.: 97\% & 
Classify track faults (wheel burn, loose bolt) from acoustic signals. \\
\cline{1-7}
\cite{sun2020fault} & 2020 & 
Train doors & SVM (+MFCC features) & 
Acoustic & Acc.: $>$95\% & 
Classify plug-door fault types from acoustic measurements. \\
\cline{1-7}
\cite{galvez2021fault} & 2021 & 
HVAC system & Gradient Boosted Trees (GBT) & 
Multivariate & Acc.: 92.2\% & 
Detect air filter faults in railway HVAC from sensor data. \\
\cline{1-7}
\cite{siddiqui2022iot} & 2022 & 
Rail track & MLP + MFCC & 
Acoustic & Acc.: 98.4\% & 
Classify rail structural faults from IoT acoustic data. \\
\cline{1-7}
\cite{ham2019comparative} & 2019 & 
Train doors & KNN vs.\ CNN & 
Current & Acc.: 99.5\% & 
Detect abnormal current patterns in train door motors. \\
\cline{1-7}
\cite{shafique2023mfcc} & 2025 & 
Rail track & SVM, GRU, LSTM & 
Acoustic & Acc.: $>$96\% & 
Classify track faults using combined MFCC and CQT features. \\

\cline{1-7}
\end{tabular}
\end{table}
\subsubsection{Vision-Based Deep Classifiers}

Vision-based classification is the dominant paradigm in the surveyed literature, covering rail surface defects, fasteners, joints, and wheel anomalies from wayside and onboard imagery.

Standard CNNs with transfer learning (Residual Network-ResNet~\cite{koonce2021resnet}, Visual Geometry Group-VGG~\cite{tammina2019transfer}) effectively handle data scarcity in visual inspection~\cite{sresakoolchai2022railway, rakshit2022railway}. Ensemble CNN approaches combining VGG-19, MobileNetV3, and ResNet50 achieve 99\% accuracy on railway defect images~\cite{bouguettaya2024ensemble}, while YOLO-based architectures (You Only Look Once) consistently outperform static CNN classifiers for binary fault classification~\cite{autores2025yolo11}.

For real-time defect localization, attention-enhanced YOLO~\cite{jiang2022review} variants achieve strong speed-accuracy trade-offs under operational constraints~\cite{gao2023railway, kaya2025detection, brintha2024fodyolo, zhukov2024hybrid}. Lightweight transformer-based detectors such as RFD-DETR (Railway Fastener Defect Detection Transformer) further improve fastener defect detection under size-constrained deployment~\cite{zhou2025rfddetr}. Mask R-CNN achieves 98\% precision for missing fastener detection from drone imagery~\cite{yilmazer2022mask}, and hybrid segmentation-detection pipelines extend coverage to foreign object localization~\cite{ning2024research}.

Vision Transformers (ViT)\cite{khan2022transformers} overcome the locality bias of CNNs by capturing global structural dependencies. ViT and DeiT(Data-efficient Image Transformer) outperform CNN baselines for fastener (DeiT: 95.0\% vs VGG16: 91.5\%)~\cite{smart2025vit, mohammadi2025rail, phaphuangwittayakul2024railtrack}. Non-destructive testing pipelines further extend classification to subsurface defects by fusing camera and ultrasonic signals~\cite{xiong2023detection}.In parallel, Wang et al.~\cite{wang2024hierarchical} proposed a hierarchical transfer-generative framework that automates multiple analytical tasks in rail surface defect inspection simultaneously.

Table~\ref{tab:cl_b} summarizes the surveyed vision-based classification works.

\begin{table}[htbp]
\centering
\caption{Vision-based deep classifiers 
for railway anomaly detection.
All methods are supervised. 
Acc.: Accuracy; mAP: mean Average 
Precision; mIoU: mean Intersection 
over Union; Prec.: Precision.}
\label{tab:cl_b}
\renewcommand{\arraystretch}{1.25}
\scriptsize
\setlength{\tabcolsep}{3pt}
\begin{tabular}{@{}|p{0.55cm}|p{0.6cm}|
p{1.2cm}|p{1.8cm}|p{1.6cm}|
p{1.4cm}|p{3.5cm}|@{}}
\cline{1-7}
\textbf{Ref.} & \textbf{Year} & 
\textbf{System} & \textbf{Architecture} & 
\textbf{Modality} & \textbf{Metric} & 
\textbf{AD Task} \\
\cline{1-7}

\cite{sresakoolchai2022railway} & 2022 & 
Track geometry & CNN, DNN & 
Track geometry signals & Acc.: 94.3\% & 
Classify rail geometry defects. \\
\cline{1-7}
\cite{rakshit2022railway} & 2022 & 
Rail surface & ResNet-50, VGG-16 & 
RGB images & Acc.: 83.8\% & 
Detect surface defects (cracks, corrosion). \\
\cline{1-7}
\cite{bouguettaya2024ensemble} & 2024 & 
Rail and wheel & Ensemble CNN & 
RGB images & Acc.: 99\% & 
Classify rail and wheel surface defects. \\
\cline{1-7}
\cite{autores2025yolo11} & 2024 & 
Rail track & YOLOv11 + TL(Transfer Learning) & 
RGB images & Acc.: 92.5\% & 
Binary classification of track images: defective vs.\ normal. \\
\cline{1-7}
\cite{gao2023railway} & 2023 & 
Rail fasteners & CSP-Darknet53 & 
RGB + Depth & mAP: 86.6\% & 
Detect and localize missing and broken fasteners. \\
\cline{1-7}
\cite{kaya2025detection} & 2025 & 
Rail joints & SqueezeNet & 
RGB images & Acc.: 98.2\% & 
Detect and segment spark erosion 
on insulated rail joints. \\
\cline{1-7}
\cite{shaikh2025contribution} & 2025 & 
Wheel surface & YOLOv5-seg & 
RGB images & mAP: 92\% & 
Detect and segment wheel defects (flats, shelling, cracks, discoloration). \\
\cline{1-7}
\cite{brintha2024fodyolo} & 2024 & 
Fasteners + track & FOD-YOLONet & 
RGB images & Acc: 98.14\% & 
Simultaneously detect fastener faults and foreign objects on tracks. \\
\cline{1-7}
\cite{zhou2025rfddetr} & 2025 & 
Rail fasteners & RFD-DETR & 
RGB images & mAP: 98.27\% & 
Detect and localize fastener defects. \\
\cline{1-7}
\cite{zhukov2024hybrid} & 2024 & 
Rail surface & YOLOv5/v8 + CBAM & 
RGB images & Acc.: 80.3\% & 
Detect and classify rail defects. \\
\cline{1-7}
\cite{yilmazer2022mask} & 2022 & 
Rail fasteners & Mask R-CNN & 
RGB (drone) & Prec.: 98\% & 
Detect and segment missing fasteners from drone imagery. \\
\cline{1-7}
\cite{ning2024research} & 2024 & 
Track obstacles & U-Net + YOLOv5 & 
RGB images & mAP@0.5: 77.9\% & 
Detect foreign objects on tracks and assess intrusion risk. \\
\cline{1-7}
\cite{xiong2023detection} & 2023 & 
Rail welds & CNN + SVM & 
RGB images & Qualitative & 
Detect internal and surface weld flaws. \\
\cline{1-7}
\cite{mohammadi2025rail} & 2025 & 
Rail surface & CNN vs.\ ViT & 
RGB images & Acc.: $\sim$95\% & 
Classify rail surface and fastener defects comparing CNN and ViT. \\
\cline{1-7}
\cite{smart2025vit} & 2025 & 
Rail fasteners & ViT, DeiT & 
RGB images & Acc.: 95.0\% & 
Detect fastener defects. \\
\cline{1-7}
\cite{phaphuangwittayakul2024railtrack} & 2024 & 
Multi-component & RailTrack-DaViT & 
RGB images & Acc.: $>$98\% & 
Classify defects across rail, fastener, fishplate, and multi-fault categories. \\

\cline{1-7}
\end{tabular}
\end{table}

\subsubsection{Time-Series Signal Classifiers}

Rolling stock fault diagnosis relies on temporal signals from onboard sensors rather than visual inspection. CNN-based classifiers applied to spectrograms and time-frequency representations effectively detect wheel flats and braking system faults~\cite{shim2023methodology, wang2021efficient}. GRU-based hierarchical models enable ultra-low-latency multiclass fault classification in traction systems~\cite{liu2020real}, while 1D-CNN and LSTM architectures classify wheelset and suspension faults directly from raw acceleration signals~\cite{shafique2022wheelset}. Multimodal classifiers combining visual and acoustic streams extend this paradigm to passenger behavioral anomaly detection~\cite{tang2023comparative}.Table~\ref{tab:cl_c} summarizes the surveyed time-series classifiers.

\begin{table}[htbp]
\centering
\caption{Time-series signal classifiers 
for railway anomaly detection. 
All methods are supervised. 
Acc.: Accuracy; AUC: Area Under 
the ROC Curve; NR: Not Reported.}
\label{tab:cl_c}
\renewcommand{\arraystretch}{1.25}
\scriptsize
\setlength{\tabcolsep}{2.5pt}
\begin{tabular}{@{}|p{0.55cm}|p{0.6cm}|
p{1.2cm}|p{1.8cm}|p{1.6cm}|
p{1.4cm}|p{3.5cm}|@{}}

\cline{1-7}
\textbf{Ref.} & \textbf{Year} & 
\textbf{System} & \textbf{Architecture} & 
\textbf{Modality} & \textbf{Metric} & 
\textbf{AD Task} \\
\cline{1-7}

\cite{shim2023methodology} & 2023 & 
Wheel-rail & LeNet-5 CNN & 
Vibration signals & Acc.: 97\%, AUC: 0.97 & 
Detect and classify wheel flat 
defects from vibration spectrograms. \\
\cline{1-7}
\cite{wang2021efficient} & 2021 & 
Braking system & Multivariate CNN & 
Pressure + force & Acc.: 93\% & Classify braking faults from multivariate signals. \\
\cline{1-7}
\cite{liu2020real} & 2020 & 
Traction system & GRU hierarchical & 
Multivariate & Acc.: $>$93\% & 
Classify 29 traction fault types from high-frequency sensor streams. \\
\cline{1-7}
\cite{shafique2022wheelset} & 2023 & 
Wheelset & 1D-CNN, LSTM & 
Acceleration & NR & 
Detect and classify wheelset faults from raw acceleration time series. \\

\cline{1-7}
\cite{tang2023comparative} & 2023 & 
Passenger cabin & SlowFast CNN & 
Video + acoustic & Acc.: 85.1\% & 
Classify abnormal passenger behavioral events. \\

\cline{1-7}
\end{tabular}
\end{table}
\vspace{1ex}
\vspace{1ex}
\noindent\textbf{Summary and limitations:} Classification-based approaches offer high accuracy, interpretable outputs, and efficient deployment on embedded hardware~\cite{kaya2025detection, wang2021efficient}. However, three structural limitations constrain their applicability in open railway environments. First, they require curated labeled datasets per fault type, which is costly for rare or novel anomalies~\cite{chalapathy2019deep, pang2021deep}. Second, models trained on a closed fault set fail silently on out-of-distribution anomalies (a critical safety concern in safety-critical railway contexts)~\cite{rakshit2022railway}. Third, class imbalance between normal and fault samples requires active mitigation through loss reweighting or data augmentation~\cite{gao2023railway, chawla2002smote}.

\vspace{1ex}
\noindent\textbf{Cross-category observations:}
Across the three groups summarized in Tables~\ref{tab:cl_a}-\ref{tab:cl_c}, several trends emerge. Vision-based classifiers dominate the surveyed literature, reflecting the widespread deployment of wayside cameras and the availability of large annotated image datasets for infrastructure inspection. Classical ML remains competitive on small acoustic and vibration datasets but is systematically outperformed by deep models on raw high-dimensional inputs~\cite{ham2019comparative, sresakoolchai2022railway}. Time-series classifiers achieve strong results on rolling stock subsystems but are predominantly validated on simulated or single-operator datasets, raising concerns about operational generalizability~\cite{liu2020real, shafique2022wheelset}. Transformer-based architectures (ViT, DeiT, DETR) consistently outperform CNN baselines across visual inspection tasks at the cost of higher computational demands~\cite{mohammadi2025rail, smart2025vit, zhou2025rfddetr}. Finally, the heterogeneity of reported metrics confirms that standardized benchmarking remains an open challenge for the railway anomaly detection community~\cite{pang2021deep, chalapathy2019deep}. These limitations motivate the complementary reconstruction- and prediction-based paradigms reviewed in subsequent sections.

\subsection{Prediction-Based Approaches}
\label{sec:prediction}

Prediction-based anomaly detection learns the normal temporal dynamics of a system and flags observations that deviate significantly from predicted values~\cite{zamanzadeh2024deep, correia2024online, pang2021deep}. Unlike classification-based approaches, this paradigm is typically trained on normal operational data, reducing dependence on fault labels, making it particularly suitable for railway subsystems where fault examples are rare or unavailable such as traction motors, braking units, Heating Ventilation and Air Conditioning (HVAC) systems, and track geometry monitoring~\cite{iqbal2024anomaly, chalapathy2019deep}.

\begin{figure}[h]
\centering
\includegraphics[width=\linewidth]
{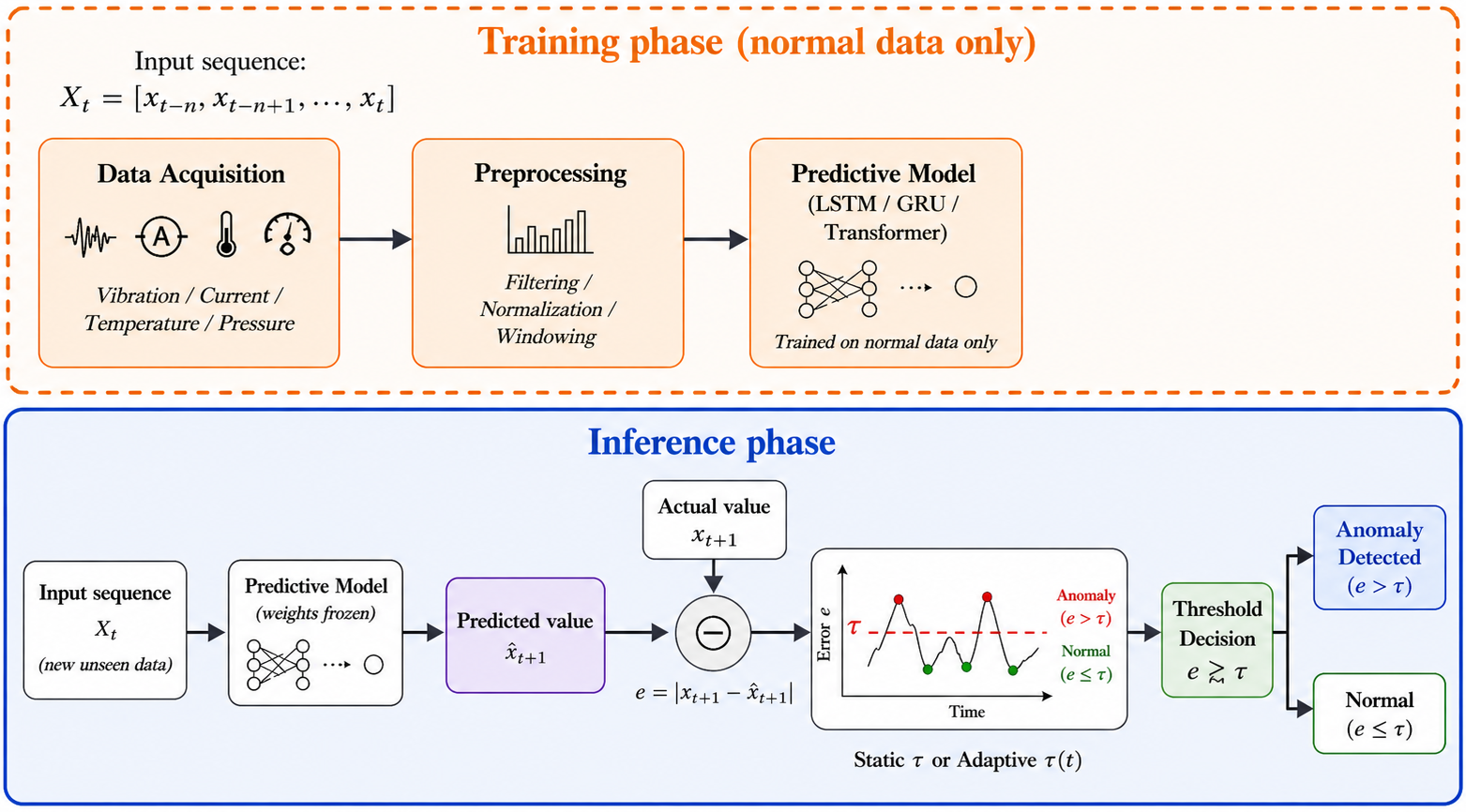}
\caption{Typical pipeline for 
prediction-based anomaly detection 
in railway systems. }
\label{fig:prediction-pipeline}
\end{figure}

As illustrated in Figure~\ref{fig:prediction-pipeline}, a deep predictive model learns to predict the next system state $\hat{x}_{t+1}$ from a sliding window of past observations $X_t = [x_{t-n}, \ldots, x_t]$. An anomaly score is computed as the prediction error $e = |x_{t+1} - \hat{x}_{t+1}|$, and a static or adaptive threshold triggers a fault alert when $e$ exceeds a predefined limit. The model is trained exclusively on normal data, enabling detection of previously unseen fault types without any fault labels~\cite{zamanzadeh2024deep}.

Based on our analysis of the surveyed studies, we organize prediction-based railway studies into three architecture-oriented groups ( recurrent sequence predictors, attention and transformer predictors, and convolutional temporal predictors)based on the dominant architectural family used in each work. For each group, a summary table lists the surveyed works.

\subsubsection{Recurrent Sequence Predictors}

Recurrent Neural Networks (RNNs) and their gated variants constitute the dominant family for prediction-based anomaly detection in railway systems~\cite{zamanzadeh2024deep, pang2021deep}. These architectures maintain a hidden state that evolves over time, enabling them to learn normal operational dynamics from sequential sensor data and detect anomalies as deviations between predicted and observed values.

Standard RNNs have been applied to anticipate unsafe operational conditions such as abnormal running behavior and derailment risk from multivariate onboard sensor streams~\cite{lee2021real}. However, standard recurrent networks suffer from gradient instability over long sequences, which limits their use in complex railway monitoring scenarios.

LSTM networks overcome this limitation through gated memory cells that selectively retain or discard information across time steps. In railway applications, LSTM predictors trained exclusively on normal operational data have been applied across multiple subsystems. Wang et al.~\cite{wang2022improved} trained a modified Long Short-Term Memory on normal data from heating ventilation and air conditioning units, and compressors to flag anomalies based on prediction error thresholds. Yokouchi and Kondo~\cite{yokouchi2021lstm} applied LSTM to condition monitoring data from railway vehicle air conditioning units, demonstrating that prediction-based detection can identify abnormal behavior without any fault labels. Hao et al.~\cite{hao2023track} applied LSTM to predict future track geometry parameters from inertial measurements, enabling early detection of misalignment before safety thresholds are exceeded. Ribeiro et al.~\cite{ribeiro2023lstm} further demonstrated that LSTM predictors can anticipate rolling stock equipment failures hours in advance using the MetroPT-3 dataset of compressor monitoring signals. Chen et al.~\cite{chen2024lstm} introduced a deep LSTM method for suspension fault detection, modeling normal vibration responses and flagging deviations from the learned baseline~(Table~\ref{tab:pred_a}).

\begin{table}[htbp]
\centering
\caption{Recurrent sequence predictors for 
railway anomaly detection. Models are trained 
on normal data only unless otherwise noted. 
F1: F1-score; Acc.: Accuracy; 
AUC: Area Under the ROC Curve; 
NR: not reported.}
\label{tab:pred_a}
\renewcommand{\arraystretch}{1.25}
\scriptsize
\setlength{\tabcolsep}{2.5pt}
\begin{tabular}{@{}|p{0.55cm}|p{0.6cm}|
p{1.3cm}|p{1.8cm}|p{1.6cm}|p{5cm}|@{}}
\cline{1-6}
\textbf{Ref.} & \textbf{Year} & 
\textbf{System} & 
\textbf{Architecture} & 
\textbf{Modality}  & 
\textbf{AD Task} \\
\cline{1-6}

\cite{lee2021real} & 2021 &
Rolling stock &
LSTM, RNN+ DNN &
Multivariate signals &

Predict derailment risk and abnormal running behavior from onboard sensors. \\
\cline{1-6}

\cite{wang2022improved} & 2022 &
HVAC, compressors &
Modified Long Short-Term Memory &Multivariate signals &

Predict normal system state and flag anomalies in multiple rolling stock subsystems. \\
\cline{1-6}

\cite{yokouchi2021lstm} & 2021 &
Air conditioning unit &
LSTM, RNN &
Temperature + current &

Detect anomalies in vehicle air conditioning units from monitoring data deviations. \\
\cline{1-6}

\cite{shimizu2022real} & 2022 &
Air conditioning unit &
LSTM &
Current signals &

Detect abnormal air conditioning units recorded in vehicule on operation. \\
\cline{1-6}

\cite{hao2023track} & 2023 &
Track geometry &
LSTM&
Inertial signals &

Forecast track geometry parameters to detect misalignment before safety limits are exceeded. \\
\cline{1-6}

\cite{ribeiro2023lstm} & 2023 &
Train traction converter cooling &
LSTM &
Multivariate signals 
 &
Predict rolling stock equipment failures hours in advance from sensor streams. \\
\cline{1-6}

\cite{chen2024lstm} & 2024 &
Suspension &
Deep LSTM &
Vibration signals &
Detect suspension faults by 
measuring deviations from 
learned normal vibration. \\
\cline{1-6}

\end{tabular}
\end{table}
\subsubsection{Attention and Transformer Predictors}

Attention mechanisms and Transformer architectures extend recurrent predictors by explicitly weighting informative sensor channels and critical time intervals, improving both accuracy and interpretability~\cite{correia2024online, zamanzadeh2024deep}.

Dual Attention Recurrent Neural Networks integrate input-level and temporal attention within an encoder-decoder structure. Liu et al.~\cite{liu2021short} applied this architecture to real-time wind risk prediction along high-speed rail corridors, where the attention mechanism captures spatial-temporal correlations across distributed weather sensor networks under weak supervision. Transformer-based architectures replace recurrence entirely with self-attention, enabling parallel computation and superior modeling of long-range temporal dependencies. Ding et al.~\cite{ding2026integrated} proposed Improved EmpiricalWavelet Transform (IEWT), a Transformer-based model for railway axle bearing fault prediction from vibration signals, demonstrating improved performance over recurrent baselines on long temporal sequences~(Table~\ref{tab:pred_b}).

\begin{table}[htbp]
\centering
\caption{Attention and Transformer predictors 
for railway anomaly detection. 
AUC: Area Under the ROC Curve; 
NR: not reported.}
\label{tab:pred_b}
\renewcommand{\arraystretch}{1.25}
\scriptsize
\setlength{\tabcolsep}{2.5pt}
\begin{tabular}{@{}|p{0.55cm}|p{0.6cm}|
p{1.5cm}|p{2.5cm}|p{1.6cm}|p{4.1cm}|@{}}
\cline{1-6}
\textbf{Ref.} & \textbf{Year} & 
\textbf{System} & 
\textbf{Architecture} & 
\textbf{Modality}  & 
\textbf{AD Task} \\
\cline{1-6}
\cite{liu2021short} & 2021 &
High-speed rail wind monitoring &
Multiple Attention Layer based Multi-Instance Learning
(MAL-MIL) &
Meteorological signals &
Predict wind hazard probability in real-time along high-speed rail corridors. \\
\cline{1-6}

\cite{ding2026integrated} & 2025 &
Axle bearing &
Improved Empirical Wavelet Transform (IEWT)+CNN &
Vibration signals &

Predict bearing degradation and detect faults from vibration time series. \\

\cline{1-6}

\end{tabular}
\end{table}

\subsubsection{Convolutional Temporal Predictors}

One-dimensional CNNs and Temporal Convolutional Networks (TCN) apply convolutional filters along the temporal dimension to capture local temporal patterns with significantly lower computational cost than recurrent architectures~\cite{chalapathy2019deep}. This makes them well-suited for embedded and edge deployment in resource-constrained railway environments.

Galvez et al.~\cite{galvez2022hybrid} combined physics-based degradation models with a one-dimensional CNN predictor to estimate the remaining useful life of railway heating ventilation and air conditioning air filters, detecting anomalies as deviations from the learned degradation trajectory. Haghbin et al.~\cite{haghbin2024cnn1d} applied 1D-CNN to predict rail corrugation profiles directly from vehicle acceleration measurements, achieving prediction errors below 5\% and demonstrating the feasibility of real-time infrastructure monitoring from onboard sensors. Zhang et al.~\cite{zhang2024tcngru} proposed a hybrid architecture combining self-attention, bidirectional TCNs, and bidirectional GRUs for predicting the evolution trend of rail corrugation, outperforming individual recurrent and convolutional baselines on a multi-domain feature set extracted from acceleration signals~(Table~\ref{tab:pred_c}).

\begin{table}[htbp]
\centering
\caption{Convolutional temporal predictors 
for railway anomaly detection. 
Acc.: Accuracy; NR: not reported; 
TCN: Temporal Convolutional Network; 
GRU: Gated Recurrent Unit.}
\label{tab:pred_c}
\renewcommand{\arraystretch}{1.25}
\scriptsize
\setlength{\tabcolsep}{2.5pt}
\begin{tabular}{@{}|p{0.55cm}|p{0.6cm}|
p{1.9cm}|p{1.8cm}|p{1.6cm}|p{4.4cm}|@{}}
\cline{1-6}
\textbf{Ref.} & \textbf{Year} & 
\textbf{System} & 
\textbf{Architecture} & 
\textbf{Modality}  & 
\textbf{AD Task} \\
\cline{1-6}

\cite{galvez2022hybrid} & 2022 &
Heating ventilation air conditioning 
filter &
1D CNN + physics model  &
Multivariate signals &
Estimate remaining useful life of air filter and detect degradation anomalies. \\
\cline{1-6}

\cite{haghbin2024cnn1d} & 2024 &
Rail corrugation &
1D CNN &
Acceleration signals &
Predict rail corrugation profile from vehicle acceleration to detect surface anomalies. \\
\cline{1-6}

\cite{zhang2024tcngru} & 2024 &
Rail corrugation &
Self-attention + TCN + GRU &
Acceleration signals &
Predict corrugation evolution trend and detect anomalous degradation progression. \\
\cline{1-6}

\end{tabular}
\end{table}
\vspace{1ex}
\noindent\textbf{Summary and limitations:}
Prediction-based approaches offer flexibility and generalizability for railway monitoring, requiring only normal operational data and enabling the detection of previously unseen fault types~\cite{pang2021deep, chalapathy2019deep}. However, three limitations constrain their applicability. First, prediction errors are sensitive to sensor noise and operational drift, which may generate false alarms unrelated to genuine faults~\cite{correia2024online}. Second, threshold calibration is non-trivial and must balance detection sensitivity against false alarm rates~\cite{zamanzadeh2024deep}. Third, models trained on normal data from one operational context often degrade significantly when deployed under different speed, load, or environmental conditions~\cite{wang2022improved, hao2023track}.

\vspace{1ex}
\noindent\textbf{Cross-category observations:}
Across the three groups summarized in Tables~\ref{tab:pred_a}-\ref{tab:pred_c}, LSTM architectures dominate the surveyed railway literature, reflecting their established stability over long sensor sequences and their suitability for multivariate onboard monitoring~\cite{wang2022improved, yokouchi2021lstm, ribeiro2023lstm}.Attention and Transformer-based predictors are emerging as strong alternatives for both rolling stock and infrastructure monitoring, have demonstrated competitive or superior performance compared to recurrent baselines in several railway studies when long-range temporal dependencies are present~\cite{ding2026integrated, liu2021short}. Convolutional temporal predictors represent a practical option for resource-constrained deployments due to their lower computational cost where inference latency and memory footprint are critical, particularly for edge-side track monitoring applications~\cite{galvez2022hybrid, haghbin2024cnn1d}. Finally, the majority of surveyed studies validate their approaches on proprietary or single-operator datasets, limiting conclusions about cross-fleet generalizability (a gap that remains open for future work)~\cite{davari2021predictive, zhang2024tcngru}.

\subsection{Reconstruction-Based Approaches}
\label{sec:reconstruction}

Reconstruction-based anomaly detection trains models exclusively on normal data to learn compact representations of healthy system behavior. At inference, inputs that deviate significantly from learned normal patterns produce elevated reconstruction errors, which serve as anomaly scores~\cite{berahmand2024autoencoders, chalapathy2019deep}. Anomalies are flagged when the reconstruction error $e = \lVert x - \hat{x} \rVert$ exceeds a static or adaptive threshold. This paradigm requires no fault labels, making it particularly suited to railway contexts where annotated fault data are scarce or unavailable~\cite{pang2021deep}.

\begin{figure}[h]
\centering
\includegraphics[width=\linewidth]
{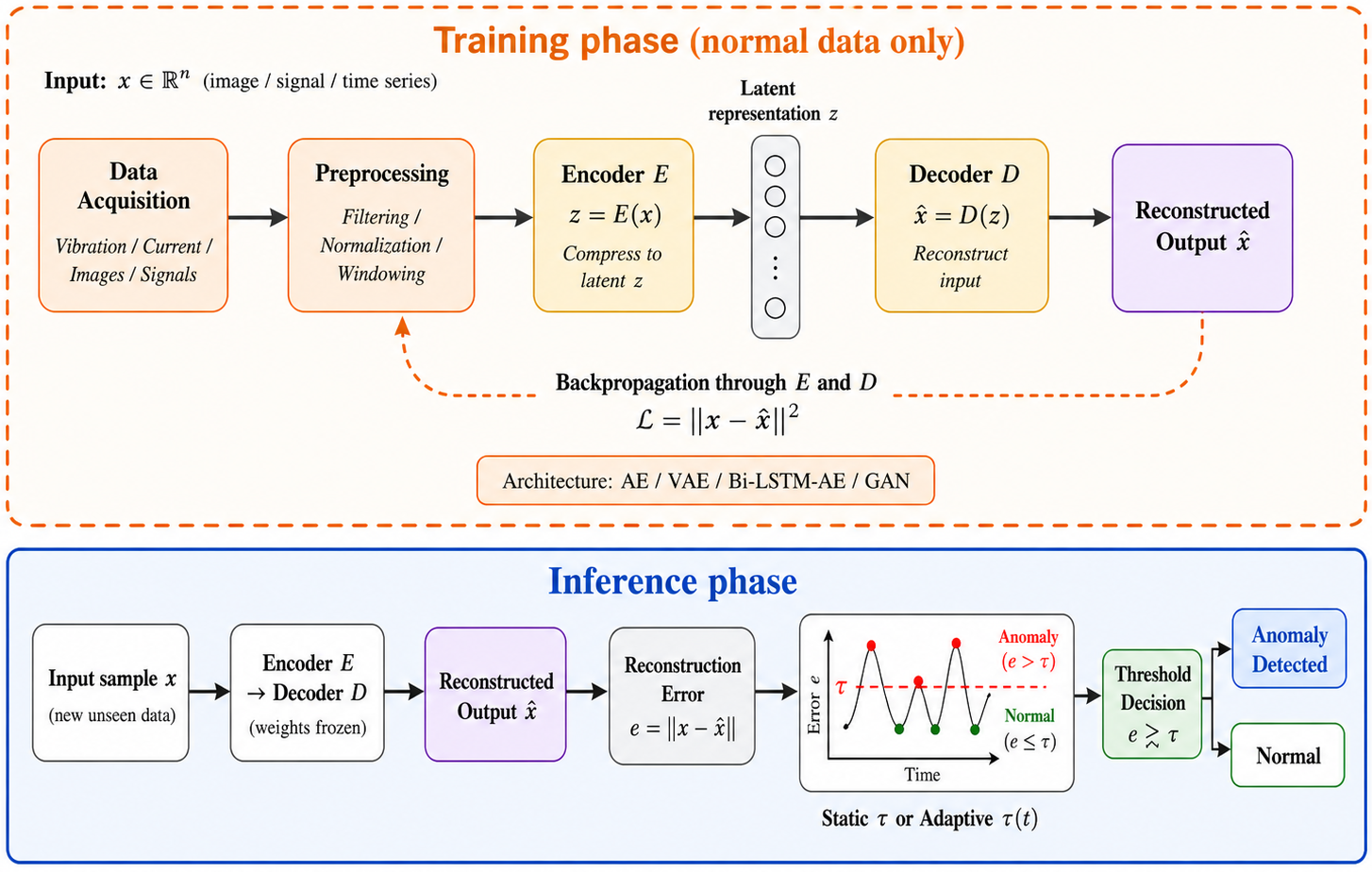}
\caption{Typical pipeline for 
reconstruction-based anomaly detection 
in railway systems.}
\label{fig:Reconstruction-pipeline}
\end{figure}

As illustrated in Figure~\ref{fig:Reconstruction-pipeline}, the pipeline consists of an encoder that compresses the input $x \in \mathbb{R}^n$ into a latent representation $z$, a decoder that reconstructs $\hat{x} = D(E(x))$, and an anomaly scoring module that compares the input and its reconstruction. The model is trained exclusively on normal data by minimizing the reconstruction loss $\mathcal{L} = \lVert x - \hat{x} \rVert^2$.

In this survey, reconstruction-based railway studies are organized into two groups according to their core architectural family: standard and recurrent autoencoders, and generative adversarial models. For each group, a summary table lists the surveyed works.

\subsubsection{Standard and Recurrent Autoencoders}

Standard Autoencoders (AEs) and their recurrent variants constitute the dominant family for reconstruction-based anomaly detection in railway systems. These architectures compress input data into a bottleneck latent space and reconstruct it, flagging anomalies through elevated reconstruction error on unseen inputs.

Sparse Autoencoders (SAEs) extend standard AEs by regularizing hidden activations to promote sparse and interpretable latent representations~\cite{chang2019sparse}. Davari et al.~\cite{davari2021predictive} applied separate SAE models to analog and digital sensor streamsfrom metro train Air Production Units (APUs), demonstrating that reconstruction-based detection can identify operational anomalies in heterogeneous sensor environments without fault labels. A subsequent study on the same MetroPT-3 dataset~\cite{davari2023lstm_ae} compared Long Short-Term Memory Autoencoders (LSTM-AEs) with SAEs, finding that LSTM-AEs achieve superior F1-score and recall by capturing temporal dependencies in pressure and current signals. Kang et al.~\cite{kang2021lstm_ae} applied a one-class LSTM-AE to detect anomalies in the brake operating unit of metro vehicles, training exclusively on healthy current signals and flagging deviations with a reconstruction error threshold.

Bidirectional LSTM Autoencoders (Bi-LSTM-AEs) extend recurrent AEs by processing sequences in both forward and backward directions, improving sensitivity to subtle temporal anomalies. Correia et al.~\cite{correia2024online} extended LSTM-AE-based reconstruction to online anomaly detection in multivariate railway sensor streams, demonstrating that reconstruction errors evaluated with the Mahalanobis distance improve detection robustness compared to mean squared error thresholding. Seo et al.~\cite{seo2024earthquake} applied autoencoder-based reconstruction to vibration signals from Korean high-speed trains for unsupervised earthquake anomaly detection, training only on normal operational vibration data and outperforming the conventional Short-Time Average over Long-Time Average detector for peak ground acceleration events above the derailment threshold.

Variational Autoencoders (VAEs) extend standard AEs by learning a probabilistic latent distribution $q(z|x)$ rather than a fixed encoding. The decoder samples from this distribution to reconstruct the input, and anomalies are detected when the reconstruction likelihood or the latent deviation from the learned normal distribution is low~\cite{berahmand2024autoencoders, zamanzadeh2024deep}.

Yuan et al.~\cite{yuan2021unsupervised} applied a Convolutional VAE (CVAE) to detect squat defects in rail surfaces from Axle Box Acceleration (ABA) signals, using an Elliptic Envelope detector on latent features to identify deviations from the normal distribution. The model successfully identified defects as shallow as 0.03 mm without any fault labels, demonstrating the sensitivity of probabilistic reconstruction to subtle infrastructure anomalies~(Table~\ref{tab:rec_a}).

\begin{table}[htbp]
\centering
\caption{Standard and recurrent autoencoder 
approaches for railway anomaly detection. }
\label{tab:rec_a}
\renewcommand{\arraystretch}{1.25}
\scriptsize
\setlength{\tabcolsep}{3pt}
\begin{tabular}{@{}|p{0.55cm}|p{0.5cm}|
p{1.5cm}|p{1.8cm}|p{1.6cm}|
p{4.8cm}|@{}}
\cline{1-6}
\textbf{Ref.} & \textbf{Year} & 
\textbf{Subsystem} & 
\textbf{Architecture} & 
\textbf{Modality} & 
\textbf{AD Task} \\
\cline{1-6}

\cite{davari2021predictive} & 2021 &
Air production unit &
SAE &
Pressure + current &
Detect anomalies in APU analog and digital sensor streams without fault labels. \\
\cline{1-6}

\cite{davari2023lstm_ae} & 2024 &
Air production unit &
LSTM-AE vs.\ SAE &
Pressure + current &
Detect APU anomalies using temporal reconstruction on MetroPT-3 dataset. \\
\cline{1-6}

\cite{kang2021lstm_ae} & 2021 &
Braking system &
One-class LSTM-AE &
Current signals &
Detect anomalies in metro brake operating unit from current signal reconstruction. \\
\cline{1-6}

\cite{correia2024online} & 2024 &
Rolling stock &
LSTM-AE &
Multivariate signals &
Online reconstruction-based anomaly detection in multivariate railway sensor streams. \\
\cline{1-6}

\cite{seo2024earthquake} & 2024 &
High-speed rail &
Convolutional AE &
Vibration signals &
Detect seismic anomalies from train vibration using normal-only autoencoder reconstruction. \\
\cline{1-6}
\cite{yuan2021unsupervised} & 2021 &
Rail surface &
CVAE &
ABA signals &
Detect rail squat defects as shallow as 0.03 mm from normal ABA signal reconstruction. \\
\cline{1-6}
\end{tabular}
\end{table}

\subsubsection{Generative Adversarial Reconstruction Models}

Generative Adversarial Networks (GANs) for anomaly detection train a generator $G$ to reconstruct normal inputs and a discriminator $D$ to distinguish real from generated samples. Anomalies are flagged when reconstruction errors (either pixel-wise or perceptual)exceed a threshold. This family offers superior expressiveness for high-dimensional visual data but requires careful training stabilization~\cite{di2019survey}.

Wang et al.~\cite{wang2022deep} proposed a GAN-based framework to detect foreign objects on railway tracks using only normal track images, reconstructing clean scenes and flagging deviations via error maps. Wang et al.~\cite{wang2024undercarriage} applied an adversarial memory-enhanced reconstruction network to metro vehicle undercarriage images, combining a memory module with adversarial training to prevent the model from reconstructing anomalous patterns (a known failure mode of standard GANs in reconstruction)~\cite{chalapathy2019deep}. 

For visual track inspection, transformer-based reconstruction architectures have recently emerged as an alternative to convolutional autoencoders. Wang et al.~\cite{wang2023memformer} proposed MemFormer, a memory-augmented Convolutional Neural Network (CNN) and Transformer model that reconstructs normal metro track images and flags anomalies via reconstruction error. The memory module constrains the model to reconstruct only normal patterns, improving robustness to the anomaly reconstruction problem. A similar approach applied to railway turnout inspection, Rail-PatchCore~\cite{railpatchcore2025}, combines unsupervised feature-level reconstruction with a dual-dimensional channel attention module to reduce background noise interference in complex turnout environments. Wang et al.~\cite{wang2024anomaly} introduced an anomaly-free representation learning approach for railway foreign object detection that learns discriminative features exclusively from normal track images. For pantograph inspection, an unsupervised reverse distillation network~\cite{padn2024pantograph} reconstructs normal pantograph image features through an encoder-decoder architecture trained on defect-free images, achieving image-level and pixel-level reconstruction scores competitive with supervised baselines~(Table~\ref{tab:rec_c}).

\begin{table*}[htbp]
\centering
\caption{Generative adversarial and 
transformer reconstruction models for 
railway anomaly detection. 
All methods are trained on normal data only.
GAN: Generative Adversarial Network; 
CNN: Convolutional Neural Network.}
\label{tab:rec_c}
\renewcommand{\arraystretch}{1.25}
\scriptsize
\setlength{\tabcolsep}{3pt}
\begin{tabular}{@{}|p{0.55cm}|p{0.5cm}|
p{1.5cm}|p{1.8cm}|p{1.6cm}|
p{4.8cm}|@{}}
\cline{1-6}
\textbf{Ref.} & \textbf{Year} & 
\textbf{Subsystem} & 
\textbf{Architecture} & 
\textbf{Modality} & 
\textbf{AD Task} \\
\cline{1-6}

\cite{wang2022deep} & 2022 &
Track obstacles &
GAN (adversarial AE) &
RGB images &
Detect foreign objects on tracks by reconstructing normal track scenes from onboard cameras. \\
\cline{1-6}

\cite{wang2024undercarriage} & 2024 &
Metro undercarriage &
GAN + memory module &
RGB images &
Detect undercarriage anomalies via adversarial memory-enhanced image reconstruction. \\
\cline{1-6}

\cite{wang2023memformer} & 2024 &
Metro tracks &
CNN + Transformer &
RGB images &
Detect track surface anomalies by reconstructing normal track images with memory augmentation. \\
\cline{1-6}

\cite{railpatchcore2025} & 2025 &
Railway turnout &
PatchCore + attention &
RGB images &
Detect visual anomalies in turnout environments using unsupervised feature reconstruction. \\
\cline{1-6}

\cite{padn2024pantograph} & 2024 &
Pantograph &
Reverse distillation network &
RGB images &
Detect pantograph anomalies via unsupervised reverse distillation image reconstruction. \\
\cline{1-6}

\end{tabular}
\end{table*}

\vspace{1ex}
\noindent\textbf{Summary and limitations:}
Reconstruction-based approaches offer label-free anomaly detection across diverse railway modalities (signals, images, and vibrations)and generalize to fault types not seen during training~\cite{pang2021deep, chalapathy2019deep}. However, three limitations constrain their applicability in real railway environments. First, standard autoencoders may reconstruct anomalous inputs with low error if the anomaly shares low-level features with normal data (a phenomenon known as the anomaly reconstruction problem)~\cite{chalapathy2019deep}. Second, reconstruction thresholds are typically calibrated heuristically and may not generalize across different operational conditions or fleets~\cite{correia2024online}. Third, generative models such as GANs require careful training stabilization and are sensitive to hyperparameter choices~\cite{di2019survey}.

\vspace{1ex}
\noindent\textbf{Cross-category observations:}
Across the three groups summarized in Tables~\ref{tab:rec_a}-\ref{tab:rec_c}, recurrent autoencoders(particularly LSTM-AE and Bi-LSTM-AE variants)dominate the surveyed railway literature for signal-based monitoring, reflecting their ability to capture temporal dependencies in multivariate sensor streams~\cite{ kang2021lstm_ae, correia2024online}. For visual inspection tasks, transformer and memory-augmented reconstruction models are emerging as alternatives to convolutional autoencoders, demonstrating improved robustness against the anomaly reconstruction problem~\cite{wang2023memformer,railpatchcore2025}. Probabilistic models such as VAEs remain less explored in the railway domain despite their demonstrated sensitivity to subtle structural anomalies~\cite{yuan2021unsupervised}. Finally, as with prediction-based approaches, the majority of surveyed reconstruction studies are validated on single-operator or proprietary datasets, limiting conclusions about cross-fleet generalizability~\cite{davari2021predictive, wang2024undercarriage}.

\subsection{Hybrid Approaches}
\label{sec:hybrid}

None of the three paradigms reviewed in the preceding sections - classification, reconstruction, and prediction - is universally sufficient for railway anomaly detection. Classification-based models achieve high accuracy on known fault types but fail on unseen anomalies. Reconstruction-based models detect novel deviations without labels but produce elevated false alarm rates in noisy environments. Prediction-based models generalize well to temporal anomalies but require stable operating conditions. Hybrid approaches address these limitations by combining two or more paradigms within a unified framework, exploiting their complementary strengths~\cite{zamanzadeh2024deep, chalapathy2019deep}.

\begin{figure}[h]
\centering
\includegraphics[width=\linewidth]
{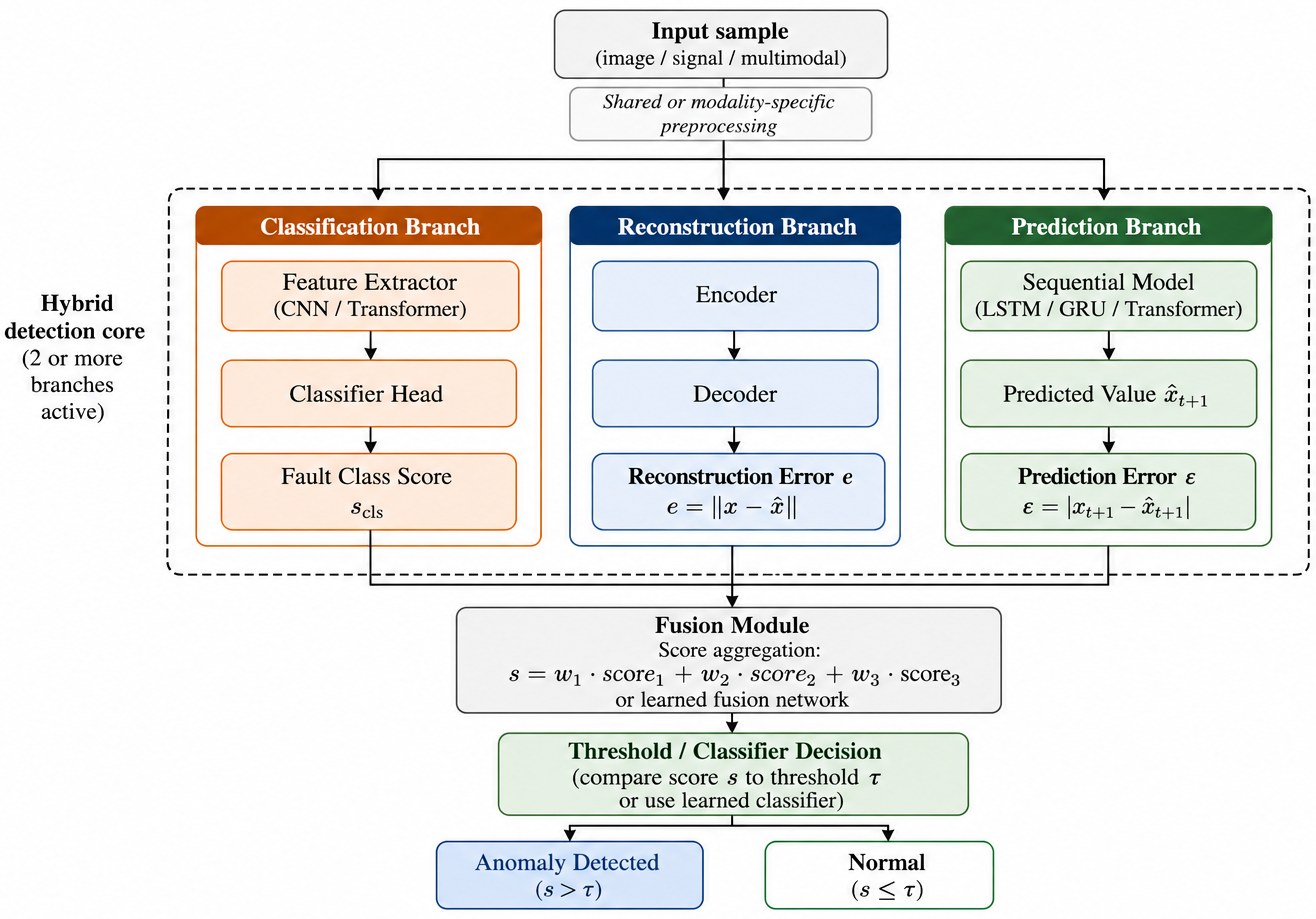}
\caption{Typical pipeline for hybrid 
anomaly detection in railway systems. Two or more detection paradigms process the input in parallel. Their outputs are fused into a unified anomaly score for the final decision.}
\label{fig:Hybrid-pipeline}
\end{figure}

As illustrated in Figure~\ref{fig:Hybrid-pipeline}, hybrid frameworks process the input through parallel or sequential branches (each implementing a different detection principle) and fuse their outputs into a unified anomaly score or decision. Fusion can be rule-based, score-weighted, or learned end-to-end depending on the availability of labeled data and the complexity of the target scenario.

In the surveyed railway literature, hybrid approaches appear in two main forms. The first form combines reconstruction and prediction within a single generative model, jointly learning to reconstruct the current state and predict future states, with anomalies scored as combined deviations from both objectives. The second form integrates supervised detection with unsupervised reconstruction or rule-based reasoning, enabling simultaneous identification of known and unknown anomalies within the same pipeline~(Table~\ref{tab:h}).

Laurendin et al.~\cite{laurendin2021hazardous} proposed a hybrid framework for detecting hazardous events near train doors, combining reconstruction and prediction within a conditional Generative Adversarial Network (cGAN) with a U-Net generator. The model jointly reconstructs the current video frame and predicts optical flow, scoring anomalies as deviations in both image appearance and motion patterns. This dual-objective design enables detection of obstructed doors, irregular passenger movements, and mechanical faults that would be missed by either objective alone.

Gao et al.~\cite{gao2020automatic} presented a multimodal hybrid system for pantograph-catenary inspection that integrates classification, reconstruction, and prediction across four sensing modalities: Red Green Blue (RGB) imaging, infrared thermography, ultraviolet discharge detection, and vibration signals. Each modality is processed by a specialized module (object detectors for known defect localization, autoencoders for signal deviation detection, and time-series models for behavioral forecasting) with outputs interpreted jointly. This modular architecture demonstrates that combining complementary detection principles and sensing modalities can achieve broader anomaly coverage than any single approach.

Chen et al.~\cite{chen2022meta} proposed a meta-learning framework combining Generative Adversarial Networks with structural similarity-based loss functions for anomaly detection in high-speed rail inspection. The framework learns to detect anomalies from limited labeled examples by combining generative reconstruction with a supervised meta-learner, reducing inspection workload by over 99.7\% when deployed on five high-speed railway lines in China.

Sun et al.~\cite{railfdnet2025} introduced RailFDNet, a hybrid supervision model for railway anomalous object detection that combines an unsupervised normalizing flow-based backbone with a supervised transformer-based anomaly decoder. The unsupervised component enhances the feature discrepancy between normal and abnormal data, while the supervised decoder uses artificially generated anomaly images to improve precise localization. This dual-stage training strategy enables the model to handle both the absence of real anomaly labels and the need for accurate spatial localization in open railway environments.

Wu et al.~\cite{wu2023railtrack} proposed a hybrid deep learning framework for automatic railroad track component inspection that combines convolutional feature extraction with rule-based geometric reasoning. The supervised classification component identifies track component types, while the rule-based module applies domain-specific geometric constraints to flag anomalous spatial configurations that the neural network alone cannot capture.

\begin{table}[htbp]
\centering
\caption{Hybrid approaches for railway 
anomaly detection. 
IR: Infrared; UV: Ultraviolet.}
\label{tab:h}
\renewcommand{\arraystretch}{1.25}
\scriptsize
\setlength{\tabcolsep}{3pt}
\begin{tabular}{@{}|p{0.55cm}|p{0.5cm}|
p{1.4cm}|p{2.4cm}|p{1.5cm}|
p{4.4cm}|@{}}
\cline{1-6}
\textbf{Ref.} & \textbf{Year} & 
\textbf{Subsystem} & 
\textbf{Architecture} & 
\textbf{Modality} & 
\textbf{AD Task} \\
\cline{1-6}

\cite{laurendin2021hazardous} & 2021 &
Train doors &
Conditional GAN + U-Net 
(reconstruction + prediction) &
Video &
Detect hazardous door events by jointly reconstructing frames and predicting optical flow. \\
\cline{1-6}

\cite{gao2020automatic} & 2020 &
Pantograph-catenary &
YOLO + AE + LSTM 
(classification + reconstruction 
+ prediction) &
RGB + IR + UV + vibration &
Detect pantograph anomalies across four modalities using parallel specialized detection modules. \\
\cline{1-6}

\cite{chen2022meta} & 2022 &
Track infrastructure &
GAN + meta-learner 
(reconstruction + supervised) &
RGB images &
Detect track anomalies from limited labeled examples using generative reconstruction and meta-learning. \\
\cline{1-6}

\cite{railfdnet2025} & 2025 &
Track obstacles &
Normalizing flow + Transformer 
(unsupervised + supervised) &
RGB images &
Detect and localize anomalous objects using hybrid unsupervised backbone and supervised decoder. \\
\cline{1-6}

\cite{wu2023railtrack} & 2023 &
Track components &
CNN + rule-based module 
(classification + rules) &
RGB images &
Inspect track components by combining learned classification with geometric domain rules. \\
\cline{1-6}

\end{tabular}
\end{table}

\vspace{1ex}
\noindent\textbf{Summary and limitations:}
Hybrid approaches offer broader anomaly coverage, greater robustness to label scarcity, and improved generalization compared to single-paradigm methods~\cite{pang2021deep, zamanzadeh2024deep}. They are particularly suited to safety-critical railway scenarios where both known and unknown fault types must be handled simultaneously. However, three limitations constrain their deployment. First, multi-branch architectures are more complex to design, train, and calibrate than single-paradigm models, particularly under real-time inference constraints~\cite{gao2020automatic}. Second, fusion strategies (whether rule-based or learned) require careful design to avoid one branch dominating the anomaly score~\cite{laurendin2021hazardous}. Third, hybrid models that process multiple modalities simultaneously impose higher computational and memory costs, which may limit their applicability on resource-constrained embedded platforms~\cite{chalapathy2019deep}.

\noindent
\textbf{General Interpretation and Comparative Analysis:}

The surveyed approaches reveal complementary strengths across four paradigms (classification, prediction, reconstruction, and hybrid) each suited to specific railway contexts.

Classification-based methods excel when labeled datasets are available and fault types are well-defined. CNN-based models achieve high accuracy in visual inspection tasks such as fastener detection~\cite{gao2023railway} and pantograph monitoring~\cite{gao2020automatic}, but their reliance on supervised learning limits adaptability to unseen anomalies.

Prediction-based methods (e.g., LSTM, GRU) are effective for continuous monitoring of multivariate time series, enabling early fault detection without anomaly labels~\cite{wang2022improved}. They capture temporal dynamics but require retraining under non-stationary conditions.

Reconstruction-based methods (AEs, VAEs, GANs) address scenarios with scarce or unlabeled fault data by modeling normal behavior and detecting deviations~\cite{davari2021predictive, yuan2021unsupervised, wang2022deep}. While robust for rare anomalies, defining interpretable thresholds remains challenging.

Hybrid approaches combine multiple paradigms for complex, multimodal environments. Examples include appearance-motion fusion for door hazard detection~\cite{laurendin2021hazardous} and integrated classification-prediction pipelines for pantograph monitoring~\cite{gao2020automatic}, offering flexibility at the cost of higher computational complexity.

In practice, deployment strategies often mix paradigms: classification for frequent visual defects, prediction for real-time sensor forecasting, reconstruction for rare failures, and hybrid models for safety-critical multimodal monitoring.

\section{Performance Evaluation for 
Railway Anomaly Detection}
\label{sec:4}

Selecting the right evaluation metric is as important as selecting the right model. In railway anomaly detection, faults are rare, error costs are asymmetric, and models must often run on embedded hardware under strict latency constraints. Standard metrics designed for balanced classification tasks frequently misrepresent performance in this context. This section reviews which metrics are appropriate for each detection task, identifies the limitations of common reporting practices observed across the 68 primary studies of this survey, and proposes concrete evaluation criteria better aligned with railway safety and deployment requirements~\cite{pang2021deep, chalapathy2019deep}.

\subsection{Metrics by Detection Task}

The appropriate metric depends on the model output and the detection objective. Table~\ref{tab:ad_metrics_summary} maps each metric to its detection task, explains what it measures, and highlights when it is appropriate or insufficient in railway contexts.

\begin{table}[H]
\centering
\caption{Evaluation metrics for railway 
anomaly detection.}
\label{tab:ad_metrics_summary}
\scriptsize
\renewcommand{\arraystretch}{1.3}
\begin{tabular}{@{}|p{1.7cm}|p{1.3cm}|
p{3.5cm}|p{4cm}|@{}}
\cline{1-4}
\textbf{Task} & 
\textbf{Metric} & 
\textbf{What it measures} & 
\textbf{Use in railway / Limitation} \\
\cline{1-4}

\textbf{Classification} &
Accuracy & 
Overall fraction of correct 
predictions. & 
\textbf{Avoid:} unreliable when faults represent $<$1\% of data (a model predicting "normal" for every sample achieves 99\% accuracy.) \\
\cline{2-4}
&
Recall & 
Fraction of actual faults 
correctly detected. & 
\textbf{Prioritize} in safety-critical scenarios (braking): a missed fault is more dangerous than a false alarm~\cite{shahin2023using}. \\
\cline{2-4}
&
Precision & 
Fraction of detected anomalies 
that are genuine faults. & 
\textbf{Use} when false alarms are operationally costly (unnecessary maintenance stops). \\
\cline{2-4}
&
F1-score & 
Harmonic mean of precision 
and recall. & 
\textbf{Preferred} for imbalanced binary detection. \\
\cline{2-4}
&
Specificity & 
Fraction of normal samples 
correctly identified. & 
\textbf{Use} to control false alarm rates in continuous monitoring; complements recall. \\

\cline{1-4}
\textbf{Anomaly scoring (unsupervised)} &
AUROC & 
Ranking quality of anomaly 
scores across all thresholds. & 
\textbf{Use} for threshold-independent comparison; \textbf{avoid} when anomaly prevalence is very low (PR-AUC is preferable)~\cite{vskvara2023auc}. \\
\cline{2-4}
&
PR-AUC & 
Area under the precision-recall curve; focuses on rare faults. & \textbf{Preferred} for reconstruction- and prediction-based approaches with severe class imbalance~\cite{pang2021deep}. \\
\cline{1-4}

\textbf{Forecasting / RUL} &
MAE & 
Mean absolute prediction error; 
robust to outliers. & 
\textbf{Use} for stable degradation signals (pressure, current, HVAC). \\
\cline{2-4}
&
RMSE & 
Root mean squared error; 
penalizes large deviations. & \textbf{Use} for detecting sudden spikes (temperature, vibration); sensitive to outliers. \\

\cline{1-4}
\textbf{Localization / Segmentation} &
mIoU & 
Intersection over union between 
predicted and ground-truth 
anomaly regions. & 
\textbf{Standard} for visual inspection tasks (cracks, surface defects). \\
\cline{2-4}
&
Pixel Accuracy & 
Fraction of correctly 
labeled pixels. & 
\textbf{Avoid} when the background dominates (biased toward non-anomalous regions). \\

\cline{1-4}
\end{tabular}
\end{table}

\subsection{Why Standard Metrics Are Insufficient for Railway Systems}

Three structural properties of railway anomaly detection make standard accuracy-oriented metrics unreliable.
 \begin{itemize}[noitemsep, topsep=2pt]
     \item \textbf{Class imbalance:}
     Operational fault rates in railway systems are typically below 1\%. Under such imbalance, accuracy and AUROC can both be misleading: a model that never detects any fault can report 99\% accuracy and a non-trivial AUROC~\cite{vskvara2023auc, pang2021deep}. PR-AUC is more informative because it focuses exclusively on the anomaly class and penalizes models that miss rare faults~\cite{malhotra2015long}.

\item\textbf{Asymmetric error costs:}
False negatives and false positives carry fundamentally different consequences. Missing a brake fault (FN) may cause a safety incident costing EUR~100{,}000 or more, while a false alarm (FP) typically triggers an unnecessary inspection costing a few hundred euros. Standard metrics treat both errors as equally costly, which is inappropriate for safety-critical monitoring~\cite{shahin2023using, ma2024cost}. Cost-sensitive evaluation assigns asymmetric penalties $c_{FN} \gg c_{FP}$ and reports expected operational cost:
\[
\bar{\mathcal{C}} = \frac{1}{N}
\bigl(c_{TP}\,TP + c_{FP}\,FP 
+ c_{FN}\,FN + c_{TN}\,TN\bigr)
\]
This formulation makes model selection directly accountable to operational and safety objectives~\cite{ma2024cost, bodo2021cost}.

\item\textbf{Deployment gap:}
High offline accuracy does not guarantee deployability. A model achieving 99\% F1-score on a GPU server may fail to meet a 100\,ms latency budget on an embedded platform. For axle temperature monitored at 10\,Hz, a 2\,s detection latency implies a delay of 20 samples (even 98\% AUROC cannot compensate for this operational gap)~\cite{benmeziane2021comprehensive, tsiktsiris2024multimodal}. Inference latency, memory footprint, and throughput must be reported alongside accuracy metrics for any deployment-oriented study.
 \end{itemize}

\subsection{Evaluation Practices in the 
Surveyed Literature}

Table~\ref{tab:metric_adoption} summarizes 
how evaluation metrics are used across 
the 68 primary studies included in this survey.

\begin{table}[htbp]
\centering
\caption{Metric adoption across the 68 surveyed studies. Inference latency is counted when at least one quantitative inference-time indicator is explicitly reported (execution time, frames per second, or end-to-end latency).}
\label{tab:metric_adoption}
\scriptsize
\renewcommand{\arraystretch}{1.25}
\begin{tabular}{@{}|p{3.8cm}|r|c|@{}}
\cline{1-3}
\textbf{Metric} & 
\textbf{Studies} & 
\textbf{\%} \\
\cline{1-3}
Accuracy & 42 & 61.8\% \\
Precision / Recall & 38 & 55.9\% \\
F1-score & 36 & 52.9\% \\
mAP / mIoU & 14 & 20.6\% \\
AUROC & 18 & 26.5\% \\
RMSE / MAE & 11 & 16.2\% \\
PR-AUC & 5 & 7.4\% \\
Inference latency & 7 & 10.3\% \\
Cost-sensitive evaluation & 2 & 2.9\% \\
Memory / energy footprint & 5 & 7.4\% \\
\cline{1-3}
\end{tabular}
\end{table}

Three key observations emerge from this analysis. First, accuracy remains the most frequently reported metric, despite being poorly suited to imbalanced railway fault detection scenarios. Second, PR-AUC, which is more appropriate for rare-event detection, is reported in fewer than 8\% of studies. Third, deployment-related metrics are largely underreported: inference latency is reported in only 7 studies, and memory or energy footprint in only 5 studies. This confirms that practical deployment considerations remain secondary in the current literature~\cite{benmeziane2021comprehensive}.

\noindent\textbf{Minimum reporting standard:}Based on these findings, future studies should 
report at minimum: 
(i)~F1-score and PR-AUC for imbalanced detection; 
(ii)~false negative rate for safety-critical scenarios; 
(iii)~inference latency measured on the target platform; and 
(iv)~cost-sensitive evaluation when asymmetric error costs 
can be estimated. These criteria are consistent with the deployment-oriented decision framework presented in Section~\ref{sec:decision_framework}.

\section{Deployment of Deep Learning Models in Railway Systems}
\label{sec:deployment}

Deploying deep learning models in railway systems requires balancing predictive performance with strict operational constraints, including latency, memory footprint, energy consumption, and safety compliance~\cite{kljaic2023overview}. Railway systems operate on resource-constrained embedded platforms under harsh conditions (vibration, thermal stress, electromagnetic interference) where long-term reliability is mandatory~\cite{radosavljevic2025railway}. As a result, model design cannot be decoupled from hardware and deployment constraints.

Railway applications must comply with safety standards such as EN~50155 (environmental requirements for onboard electronics) and EN~50128 (software safety integrity levels SIL1--SIL4)~\cite{donnarumma201950128}. However, no standardized certification pathway currently exists for deep learning models, due to their non-deterministic behavior and limited interpretability~\cite{radosavljevic2025railway}. This lack of certification remains a major barrier to deploying neural networks in safety-critical railway subsystems.

Despite the importance of these constraints, deployment remains the least documented aspect of the railway anomaly detection literature. While recent work has begun to address this gap for specific subsystems such as track defect detection~\cite{radosavljevic2025railway}, no prior survey provides a deployment-oriented analysis spanning infrastructure, rolling stock, and environmental anomaly detection across heterogeneous hardware platforms. Among the studies surveyed in this work, few explicitly report at least one quantitative deployment metric such as inference latency, power consumption, or accuracy degradation under model compression. The following analysis combines this limited railway-specific evidence with established practices from the broader embedded deep learning literature, adapted to railway constraints. Each subsection distinguishes between findings from surveyed railway studies, requirements from railway engineering standards, and guidance from the general embedded AI literature.

\subsection{Deployment Evidence in the Surveyed Literature}
\label{sec:deploy_evidence}

To assess the current state of deployment in railway anomaly detection, we examined all surveyed studies for quantitative deployment metrics. Table~\ref{tab:deploy_metrics} summarizes the studies that report at least one hardware-level measurement. The remaining studies evaluate their models exclusively on desktop or server-grade GPUs without reporting inference latency on a target embedded platform, power consumption, or accuracy impact of model compression.

\begin{table*}[htpb]
\centering
\caption{Deployment metrics reported across the surveyed studies. Memory values refer to platform RAM capacity (GB) or FPGA on-chip resource usage (BRAM), depending on reporting conventions in each study. NR: not reported.}
\label{tab:deploy_metrics}
\renewcommand{\arraystretch}{1.25}
\scriptsize
\setlength{\tabcolsep}{3pt}
\begin{tabular}{|p{0.6cm}|p{0.7cm}|p{1.6cm}|p{1.7cm}|p{1cm}|p{1cm}|p{0.9cm}|p{1.2cm}|p{1.4cm}|}
\hline
\textbf{Ref.} & \textbf{Year} & \textbf{Sub-system} & \textbf{Platform} & \textbf{Class} & \textbf{Latency (ms)} & \textbf{Power (W)} & \textbf{Accuracy} & \textbf{Memory} \\
\hline
\cite{fu2025efficient} & 2025 & Track faults (vision) & Zynq 7Z020 / KV260 & FPGA & 0.7--10 & 2.3--3.9 & 93.4\% & 4.5--11.6\,Mb BRAM \\
\hline
\cite{liu2020real} & 2020 & Traction faults & Xilinx VCU128 & FPGA & $<$1 & NR & $>$93\% & FPGA resources \\
\hline
\cite{li2024edge} & 2024 & Track faults (vision) & Xilinx ZCU104 & FPGA & 2.1 & 6.9 & 88.9\% & 230.5 BRAM \\
\hline
\cite{cui2019real} & 2020 & Fastener (3D) & Jetson TX2 + Atom & GPU + CPU & $\sim$15 & NR & 99.7\% & 8\,GB \\
\hline
\cite{tsiktsiris2024multimodal} & 2024 & In-cabin events & Jetson AGX Xavier & GPU edge & $\sim$1200 & 10--30 & 85.1\% & 32\,GB \\
\hline
\cite{diaz2025real} & 2025 & Passenger counting & Jetson Nano & GPU edge & 80--100 & $\sim$5--10 & 96.85\% & 4\,GB \\
\hline
\cite{chavhan2025bi} & 2025 & Track faults (IoT) & Jetson Xavier + ESP32 & GPU + IoT & 105--160 & NR & 90--96\% & $\sim$6\,GB \\
\hline
\end{tabular}
\end{table*}

Four observations emerge from this table.

\textbf{Platform diversity is narrow:} Three studies deploy on Field-Programmable Gate Array (FPGA) platforms~\cite{fu2025efficient, liu2020real, li2024edge}, three use NVIDIA Jetson-family GPU edge devices~\cite{cui2019real, tsiktsiris2024multimodal, diaz2025real}, and one combines a Jetson platform with IoT modules~\cite{chavhan2025bi}.

\textbf{Latency spans three orders of magnitude:} Inference times range from 0.7\,ms for binary classification on an FPGA~\cite{fu2025efficient} to approximately 1200\,ms for a full multimodal pipeline on a Jetson AGX Xavier~\cite{tsiktsiris2024multimodal}. This range reflects differences in task complexity and input resolution rather than platform capability alone, which underlines the need for standardized deployment benchmarks in the railway domain.

\textbf{Power consumption is rarely measured:} Only three studies report power measurements~\cite{fu2025efficient, li2024edge, tsiktsiris2024multimodal}. FPGA-based systems operate at 2.3--6.9\,W~\cite{fu2025efficient, li2024edge}, an order of magnitude below the 10--30\,W reported for GPU edge platforms~\cite{tsiktsiris2024multimodal}. Fu et al.~\cite{fu2025efficient} introduce a cost-energy efficiency metric (CEES\,=\,throughput\,/\,power\,$\times$\,cost), demonstrating 101.9$\times$ and 33.8$\times$ improvements over CPU and GPU platforms respectively. The remaining studies do not report power consumption, making it impossible to assess their suitability for onboard deployment where power budgets are typically constrained to 500\,W or less for the full monitoring system~\cite{radosavljevic2025railway}.

\textbf{Model optimization is inconsistent:} Model compression techniques (reviewed in detail in Section~\ref{sec:model_optimization}) are applied unevenly. Fu et al.~\cite{fu2025efficient} provide the most complete pipeline, combining extreme quantization with custom hardware resource balancing. Li et al.~\cite{li2024edge} apply reduced-precision arithmetic with no accuracy loss. Tsiktsiris et al.~\cite{tsiktsiris2024multimodal} use hardware-accelerated inference optimization with approximately 0.5\% accuracy drop. The remaining studies apply minimal or no optimization.

This analysis confirms that real-time inference on embedded platforms is technically feasible for specific railway tasks. However, the evidence base remains too sparse to support general conclusions about deployability across subsystems or operational conditions. Establishing deployment reporting as a minimum standard (including inference latency, power consumption, and accuracy impact of compression) would significantly improve the comparability of future studies.

\subsection{Hardware Platforms for Railway Deployment}
\label{sec:deploy_platforms}

Based on the surveyed literature and the broader embedded AI domain, embedded platforms for railway anomaly detection fall into five classes.
\begin{itemize}[noitemsep, topsep=2pt]
    \item \textbf{General-purpose CPU/GPU systems} 
are primarily used for training and offline evaluation. They provide high computational throughput and large memory capacity. However, their high power consumption, non-deterministic latency, and large physical footprint make them unsuitable for onboard railway deployment~\cite{gao2023railway, kaya2025detection, edgeai2025survey}.

\item\textbf{GPU-accelerated edge platforms}, 
such as the NVIDIA Jetson family (TX2, Xavier NX, Orin), offer a practical compromise between performance and deployability~\cite{tsiktsiris2024multimodal, cui2019real}. Their software ecosystem supports rapid deployment, although latency can vary under concurrent workloads~\cite{edgeai2025survey}.

\item\textbf{Field-Programmable Gate Array (FPGA)-based systems} provide deterministic execution and ultra-low latency, making them well-suited for safety-critical subsystems such as braking monitoring~\cite{liu2020real, nechi2023fpga, li2024edge}. Their energy efficiency is high due to custom dataflow and fixed-point computation, at the cost of high development complexity and limited flexibility.

\item\textbf{Embedded Central Processing Unit platforms}, including ARM Cortex-A and Intel Atom processors, are used for preprocessing, lightweight inference, and Train Control and Monitoring System (TCMS) integration. Their limited parallelism restricts them to lightweight models or low-frequency monitoring tasks~\cite{siddiqui2022iot, wang2022improved, benmeziane2021comprehensive}.

\item\textbf{Neural Processing Units (NPUs) and AI accelerators} (e.g., ARM Ethos, Hailo-8) are purpose-built for neural network inference and achieve high throughput at low power consumption through architectures optimized for matrix operations and quantized computation~\cite{edgeai2025survey}. Although currently underrepresented in the railway anomaly detection literature, NPUs represent a strong candidate for future onboard deployment due to their energy efficiency and compact form factor.

\end{itemize}
Table~\ref{tab:platforms} summarizes their trade-offs and typical use cases.

\begin{table*}[htpb]
\centering
\caption{Hardware platforms for deep learning deployment in railway anomaly detection systems. The \textbf{Railway evidence} column indicates whether the platform has been validated in a surveyed railway study (Yes) or is projected from the general embedded AI literature (Projected).}
\label{tab:platforms}
\renewcommand{\arraystretch}{1.25}
\scriptsize
\setlength{\tabcolsep}{3pt}
\begin{tabular}{|p{1.3cm}|p{1.1cm}|p{3.5cm}|p{3.8cm}|p{1.3cm}|}
\hline
\textbf{Platform} & \textbf{Devices} & \textbf{When to Choose} & \textbf{Key Limitation} & \textbf{Railway evidence} \\
\hline
CPU/GPU server & Intel Xeon, RTX & Offline training with high compute requirements, no onboard constraints~\cite{gao2023railway, kaya2025detection}. & High power ($>$200\,W); non-deterministic latency incompatible with real-time safety-critical systems~\cite{benmeziane2021comprehensive, edgeai2025survey}. & Yes (training only) \\
\hline
GPU edge & Jetson TX2, Xavier, Orin & Real-time vision and multimodal tasks under moderate power constraints~\cite{tsiktsiris2024multimodal, cui2019real}. &
Latency variability under concurrent workloads; limited for real-time constraints~\cite{edgeai2025survey}. & Yes \\
\hline
FPGA & Xilinx Zynq, ZCU, VCU & Safety-critical subsystems requiring deterministic and low latency~\cite{liu2020real, nechi2023fpga}. &
High development complexity; long design cycles; limited flexibility for model updates~\cite{nechi2023fpga}. & Yes \\
\hline
Embedded CPU & ARM Cortex-A, Intel Atom & Preprocessing, lightweight inference~\cite{siddiqui2022iot, wang2022improved}. &
Low parallelism; unsuitable for complex deep learning models or high-frequency inference~\cite{benmeziane2021comprehensive}. & Partial \\
\hline
NPU / AI accel. & Hailo-8, ARM Ethos & Low-power onboard inference with dedicated neural acceleration; strong candidate for future embedded deployment~\cite{edgeai2025survey}. &
Limited validation in railway applications; integration and certification challenges remain open. & Projected \\
\hline
\end{tabular}
\end{table*}

\subsection{Deployment Architectures: Edge, Cloud, and Hybrid}
\label{sec:deploy_arch}

Deployment architecture directly impacts end-to-end latency, availability under degraded connectivity, and integration with maintenance workflows~\cite{gong2024train, armijo2024integration, amin2025integrating}.

\begin{itemize}[noitemsep, topsep=2pt]
    \item \textbf{Edge deployment} executes inference close to sensors and is suited to anomalies requiring predictable response times or local autonomy. European railway architectures emphasize edge and wayside processing for continuous detection~\cite{gong2024train}. Edge offers low latency and resilience to connectivity outages, but is constrained by compute, memory, and power budgets. It is best suited for safety-critical events such as track intrusion and door obstruction.

\item\textbf{Cloud deployment} centralizes storage and compute for fleet-level analytics, long-horizon trend analysis, and model retraining~\cite{amin2025integrating}. It provides elastic compute and cross-fleet aggregation, but network-induced latency variability makes it unsuitable for real-time detection. It is best suited for slow degradation analytics and retrospective root-cause analysis.

\item\textbf{Hybrid edge--cloud deployment} combines fast edge screening with centralized refinement: edge nodes triage events and forward only informative content, while centralized services correlate anomalies across assets and update models~\cite{armijo2024integration}. This architecture balances responsiveness and scalability at the cost of higher orchestration complexity. It is best suited for multimodal, network-wide deployments requiring both immediate alerts and fleet-level context.
\end{itemize}

In practice, deployment selection is driven by latency determinism and connectivity constraints: safety-critical tasks with strict deadlines require edge deployment, while non-critical analytics can tolerate cloud-based processing. Hybrid architectures are increasingly necessary for multimodal railway monitoring systems that must simultaneously provide immediate local alerts and support fleet-level decision making.

\subsection{Real-Time and Non-Real-Time Anomalies}
\label{sec:realtime}

Railway anomalies differ in urgency, which directly influences hardware selection and algorithm design. In this context, real-time refers to predictable (deterministic) response time rather than simply low average latency~\cite{kljaic2023overview}.
\begin{itemize}[noitemsep, topsep=2pt]
    \item \textbf{Real-time anomalies} are safety-critical events (sudden motor failures, braking faults, track intrusions) requiring bounded inference latency. These are further divided into \textbf{hard real-time} (e.g., braking protection, where deadline misses are unacceptable) and \textbf{soft real-time} (e.g., door monitoring, where occasional misses degrade performance without immediate safety impact). Hard real-time constraints favor deterministic platforms such as FPGAs~\cite{liu2020real, fu2025efficient}, while soft real-time tasks are typically handled by GPU edge devices~\cite{tsiktsiris2024multimodal, diaz2025real}. For video-based inspection at operational speeds below 160\,km/h, a frame rate of approximately 30\,fps at 1504-pixel resolution is required for continuous coverage~\cite{radosavljevic2025railway}, corresponding to a per-frame budget of approximately 33\,ms. For 60\,fps video streams, this budget tightens to 16.7\,ms~\cite{fu2025efficient}.

\item\textbf{Non-real-time anomalies} are gradual degradations (HVAC inefficiency, bearing wear, insulation aging) that allow offline or batch analysis integrated into condition-based maintenance~\cite{davari2021predictive, galvez2021fault}. These tasks prioritize robustness and interpretability over latency, and can be deployed on cloud or server platforms.
\end{itemize}

This distinction is operationally important because it directly constrains feasible model architectures, optimization strategies, and hardware platforms, as formalized in the decision framework of Section~\ref{sec:decision_framework}.

\subsection{Model Optimization for Embedded Railway Deployment}
\label{sec:model_optimization}

\begin{figure}[htbp]
\centering
\includegraphics[width=\linewidth]{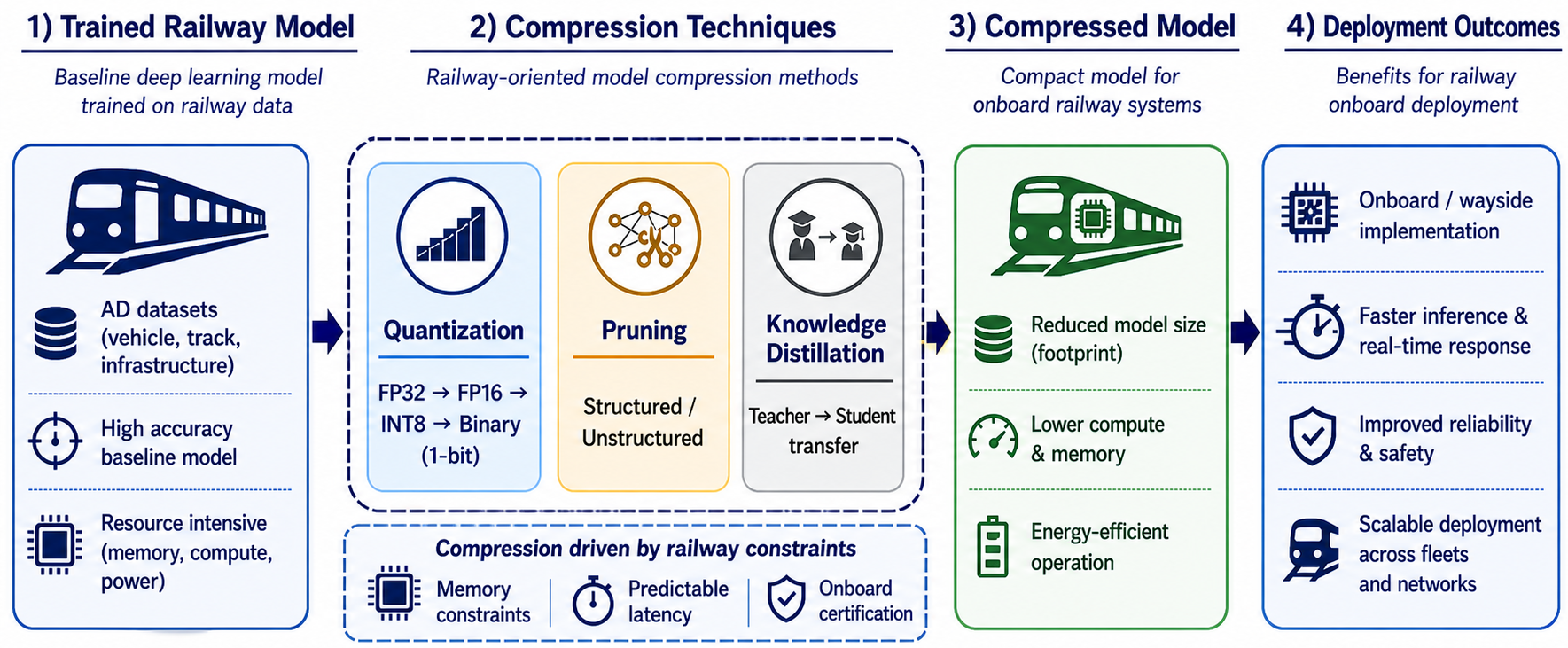}
\caption{Model compression workflow: quantization, pruning, and knowledge distillation transform a baseline model into a compact version for embedded railway deployment.}
\label{fig:model_compression}
\end{figure}

Unlike general embedded AI applications, railway deployment imposes additional constraints on determinism, long-term reliability, and safety certification that limit which optimization strategies are applicable in practice~\cite{radosavljevic2025railway, kljaic2023overview}. Onboard and wayside platforms further constrain memory, power, and thermal budgets, limiting the deployability of large deep learning models~\cite{benmeziane2021comprehensive, li2023model, cheng2020survey}. Six optimization strategies are identified in the surveyed literature, grouped into two categories: techniques that primarily reduce latency, and techniques that primarily reduce model size. Figure~\ref{fig:model_compression} illustrates the compression workflow.

\subsubsection{Latency-oriented techniques}
\begin{itemize}[noitemsep, topsep=2pt]
    \item \textbf{Lightweight architectures} such as MobileNetV3, SqueezeNet, and Efficient Net-Lite use depthwise separable convolutions to reduce computational cost on edge hardware~\cite{benmeziane2021comprehensive}. In the railway domain, MobileNetV3 combined with Transformer encoders achieved latency below 100\,ms for foreign object detection~\cite{ning2024research}. Radosavljevic et al.~\cite{radosavljevic2025railway} systematically evaluated six YOLO variants (v5 through v12) on two Jetson platforms, finding that YOLOv5nu consistently achieved the best accuracy--latency trade-off across both FP32 and FP16 precision levels, with mean latency of 20--50\,ms on Jetson AGX Xavier depending on dataset complexity.

\item\textbf{Input reduction} lowers computational load through frame downsampling or dimensionality reduction. Laurendin et al.~\cite{laurendin2021hazardous} applied optical flow on reduced frame windows for efficient real-time door surveillance.

\item\textbf{Hardware-accelerated inference} uses platform-specific runtimes (TensorRT, ONNX Runtime, TFLite) to optimize execution. Tsiktsiris et al.~\cite{tsiktsiris2024multimodal} achieved 2--3$\times$ speedup using INT8 quantization with TensorRT on Jetson platforms. Radosavljevic et al.~\cite{radosavljevic2025railway} confirmed that TensorRT export provides the best inference performance across all tested YOLO variants on both Jetson AGX Xavier and Orin Nano, with FP16 inference reducing latency by 40--50\% compared to FP32 with negligible accuracy loss ($<$0.3\% mAP drop in most cases).

\item\textbf{Parallel pipelines} enable asynchronous processing across CPU threads and GPU streams, reducing end-to-end latency in multimodal systems~\cite{tsiktsiris2024multimodal}.
\end{itemize}

\subsubsection{Size-oriented techniques}
\begin{itemize}[noitemsep, topsep=2pt]
    \item \textbf{Quantization} reduces numerical precision from FP32 to FP16, INT8, or lower. It is the most commonly applied compression technique in railway deployments. Fu et al.~\cite{fu2025efficient} demonstrate the most extreme case: binary neural networks (1-bit weights and activations) deployed on FPGA, achieving 93.4\% accuracy with only 0.3\% degradation from full precision and power consumption of 2.3--3.9\,W. Li et al.~\cite{li2024edge} apply mixed fixed-point quantization (12-bit and 22-bit) on FPGA with no accuracy loss. Tsiktsiris et al.~\cite{tsiktsiris2024multimodal} report INT8 quantization on GPU edge with $\sim$0.5\% accuracy drop. These results confirm that quantization is effective for railway deployment, with accuracy degradation typically below 1\% when calibration is performed carefully~\cite{jacob2018quantization}.

\item\textbf{Pruning} removes redundant weights or filters to reduce model complexity~\cite{cheng2020survey}. Fu et al.~\cite{fu2025efficient} apply entropy-based pruning to reduce network depth before FPGA deployment. Beyond this, explicit pruning strategies are rare in the railway literature; most works rely on lightweight architectures rather than formal pruning pipelines~\cite{weng2025improved}.

\item\textbf{Knowledge distillation} transfers representations from a large teacher model to a compact student model~\cite{hinton2015distilling}. This technique remains essentially absent from railway anomaly detection, though it has shown promise for fault diagnosis under hardware constraints in adjacent domains~\cite{gong2023lightweight}.
\end{itemize}

\subsubsection{Summary}

Existing railway deployments primarily exploit quantization and lightweight architecture selection. Pruning is applied in one study~\cite{fu2025efficient}. Knowledge distillation and early-exit strategies~\cite{simioni2025early, bouzidi2024efficient} remain unexplored in this domain. No study applies a systematic end-to-end compression pipeline combining multiple techniques. This pattern suggests that most railway anomaly detection models are optimized for accuracy first and adapted for deployment post hoc, rather than designed with hardware constraints as a primary objective. Establishing systematic compression pipelines tailored to railway-specific constraints (deterministic latency, memory limits, and safety certification) remains an open challenge. More advanced approaches such as hardware-aware neural architecture search and hardware--software co-design, which integrate deployment constraints directly into model design, are discussed as future research directions in Section~\ref{sec:future}.

\subsection{Deployment-Driven Decision Framework}
\label{sec:decision_framework}

The preceding subsections established three complementary views: (i)~the empirical deployment evidence from the surveyed literature (Section~\ref{sec:deploy_evidence}), (ii)~the hardware platforms and deployment architectures available for railway systems (Sections~\ref{sec:deploy_platforms}--\ref{sec:deploy_arch}), and (iii)~the model optimization techniques applied in practice (Section~\ref{sec:model_optimization}). Combined with the four-axis anomaly taxonomy (Section~\ref{sec:Railway Anomalies Categorization}) and the detection paradigms (Section~\ref{sec:3}), these elements provide the building blocks for systematic deployment planning. However, they do not directly answer the practitioner's question: \textbf{which detection paradigm, model family, and deployment configuration should be selected for a given railway scenario?}

\begin{figure}[H]
\centering
\includegraphics[width=\textwidth]{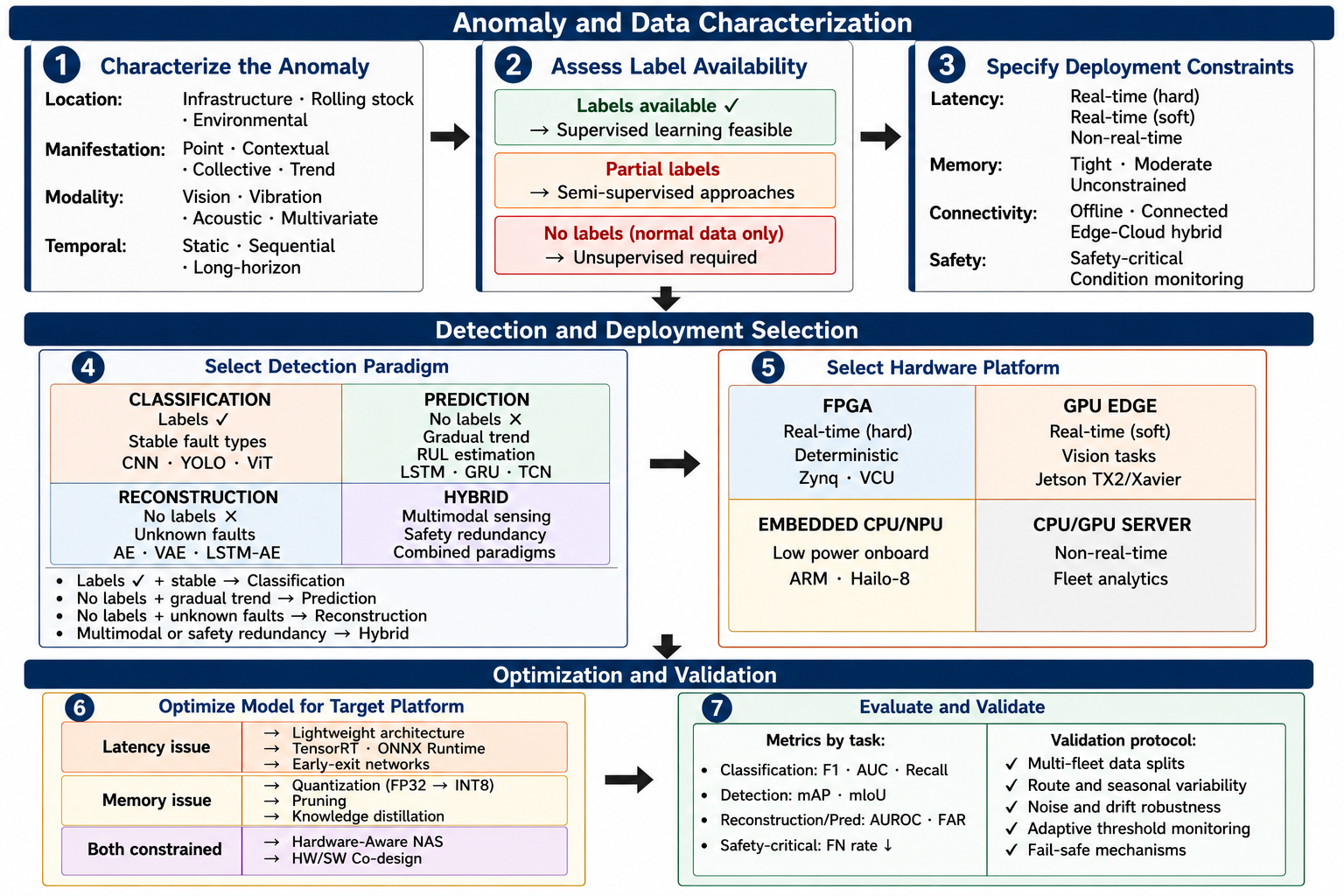}
\caption{Deployment decision framework for deep learning-based anomaly detection in railway systems.}
\label{fig:decision_framework}
\end{figure}

This subsection introduces a deployment-driven decision framework that maps (A)~anomaly and data characteristics and (B)~operational constraints to (C)~a justified selection of detection paradigm, hardware platform, optimization strategy, and evaluation protocol. The framework is operationalized in Figure~\ref{fig:decision_framework}, which summarizes the complete decision path from anomaly characterization to deployment validation. The framework synthesizes the surveyed evidence with railway engineering requirements and is intended as design guidance rather than a replacement for domain-specific risk analysis or safety certification procedures such as EN~50128~\cite{kljaic2023overview}.

\noindent\textbf{Scope:} The framework targets safety-relevant railway monitoring applications, including infrastructure inspection, onboard rolling-stock subsystems, and environmental risk scenarios.

\subsubsection{Stepwise Selection Procedure}
\label{subsubsec:decision_steps}

For a given railway anomaly detection scenario, the following seven steps should be applied sequentially.

\begin{enumerate}[topsep=2pt]

\item \textbf{Characterize the anomaly:}
Describe the target anomaly using the four-axis taxonomy introduced in Section~\ref{sec:Railway Anomalies Categorization}: location (infrastructure, rolling stock, environmental), manifestation (point, contextual, collective, trend), sensing modality (visual, vibration, acoustic, multivariate), and temporal behavior (static, sequential, long-horizon). This step determines whether spatial localization, temporal modeling, or multimodal fusion is required.

\item \textbf{Assess label availability and data regime:}
Determine whether annotated fault examples are available at scale, partially available, or absent. In railway systems, fault rarity, data confidentiality, and class imbalance frequently limit supervision~\cite{chalapathy2019deep, pang2021deep}. Normal-only learning, reconstruction, and prediction strategies are often more realistic than purely supervised classification.

\item \textbf{Specify deployment constraints:}
Define the operational constraints that affect both model and platform selection:
\begin{itemize}[noitemsep]
\item \textbf{Latency:} hard real-time ($<$10\,ms, e.g., braking protection), soft real-time ($<$100\,ms, e.g., door monitoring), or offline/batch.
\item \textbf{Compute and memory budget:} tight embedded ($<$8\,W, $<$1\,GB), moderate edge (10--30\,W, 4--32\,GB), or unconstrained.
\item \textbf{Connectivity:} offline-capable, connected, or hybrid edge-cloud.
\item \textbf{Safety requirement:} safety-critical (SIL2+) or condition-monitoring.
\end{itemize}

\item \textbf{Select the detection paradigm:}
Based on Steps~1--3, the following rules guide paradigm selection:
\begin{itemize}[noitemsep]
\item \textbf{Labels available} and stable recurring fault types $\rightarrow$ \textbf{classification-based}.
\item \textbf{No labels} and temporal or gradual degradation $\rightarrow$ \textbf{prediction-based} or \textbf{reconstruction-based}.
\item \textbf{No labels} and rare, unknown, or weakly characterized faults $\rightarrow$ \textbf{reconstruction-based}.
\item \textbf{Multimodal sensing} or safety redundancy required $\rightarrow$ \textbf{hybrid}.
\end{itemize}
Table~\ref{tab:decision_matrix_refined} provides an analytical comparison supporting these rules.

\item \textbf{Select the hardware platform:}
Platform choice is driven by the latency and power constraints specified in Step~3:
\begin{itemize}[noitemsep]
\item Hard real-time and deterministic latency $\rightarrow$ \textbf{FPGA}~\cite{liu2020real, fu2025efficient, li2024edge}.
\item Soft real-time vision or multimodal inference $\rightarrow$ \textbf{GPU edge} (e.g., Jetson TX2, Xavier, Orin)~\cite{tsiktsiris2024multimodal, cui2019real}.
\item Low-power onboard inference $\rightarrow$ \textbf{embedded CPU} or \textbf{NPU}.
\item Offline training or fleet-level analytics $\rightarrow$ \textbf{CPU/GPU server}.
\end{itemize}

\item \textbf{Optimize the model for the target platform:}
The selected model must be adapted to the target hardware budget:
\begin{itemize}[noitemsep]
\item If latency is limiting $\rightarrow$ lightweight architectures or hardware-accelerated inference~\cite{tsiktsiris2024multimodal, ning2024research, radosavljevic2025railway}.
\item If memory is limiting $\rightarrow$ quantization, pruning, or knowledge distillation~\cite{jacob2018quantization, hinton2015distilling, cheng2020survey}.
\item If both are limiting $\rightarrow$ consider hardware-aware neural architecture search or hardware--software co-design~\cite{benmeziane2021comprehensive, cai2018proxylessnas, bouzidi2022hadas}.
\end{itemize}

\item \textbf{Define the evaluation protocol and validate under operational variability:}
Evaluation must reflect the detection task and safety requirements:
\begin{itemize}[noitemsep]
\item Classification: F1-score, minority-class recall, and AUROC.
\item Detection and localization: mAP and mIoU.
\item Reconstruction and prediction: PR-AUC, anomaly-score AUROC, and false alarm rate.
\item Safety-critical deployment: false negative rate, conservative thresholds, and uncertainty monitoring.
\item All deployed systems: inference latency on target platform and power consumption.
\end{itemize}
Validation should include operational variability across fleets, routes, seasons, speeds, and maintenance states. Drift monitoring and adaptive thresholding should be considered for long-term deployment~\cite{correia2024online, zamanzadeh2024deep}.

\end{enumerate}

\subsubsection{Analytical Comparison of Detection Paradigms}
\label{subsubsec:decision_matrix}

Table~\ref{tab:decision_matrix_refined} complements the rule-based framework by comparing the four paradigms across ten engineering criteria relevant to railway deployment. It is intended as a structured summary supporting the selection rules in Step~4, not as a standalone decision tool.

\noindent\textbf{Rating scale and criteria:} Each cell is rated as High~(H), Medium~(M), or Low~(L) based on the following definitions:
\begin{itemize}[noitemsep]
\item \textbf{High (H):} The paradigm satisfies the criterion under typical railway conditions without requiring additional mechanisms. \textbf{Example:} Classification achieves high interpretability because it produces explicit fault labels that maintenance engineers can act on directly.
\item \textbf{Medium (M):} Suitability depends on implementation choices, data quality, or operational conditions, and may require moderate adaptation. \textbf{Example:} Prediction handles temporal dependencies well but requires threshold recalibration when operating conditions change (e.g., seasonal variation, fleet differences).
\item \textbf{Low (L):} The paradigm does not naturally satisfy the criterion and typically requires substantial additional mechanisms such as data augmentation, domain adaptation, or hybridization. \textbf{Example:} Classification has low coverage of unseen anomalies because it assumes a closed set of known fault types and fails silently on out-of-distribution inputs.
\end{itemize}

\begin{table}[htbp]
\centering
\caption{Analytical comparison of detection paradigms for deployment-oriented railway anomaly detection. H:~High suitability; M:~Medium; L:~Low.}
\label{tab:decision_matrix_refined}
\scriptsize
\renewcommand{\arraystretch}{1.25}
\begin{tabular}{@{}|p{3.2cm}|p{1.7cm}|p{1.7cm}|p{1.9cm}|p{1.5cm}|@{}}
\hline
\textbf{Criterion} & \textbf{Classification} & \textbf{Prediction} & \textbf{Reconstruction} & \textbf{Hybrid} \\
\hline
Suitability when fault labels are available & H & M & M & H \\
\hline
Suitability with normal data only & L & H & H & M \\
\hline
Coverage of unseen anomalies & L & M & H & H \\
\hline
Robustness to class imbalance & L & M & H & M \\
\hline
Temporal dependency handling & M & H & M--H & H \\
\hline
Tolerance to operational variability & L--M & M & M & H \\
\hline
Interpretability for maintenance & H & M & L--M & M \\
\hline
Embedded feasibility (tight budget) & H & M & M & L--M \\
\hline
Multimodal extensibility & M & M & M & H \\
\hline
Safety redundancy potential & L & M & M & H \\
\hline
\end{tabular}
\end{table}

\subsubsection{Illustrative Case Studies}
\label{subsubsec:decision_cases}

Two representative scenarios are used to illustrate how the stepwise procedure maps anomaly characteristics, label availability, deployment constraints, paradigm selection, hardware choice, model optimization, and evaluation metrics to concrete deployment recommendations.

\vspace{6pt}
\begin{enumerate}
    \item\textbf{Case~1: Real-time wheel flat detection from vibration signals}

\noindent\textbf{Step~1 Anomaly characterization:} The anomaly is a rolling-stock-level fault (wheel flat) that manifests as periodic abnormal vibration patterns, with sequential temporal behavior as the defect persists or worsens over operating cycles.

\noindent\textbf{Step~2 Label availability:} Labeled vibration datasets with annotated wheel flat events are available from test rig experiments~\cite{shim2023methodology}. The data regime supports supervised learning.

\noindent\textbf{Step~3 Deployment constraints:} Detection must occur onboard during operation. The latency requirement is soft real-time ($<$50\,ms per inference cycle). Power budget is moderate (10--30\,W). The system must operate offline (no continuous connectivity). Safety classification: condition-monitoring with escalation to safety-critical if severity thresholds are exceeded.

\noindent\textbf{Step~4 Paradigm selection:} Labels are available and the fault type is recurring and well-characterized $\rightarrow$ \textbf{classification-based} approach. A spectrogram-based CNN converts vibration windows into time-frequency images for supervised classification~\cite{shim2023methodology}.

\noindent\textbf{Step~5 Hardware platform:} Soft real-time with moderate power budget $\rightarrow$ \textbf{GPU edge platform} (e.g., Jetson Xavier NX). If deterministic latency is required for integration with braking systems, an \textbf{FPGA} should be considered instead.

\noindent\textbf{Step~6 Model optimization:} The spectrogram CNN is lightweight ($<$5M parameters). FP16 quantization via TensorRT can be used to reduce inference time and memory usage, but the final accuracy--latency trade-off must be validated on the target platform~\cite{radosavljevic2025railway}.

\noindent\textbf{Step~7 Evaluation:} Primary metrics: recall (to minimize missed flats), F1-score, and false negative rate. Validation must cover multiple wheel wear levels, train speeds (80--300\,km/h), and seasonal temperature variations. Inference latency and power consumption must be reported on the target platform.

\vspace{6pt}
\item\textbf{Case~2: Door anomaly detection from current signals}

\noindent\textbf{Step~1 Anomaly characterization:} The anomaly is a rolling-stock-level fault (door obstruction or mechanical degradation) that manifests as a collective or contextual deviation in sequential current signals recorded during each door opening/closing cycle.

\noindent\textbf{Step~2 Label availability:} Fault labels are unavailable. Only normal door cycles are recorded during routine operation~\cite{shimizu2022real}. The data regime requires unsupervised or normal-only learning.

\noindent\textbf{Step~3 Deployment constraints:} Detection should occur onboard in near-real-time (soft real-time, $<$200\,ms per cycle). Power budget is tight ($<$10\,W). Connectivity is intermittent. Safety classification: condition-monitoring.

\noindent\textbf{Step~4 Paradigm selection:} No labels, sequential temporal behavior, collective anomaly pattern $\rightarrow$ \textbf{reconstruction-based} approach. A bidirectional LSTM autoencoder trained on normal cycles flags deviations via reconstruction error~\cite{shimizu2022real}.

\noindent\textbf{Step~5 Hardware platform:} Tight power budget with soft real-time requirement $\rightarrow$ \textbf{embedded CPU} for lightweight LSTM models, or \textbf{GPU edge} (Jetson Nano at $\sim$5--10\,W) if model complexity requires GPU acceleration.

\noindent\textbf{Step~6 Model optimization:} INT8 quantization reduces the LSTM autoencoder memory footprint. No pruning or distillation is needed given the small model size ($<$500K parameters).

\noindent\textbf{Step~7 Evaluation:} Primary metrics: PR-AUC (due to severe class imbalance), false alarm rate, and reconstruction error distribution analysis. Validation must include multiple door types, passenger load conditions, and seasonal temperature effects on mechanical resistance. Threshold calibration and drift monitoring are essential for long-term deployment~\cite{correia2024online}. Inference latency and power consumption must be reported.
\end{enumerate}

Although the two detailed examples focus on onboard time-series monitoring, the same seven-step logic also applies to vision-based and offline inspection scenarios, such as image-based rail fastener or rail-surface defect detection. These scenarios are discussed within the vision-based detection paradigm in Section~5.1.2, where image-based classification and localization approaches are reviewed.


\subsubsection{Engineering Implications}
\label{subsubsec:engineering_implications}

For deployment-oriented railway anomaly detection studies, the following information should be reported at minimum:
(i)~target hardware platform and latency constraints,
(ii)~label availability and class imbalance level,
(iii)~optimization strategy applied (quantization level, pruning ratio, or distillation setup),
(iv)~decision policy and threshold calibration method, and
(v)~robustness evaluation under noise, drift, and fleet variability.

The proposed framework transforms the survey taxonomy and paradigm analysis into an end-to-end decision methodology. It connects anomaly characterization, supervision level, deployment constraints, hardware choice, model optimization, and evaluation protocol, thereby providing practitioners with a structured and reproducible process for selecting and deploying deep learning solutions for railway anomaly detection.

\section{Future Research Directions}
\label{sec:future}

Several research directions emerge from the limitations identified in this survey. 
These directions concern not only detection accuracy, but also generalization, deployment readiness, reproducibility, and safety assurance in real railway environments.

\textbf{Multimodal fusion and multi-task railway monitoring:}
Most existing studies focus on a single sensing modality and address one specific task, such as fault classification, defect detection, or condition monitoring. However, railway systems naturally generate heterogeneous data from vision, vibration, acoustic, current, voltage, log, and operational signals. Future work should therefore move toward unified multimodal architectures able to combine complementary information across subsystems. Such models should not only detect anomalies, but also support fault localization, severity estimation, and condition assessment within the same framework~\cite{liyuan2024multi, liu2026bearing}.

\textbf{Open datasets and standardized evaluation protocols:}
A major obstacle to progress remains the lack of public railway anomaly detection benchmarks. Most studies rely on proprietary datasets, use different fault definitions, and report results under incompatible experimental settings. This makes cross-study comparison difficult and limits reproducibility. Future efforts should prioritize the creation of open, well-documented datasets covering representative railway subsystems, fault types, operating conditions, and sensor modalities. In addition, standardized evaluation protocols are needed, including clear train-test split strategies, cross-route and cross-fleet validation, cost-sensitive metrics, false-alarm analysis, and inference latency reporting~\cite{pappaterra2021systematic}.

\textbf{Hardware-aware neural architecture search and hardware--software co-design:}
Current railway anomaly detection studies generally address deployment constraints after training, using compression techniques such as quantization, pruning, or knowledge distillation. Although these methods reduce model size and inference cost, they cannot fully compensate for an architecture that was not designed for the target hardware from the beginning. Future research should therefore explore hardware-aware neural architecture search (HW-NAS) as a more principled approach for railway applications. Instead of optimizing only detection performance, HW-NAS can jointly consider accuracy, latency, memory footprint, and energy consumption during the architecture search process~\cite{benmeziane2021comprehensive, cai2018proxylessnas}.

However, despite its potential, HW-NAS remains largely underexplored in railway anomaly detection for three main reasons. First, although the search is performed offline and does not increase inference cost after deployment, it increases development complexity because many candidate architectures must be trained, evaluated, or estimated under realistic hardware constraints~\cite{sinha2024multiobjective, sinha2023representation}. Second, the railway domain lacks standardized deployment-oriented benchmarks: most studies rely on task-specific or proprietary datasets and rarely report common hardware platforms, latency budgets, memory limits, or energy measurements. This prevents fair comparison of hardware-aware search strategies across railway subsystems. Third, when anomaly detection models support safety-related decision chains, automatically searched architectures require careful validation in terms of reproducibility, robustness, and auditability. More generally, current railway anomaly detection research still focuses mainly on detection accuracy and proof-of-concept validation, while systematic architecture search under hardware constraints remains uncommon.

This direction is especially relevant for railway-grade embedded platforms, where models must operate under strict constraints related to real-time response, limited resources, vibration, temperature variation, electromagnetic compatibility, and long-term reliability. Future studies should define railway-specific hardware benchmarks and latency budgets for different subsystems. Beyond HW-NAS, hardware--software co-design offers an even broader perspective by jointly optimizing the neural architecture and the deployment platform~\cite{guo2025survey, bouzidi2022hadas, ghebriout2023harmonicnas}.
Adjacent railway control studies, such as optimization-based control for maglev suspension, also highlight the importance of robustness in railway-grade systems, although they remain outside the core corpus of deep learning-based anomaly detection~\cite{teklu2023genetic}.

\textbf{Domain adaptation and cross-fleet generalization:}
Another important limitation is the lack of evaluation across fleets, routes, seasons, and operating conditions. Models trained on one vehicle type, operator, or geographic region may not generalize to another due to differences in sensors, maintenance practices, infrastructure quality, weather, loading conditions, and operational profiles. Future work should investigate domain adaptation, transfer learning, and federated learning to improve robustness across heterogeneous railway environments. These approaches could reduce the need for each operator to build large labeled datasets from scratch, while enabling knowledge transfer between fleets and preserving data confidentiality~\cite{rakshit2022railway,hao2023track, zamanzadeh2024deep}.

\textbf{Foundation models and large language models for railway monitoring:}
Foundation models and large language models open new possibilities for railway monitoring, particularly when heterogeneous data sources are available. Vision foundation models could support few-shot defect detection in inspection images, while language models could help analyze maintenance reports, event logs, and operator notes. More importantly, multimodal foundation models may enable cross-modal reasoning between sensor signals, visual observations, and textual maintenance records~\cite{chen2025large}. However, their use in railway systems remains a long-term research direction. Their high computational cost, uncertain latency, limited interpretability, and difficult certification raise important barriers for safety-critical deployment. Future research should therefore focus on efficient, explainable, and certifiable foundation-model-based solutions rather than directly transferring large generic models to railway applications.

\section{Conclusion}
\label{sec:conclusion}

This survey presented a structured and deployment-oriented review of deep learning-based anomaly detection in railway systems, covering infrastructure, rolling stock, and operational contexts. Four complementary contributions were developed to support systematic method selection and deployment planning.

A four-axis anomaly taxonomy was introduced to characterize railway anomalies by location, manifestation, sensing modality, and temporal behavior. Deep learning approaches were organized into four detection paradigms-classification-, prediction-, reconstruction-, and hybrid-based-and analyzed with respect to data availability, supervision requirements, and deployment feasibility. Evaluation practices were critically reviewed, showing that conventional accuracy-centric metrics are insufficient for safety-critical applications; an expanded framework integrating cost-sensitive assessment, interpretability, and latency reporting was proposed. Finally, a deployment-driven decision framework was introduced to translate anomaly characteristics and operational constraints into justified paradigm and architecture choices, with particular emphasis on embedded platforms, model compression, and hardware-aware design.

Taken together, these contributions establish a unified reference for researchers and engineers at the intersection of deep learning and railway systems, bridging the gap between academic performance and operational feasibility in safety-critical monitoring applications.

\scriptsize
\bibliographystyle{elsarticle-num} 
\bibliography{references}

@article{wang2024anomaly,
  title={An anomaly-free representation learning approach for efficient railway foreign object detection},
  author={Wang, Tiange and Zhang, Zijun and Xie, Min},
  journal={IEEE Transactions on Industrial Informatics},
  volume={20},
  number={10},
  pages={12125--12135},
  year={2024},
  publisher={IEEE}
}

@article{wang2024hierarchical,
  author  = {T. Wang and Z. Zhang and K. L. Tsui},
  title   = {A Hierarchical Transfer-Generative Framework for 
             Automating Multianalytical Tasks in Rail Surface 
             Defect Inspection},
  journal = {IEEE Internet of Things Journal},
  year    = {2024},
  doi     = {10.1109/JIOT.2024.3374751}
}

@ARTICLE{teklu2023genetic,
  author={Teklu, Esaias Abera and Abdissa, Chala Merga},
  journal={IEEE Access}, 
  title={Genetic Algorithm Tuned Super Twisting Sliding Mode Controller for Suspension of Maglev Train With Flexible Track}, 
  year={2023},
  volume={11},
  number={},
  pages={30955-30969},
  doi={10.1109/ACCESS.2023.3262416}}

@inproceedings{sinha2023representation,
  title={Hardware aware evolutionary neural architecture search using representation similarity metric},
  author={Sinha, Nilotpal and El Rahman Shabayek, Abd and Kacem, Anis and Rostami, Peyman and Shneider, Carl and Aouada, Djamila},
  booktitle={Proceedings of the IEEE/CVF Winter Conference on Applications of Computer Vision},
  pages={2628--2637},
  year={2024}
}

@inproceedings{sinha2024multiobjective,
  title={Multi-objective hardware aware neural architecture search using hardware cost diversity},
  author={Sinha, Nilotpal and Rostami, Peyman and El Rahman Shabayek, Abd and Kacem, Anis and Aouada, Djamila},
  booktitle={Proceedings of the IEEE/CVF Conference on Computer Vision and Pattern Recognition},
  pages={8032--8039},
  year={2024}
}

@article{gong2024train,
  title={Train-to-edge cooperative intelligence for obstacle intrusion detection in rail transit},
  author={Gong, Taiyuan and Zhu, Li and Yu, F Richard and Tang, Tao},
  journal={IEEE Transactions on Vehicular Technology},
  volume={73},
  number={6},
  pages={7669--7680},
  year={2024},
  publisher={IEEE}
}

@article{bhadouria2025advances,
  title={Advances in railway condition monitoring and fault diagnosis: Transition from industry 4.0 automation to industry 5.0 human-centric maintenance},
  author={Bhadouria, Abhishek S and Mishra, Rajesh P and Jalan, Arun K and Andrade, Ant{\'o}nio R},
  journal={Proceedings of the Institution of Mechanical Engineers, Part F: Journal of Rail and Rapid Transit},
  pages={09544097261417814},
  year={2025},
  publisher={SAGE Publications Sage UK: London, England}
}

@article{fernandez2025gan,
  title={GAN-based data augmentation of time series for fault diagnosis in railway track: HA Fern{\'a}ndez-Bobadilla et al.},
  author={Fern{\'a}ndez-Bobadilla, H{\'e}ctor A and Bouchikhi, Yahya and Martin, Ullrich},
  journal={Railway Engineering Science},
  volume={33},
  number={4},
  pages={642--683},
  year={2025},
  publisher={Springer}
}

@article{larese2025transformer,
  title={Transformer vibration forecasting for advancing rail safety and maintenance 4.0},
  author={Larese, Dar{\'\i}o C and Cerrada, Almudena Bravo and Tomei, Gabriel Dambrosio and Guerrero-L{\'o}pez, Alejandro and Olmos, Pablo M and Garc{\'\i}a, Mar{\'\i}a Jes{\'u}s G{\'o}mez},
  journal={IEEE Open Journal of Intelligent Transportation Systems},
  volume={6},
  pages={1612--1624},
  year={2025},
  publisher={IEEE}
}

@article{zhukov2024hybrid,
  title={A hybrid system for defect detection on rail lines through the fusion of object and context information},
  author={Zhukov, Alexey and Rivero, Alain and Benois-Pineau, Jenny and Zemmari, Akka and Mosbah, Mohamed},
  journal={Sensors},
  volume={24},
  number={4},
  pages={1171},
  year={2024},
  publisher={MDPI}
}

@inproceedings{yokouchi2021lstm,
  title={{LSTM}-based Anomaly Detection for 
         Railway Vehicle Air-conditioning Unit 
         using Monitoring Data},
  author={Yokouchi, T. and Kondo, M.},
  booktitle={IECON 2021 -- 47th Annual 
             Conference of the IEEE 
             Industrial Electronics Society},
  pages={1--6},
  year={2021},
  doi={10.1109/IECON48115.2021.9589473}
}

@article{ribeiro2023lstm,
  title={{LSTM}-based failure prediction for 
         railway rolling stock equipment},
  author={Ribeiro, Rita P. and others},
  journal={Expert Systems with Applications},
  volume={218},
  pages={119420},
  year={2023},
  doi={10.1016/j.eswa.2023.119420}
}

@article{kang2021lstm_ae,
  title={Anomaly detection of the brake 
         operating unit on metro vehicles 
         using a one-class {LSTM} autoencoder},
  author={Kang, Junyoung and Kim, Chang-Sun 
          and Kang, Jong-Won and 
          Gwak, Jeonghwan},
  journal={Applied Sciences},
  volume={11},
  number={19},
  pages={9290},
  year={2021},
  doi={10.3390/app11199290}
}

@inproceedings{davari2023lstm_ae,
  title={Detecting and Explaining Anomalies 
         in the Air Production Unit of a Train},
  author={Davari, Narjes and Veloso, Bruno 
          and Ribeiro, Rita P. and 
          Gama, Jo{\~a}o},
  booktitle={Proceedings of the 39th 
             ACM/SIGAPP Symposium on 
             Applied Computing},
  pages={1--8},
  year={2024},
  doi={10.1145/3605098.3635906}
}

@article{seo2024earthquake,
  title={Unsupervised anomaly detection for 
         earthquake detection on {Korea} 
         high-speed trains using 
         autoencoder-based deep learning models},
  author={Seo, Jiyeon and Kim, Youngho 
          and Ha, Jimin and others},
  journal={Scientific Reports},
  volume={14},
  pages={639},
  year={2024},
  doi={10.1038/s41598-024-51354-7}
}

@article{wang2023memformer,
  title={{MemFormer}: A memory based unified 
         model for anomaly detection on metro 
         railway tracks},
  author={Wang, Lei and Zhang, Kai and 
          Zheng, Qing and Ding, Guofu and 
          Zhang, Weihua and Chen, Dejun 
          and Liu, Bin},
  journal={Expert Systems with Applications},
  volume={238},
  pages={121824},
  year={2024},
  doi={10.1016/j.eswa.2023.121824}
}

@article{wang2024undercarriage,
  title={An undercarriage image driven 
         anomaly detection method for metro 
         vehicle based on adversarial 
         memory enhancement},
  author={Wang, Lei and Zhang, Kai 
          and others},
  journal={Proceedings of the Institution 
          of Mechanical Engineers, Part F: 
          Journal of Rail and Rapid Transit},
  year={2024},
  doi={10.1177/09544097231201519}
}

@article{chen2022meta,
  title={Meta-learning with {GANs} for 
         anomaly detection, with deployment 
         in high-speed rail inspection system},
  author={Chen, Yinan and others},
  journal={arXiv preprint arXiv:2202.05795},
  year={2022}
}

@article{cheng2020survey,
  title={A survey of model compression and acceleration for deep neural networks},
  author={Cheng, Yu and Wang, Duo and Zhou, Pan and Zhang, Tao},
  journal={arXiv preprint arXiv:1710.09282},
  year={2017}
}

@inproceedings{bouzidi2022hadas,
  title={{HADAS}: Hardware-Aware Dynamic 
         Neural Architecture Search for 
         Edge Performance Scaling},
  author={Bouzidi, Halima and Odema, 
          Mohanad and Ouarnoughi, Hamza 
          and Al Faruque, Mohammad 
          Abdullah and Niar, Smail},
  booktitle={Design, Automation and 
             Test in Europe Conference 
             (DATE)},
  year={2023},
  url={https://arxiv.org/abs/2212.03354}
}

@article{you2026gated,
  title={Gated Memory-Guided Multi-scale spatio--temporal--spectral feature fusion network for unsupervised Internet of Things time series anomaly detection},
  author={You, Peng and Chen, Peng and Li, Xi and Bian, Ang},
  journal={Engineering Applications of Artificial Intelligence},
  volume={169},
  pages={114104},
  year={2026},
  publisher={Elsevier}
}

@article{du2024study,
  title={Study on automatic tracking system of microwave deicing device for railway contact wire},
  author={Du, Guanfeng and Zhang, Hongzheng and Yu, Hanbo and Hou, Peng and He, Junbo and Cao, Shengxian and Wang, Gong and Ma, Le},
  journal={IEEE Transactions on Instrumentation and Measurement},
  volume={73},
  pages={1--11},
  year={2024},
  publisher={IEEE}
}

@article{fu2025efficient,
  title={An efficient fpga-based edge ai system for railway fault detection},
  author={Fu, Yulin and Yan, Dongwei and Li, Jiale and Ma, Sean Longyu and Sham, Chiu-Wing and Chou, Hong-fu},
  journal={IEEE Consumer Electronics Magazine},
  volume={14},
  number={4},
  pages={76--86},
  year={2025},
  publisher={IEEE}
}

@article{li2024edge,
  title={An edge ai system based on fpga platform for railway fault detection},
  author={Li, Jiale and Fu, Yulin and Yan, Dongwei and Ma, Sean Longyu and Sham, Chiu-Wing},
  booktitle={2024 IEEE 13th Global Conference on Consumer Electronics (GCCE)},
  pages={1387--1389},
  year={2024},
  organization={IEEE}
}

@article{diaz2025real,
  title={Real-Time Passenger Flow Analysis in Tram Stations Using YOLO-Based Computer Vision and Edge AI on Jetson Nano},
  author={Diaz-Santos, Sonia and Caballero-Gil, Pino and Caballero-Gil, C{\'a}ndido},
  journal={Computers},
  volume={14},
  number={11},
  pages={476},
  year={2025},
  publisher={MDPI}
}

@article{smart2025vit,
  title={Toward smart railway maintenance: AI-enhanced Non-Destructive Evaluation using Vision Transformers and CNNs for fastener defect detection},
  author={Mohammadi, Samira and Karganroudi, Sasan Sattarpanah and Adda, Mehdi and Ibrahim, Hussein},
  journal={Green Energy and Intelligent Transportation},
  volume={5},
  number={3},
  pages={100332},
  year={2026}
}

@article{liu2026bearing,
  title={Bearing Anomaly Detection Method Based on Multimodal Fusion and Self-Adversarial Learning},
  author={Liu, Han and Qin, Yong and Tu, Dilong},
  journal={Sensors},
  volume={26},
  number={2},
  pages={629},
  year={2026},
  publisher={MDPI}
}

@article{liyuan2024multi,
  title={Multi-modal fusion for multi-task fuzzy detection of rail anomalies},
  author={Liyuan, Yang and Osman, Ghazali and Rahman, Safawi Abdul and Mustapha, Muhammad Firdaus},
  journal={IEEE Access},
  volume={12},
  pages={73925--73935},
  year={2024},
  publisher={IEEE}
}

@inproceedings{donnarumma201950128,
  title={EN-50128 certification-oriented design of a safety-critical hard real-time kernel},
  author={Donnarumma, Ciro and Fara, Pietro and Serra, Gabriele and Di Leonardi, Sandro and Marinoni, Mauro},
  booktitle={2019 IEEE International Symposium on Software Reliability Engineering Workshops (ISSREW)},
  pages={314--317},
  year={2019},
  organization={IEEE}
}

@article{radosavljevic2025railway,
  title={Railway Track Defect Detection: From a Comprehensive Review of Methods to New Embedded System Modeling Perspectives},
  author={Radosavljevic, Sa{\v{s}}a and Rivero, Alain and El Ouardi, Abdelhafid and Fl{\'o}rez, Sergio Rodr{\'\i}guez},
  journal={IEEE Transactions on Instrumentation and Measurement},
  volume={75},
  pages={1--25},
  year={2025},
  publisher={IEEE}
}

@article{chavhan2025bi,
  title={Bi-LSTM-Based Train Fault Detection System Using Edge Computing and IoT for Enhanced Railway Safety},
  author={Chavhan, Suresh and Gupta, Deepak and Rodrigues, Joel JPC},
  journal={IEEE Transactions on Intelligent Transportation Systems},
  year={2025},
  publisher={IEEE}
}

@inproceedings{ghebriout2023harmonicnas,
  title={{Harmonic-NAS}: Hardware-Aware 
         Multimodal Neural Architecture 
         Search on Resource-constrained 
         Devices},
  author={Ghebriout, Mohamed Imed Eddine 
          and Bouzidi, Halima and 
          Niar, Sma{\"i}l and 
          Ouarnoughi, Hamza},
  booktitle={Asian Conference on 
             Machine Learning (ACML)},
  year={2023},
  url={https://api.semanticscholar.org/
       CorpusID:261706155}
}

@inproceedings{jacob2018quantization,
  title={Quantization and training of neural networks for efficient integer-arithmetic-only inference},
  author={Jacob, Benoit and Kligys, Skirmantas and Chen, Bo and Zhu, Menglong and Tang, Matthew and Howard, Andrew and Adam, Hartwig and Kalenichenko, Dmitry},
  booktitle={Proceedings of the IEEE conference on computer vision and pattern recognition},
  pages={2704--2713},
  year={2018}
}

@article{hinton2015distilling,
  title={Distilling the Knowledge 
         in a Neural Network},
  author={Hinton, Geoffrey and 
          Vinyals, Oriol and 
          Dean, Jeff},
  journal={arXiv preprint 
           arXiv:1503.02531},
  year={2015}
}

@article{wu2023railtrack,
  title={Automatic Railroad Track Components 
         Inspection Using Hybrid Deep 
         Learning Framework},
  author={Wu, Yongle and Chen, Ping and 
          Qin, Yong and Qian, Yao and 
          Xu, Fei and Jia, Limin},
  journal={IEEE Transactions on 
          Instrumentation and Measurement},
  volume={72},
  pages={1--15},
  year={2023},
  doi={10.1109/TIM.2022.3225031}
}

@article{haghbin2024cnn1d,
  title={Predicting Rail Corrugation Based on 
         Convolutional Neural Networks Using 
         {Vehicle's} Acceleration Measurements},
  author={Haghbin, Masoud and Chiachi{\'o}, 
          Juan and others},
  journal={Sensors},
  volume={24},
  number={14},
  pages={4627},
  year={2024},
  doi={10.3390/s24144627}
}

@article{ding2026integrated,
  title={An Integrated IEWT and CNN--Transformer Deep Architecture for Intelligent Fault Diagnosis of Bogie Axle-Box Bearings},
  author={Ding, Xiaoping and Li, Zhongqi and Tang, Minghui and Shen, Xiaoxu and Zhou, Liang},
  journal={Electronics},
  volume={15},
  number={4},
  pages={804},
  year={2026},
  publisher={MDPI}
}

@article{zhang2024tcngru,
  title={Intelligent prediction of rail 
         corrugation evolution trend based on 
         self-attention bidirectional {TCN} 
         and {GRU}},
  author={Zhang, Chao and others},
  journal={Intelligent Robotics},
  volume={4},
  number={4},
  pages={548--565},
  year={2024},
  doi={10.20517/ir.2024.20}
}

@article{weng2025improved,
  title={An improved DeepLabv3+ railway track extraction algorithm based on densely connected and attention mechanisms},
  author={Weng, Yanbin and Yang, Jie and Zhang, Changfan and He, Jing and Peng, Cheng and Jia, Lin and Xiang, Hui},
  journal={Scientific Reports},
  volume={15},
  number={1},
  pages={2556},
  year={2025},
  publisher={Nature Publishing Group UK London}
}

@phdthesis{bouzidi2024efficient,
  title={Efficient deployment of deep neural networks on hardware devices for edge ai},
  author={Bouzidi, Halima},
  year={2024},
  school={Universit{\'e} Polytechnique Hauts-de-France}
}

@article{li2023model,
  title={Model compression for deep neural networks: A survey},
  author={Li, Zhuo and Li, Hengyi and Meng, Lin},
  journal={Computers},
  volume={12},
  number={3},
  pages={60},
  year={2023},
  publisher={MDPI}
}

@article{guo2025survey,
  title={A survey: Collaborative hardware and software design in the era of large language models},
  author={Guo, Cong and Cheng, Feng and Du, Zhixu and Kiessling, James and Ku, Jonathan and Li, Shiyu and Li, Ziru and Ma, Mingyuan and Molom-Ochir, Tergel and Morris, Benjamin and others},
  journal={IEEE Circuits and Systems Magazine},
  volume={25},
  number={1},
  pages={35--57},
  year={2025},
  publisher={IEEE}
}

@article{gong2023lightweight,
  title={Lightweight fault diagnosis method in embedded system based on knowledge distillation},
  author={Gong, Ran and Wang, Chenlin and Li, Jinxiao and Xu, Yi},
  journal={Journal of Mechanical Science and Technology},
  volume={37},
  number={11},
  pages={5649--5660},
  year={2023},
  publisher={Springer}
}

@article{amin2025integrating,
  title={Integrating AI and edge computing for advanced safety at railroad grade crossings},
  author={Amin, AL and Chimba, Deo and Hasan, Kamrul},
  journal={Journal of Rail Transport Planning \& Management},
  volume={33},
  pages={100501},
  year={2025},
  publisher={Elsevier}
}

@article{armijo2024integration,
  title={Integration of railway bridge structural health monitoring into the internet of things with a digital twin: a case study},
  author={Armijo, Alberto and Zamora-S{\'a}nchez, Diego},
  journal={Sensors},
  volume={24},
  number={7},
  pages={2115},
  year={2024},
  publisher={MDPI}
}

@article{shahin2023using,
  title={Using machine learning and deep learning algorithms for downtime minimization in manufacturing systems: An early failure detection diagnostic service},
  author={Shahin, Mohammad and Chen, F Frank and Hosseinzadeh, Ali and Zand, Neda},
  journal={The international journal of advanced manufacturing technology},
  volume={128},
  number={9},
  pages={3857--3883},
  year={2023},
  publisher={Springer}
}

@inproceedings{simioni2025early,
  title={An Early Exit Deep Neural Network for Fast Inference Intrusion Detection},
  author={Simioni, Jo{\~a}o and Viegas, Eduardo Kugler and Santin, Altair and Horchulhack, Pedro},
  booktitle={Proceedings of the 40th ACM/SIGAPP Symposium on Applied Computing},
  pages={730--737},
  year={2025}
}

@inproceedings{chang2019sparse,
  title={A sparse autoencoder based hyperspectral anomaly detection algorihtm using residual of reconstruction error},
  author={Chang, Shizhen and Du, Bo and Zhang, Liangpei},
  booktitle={IGARSS 2019-2019 IEEE International Geoscience and Remote Sensing Symposium},
  pages={5488--5491},
  year={2019},
  organization={IEEE}
}

@article{kljaic2023overview,
  title={An overview of current challenges and emerging technologies to facilitate increased energy efficiency, safety, and sustainability of railway transport},
  author={Kljai{\'c}, Zdenko and Pavkovi{\'c}, Danijel and Cipek, Mihael and Trstenjak, Maja and Mlinari{\'c}, Tomislav Josip and Nik{\v{s}}i{\'c}, Mladen},
  journal={Future Internet},
  volume={15},
  number={11},
  pages={347},
  year={2023},
  publisher={MDPI}
}

@inproceedings{bodo2021cost,
  title={Cost-sensitive fault identification in predictive maintenance applications: a case study},
  author={Bodo, Roberto and Bertocco, Matteo and Bianchi, Alberto},
  booktitle={2021 IEEE International Instrumentation and Measurement Technology Conference (I2MTC)},
  pages={1--6},
  year={2021},
  organization={IEEE}
}

@article{nechi2023fpga,
  title={Fpga-based deep learning inference accelerators: Where are we standing?},
  author={Nechi, Anouar and Groth, Lukas and Mulhem, Saleh and Merchant, Farhad and Buchty, Rainer and Berekovic, Mladen},
  journal={ACM Transactions on Reconfigurable Technology and Systems},
  volume={16},
  number={4},
  pages={1--32},
  year={2023},
  publisher={ACM New York, NY}
}

@article{ma2024cost,
  title={Cost-Sensitive Dual-Stream Residual Networks for Imbalanced Classification.},
  author={Ma, Congcong and Mi, Jiaqi and Gao, Wanlin and Tao, Sha},
  journal={Computers, Materials \& Continua},
  volume={80},
  number={3},
  year={2024}
}

@article{cheng2020review,
  title={A review of intelligent fault diagnosis for high-speed trains: Qualitative approaches},
  author={Cheng, Chao and Wang, Jiuhe and Chen, Hongtian and Chen, Zhiwen and Luo, Hao and Xie, Pu},
  journal={Entropy},
  volume={23},
  number={1},
  pages={1},
  year={2020},
  publisher={MDPI}
}

@article{wrobel2024fault,
  title={Fault Detection and Diagnostic Methods for Railway Systems--A Literature Survey},
  author={Wr{\'o}bel, Jakub and Bury, Pawe{\l} and Zaj{\k{a}}c, Mateusz and Kierzkowski, Artur and Tubek, S{\l}awomir and Blaut, J{\k{e}}drzej},
  journal={Advances in Science and Technology. Research Journal},
  volume={18},
  number={6},
  year={2024}
}

@article{mohammadi2025rail,
  title={Rail Defect Classification with Deep Learning Method},
  author={Mohammadi, Samira and Karganroudi, Sasan Sattarpanah and Adda, Mehdi and Ibrahim, Hussein},
  journal={Green Energy and Intelligent Transportation},
  pages={100332},
  year={2025},
  publisher={Elsevier}
}

@article{phaphuangwittayakul2024railtrack,
  title={RailTrack-DaViT: a vision transformer-based approach for automated railway track defect detection},
  author={Phaphuangwittayakul, Aniwat and Harnpornchai, Napat and Ying, Fangli and Zhang, Jinming},
  journal={Journal of Imaging},
  volume={10},
  number={8},
  pages={192},
  year={2024},
  publisher={MDPI}
}

@article{correia2024online,
  title={Online model-based anomaly detection 
         in multivariate time series: Taxonomy, 
         survey, research challenges and future 
         directions},
  author={Correia, L. and others},
  journal={Engineering Applications of 
           Artificial Intelligence},
  volume={138},
  pages={109323},
  year={2024}
}

@article{chen2024lidar,
  title={A real-time method for railway track 
         detection and {3D} fitting based on 
         camera and {LiDAR} fusion sensing},
  author={Chen, Yongpeng and others},
  journal={Remote Sensing},
  volume={16},
  number={8},
  pages={1441},
  year={2024}
}

@article{yurekli2025das,
  title={Real-time railway hazard detection 
         using distributed acoustic sensing 
         and hybrid ensemble learning},
  author={Y{\"u}rekli, Yusuf and 
          {\"O}zarpa, Cevat and 
          Avc{\i}, {\.I}sa},
  journal={Sensors},
  volume={25},
  number={13},
  pages={3992},
  year={2025}
}

@article{bai2018empirical,
  title={An empirical evaluation of generic 
         convolutional and recurrent networks 
         for sequence modeling},
  author={Bai, Shaojie and Kolter, J Zico 
          and Koltun, Vladlen},
  journal={arXiv preprint arXiv:1803.01271},
  year={2018}
}

@article{zamanzadeh2024deep,
  title={Deep learning for time series anomaly detection: A survey},
  author={Zamanzadeh Darban, Zahra and Webb, Geoffrey I and Pan, Shirui and Aggarwal, Charu and Salehi, Mahsa},
  journal={ACM Computing Surveys},
  volume={57},
  number={1},
  pages={1--42},
  year={2024},
  publisher={ACM New York, NY}
}

@article{hadj2026contributions,
  title={Contributions and limitations of AI and machine learning in railway operations, maintenance, and safety: A literature review},
  author={Hadj-Mabrouk, Habib},
  journal={Proceedings of the Institution of Mechanical Engineers, Part O: Journal of Risk and Reliability},
  volume={240},
  number={1},
  pages={3--36},
  year={2026},
  publisher={SAGE Publications Sage UK: London, England}
}

@article{tang2023comparative,
  title={A comparative review on multi-modal sensors fusion based on deep learning},
  author={Tang, Qin and Liang, Jing and Zhu, Fangqi},
  journal={Signal Processing},
  volume={213},
  pages={109165},
  year={2023},
  publisher={Elsevier}
}

@inproceedings{butakova2017complex,
  title={Complex event processing for network anomaly detection in digital railway communication services},
  author={Butakova, Maria A and Chernov, Andrey V and Shevchuk, Petr S and Vereskun, Vladimir D},
  booktitle={2017 25th Telecommunication Forum (TELFOR)},
  pages={1--4},
  year={2017},
  organization={IEEE}
}

@article{shim2022anomaly,
  title={Anomaly detection method in railway using signal processing and deep learning},
  author={Shim, Jaeseok and Koo, Jeongseo and Park, Yongwoon and Kim, Jaehoon},
  journal={Applied Sciences},
  volume={12},
  number={24},
  pages={12901},
  year={2022},
  publisher={MDPI}
}

@article{song2024fault,
  title={A Fault Detection System for Wiring Harness Manufacturing Using Artificial Intelligence},
  author={Song, Jinwoo and Kumar, Prashant and Kim, Yonghawn and Kim, Heung Soo},
  journal={Mathematics},
  volume={12},
  number={4},
  pages={537},
  year={2024},
  publisher={MDPI}
}

@article{chalapathy2019deep,
  title={Deep learning for anomaly detection: A survey},
  author={Chalapathy, Raghavendra and Chawla, Sanjay},
  journal={arXiv preprint arXiv:1901.03407},
  year={2019}
}

@article{benmeziane2021comprehensive,
  title={A comprehensive survey on hardware-aware neural architecture search},
  author={Benmeziane, Hadjer and Maghraoui, Kaoutar El and Wistuba, Martin and Wang, Naigang},
  journal={arXiv preprint arXiv:2101.09336},
  year={2021}
}

@article{oh2022review,
  title={A review of deep learning applications for railway safety},
  author={Oh, Kyuetaek and Yoo, Mintaek and Jin, Nayoung and Ko, Jisu and Seo, Jeonguk and Joo, Hyojin and Ko, Minsam},
  journal={Applied Sciences},
  volume={12},
  number={20},
  pages={10572},
  year={2022},
  publisher={MDPI}
}

@article{tang2022literature,
  title={A literature review of Artificial Intelligence applications in railway systems},
  author={Tang, Ruifan and De Donato, Lorenzo and Besinovi{\'c}, Nikola and Flammini, Francesco and Goverde, Rob MP and Lin, Zhiyuan and Liu, Ronghui and Tang, Tianli and Vittorini, Valeria and Wang, Ziyulong},
  journal={Transportation Research Part C: Emerging Technologies},
  volume={140},
  pages={103679},
  year={2022},
  publisher={Elsevier}
}

@article{kumar2024systematic,
  title={A systematic literature review of defect detection in railways using machine vision-based inspection methods},
  author={Kumar, Ankit and Harsha, SP},
  journal={International Journal of Transportation Science and Technology},
  year={2024},
  publisher={Elsevier}
}

@article{baldyga2024anomaly,
  title={Anomaly detection in railway sensor data environments: state-of-the-art methods and empirical performance evaluation},
  author={Ba{\l}dyga, Micha{\l} and Bara{\'n}ski, Kacper and Belter, Jakub and Kalinowski, Mateusz and Weichbroth, Pawe{\l}},
  journal={Sensors},
  volume={24},
  number={8},
  pages={2633},
  year={2024},
  publisher={MDPI}
}

@article{lourencco2024time,
  title={Time series data mining for railway wheel and track monitoring: a survey},
  author={Louren{\c{c}}o, Afonso and Ribeiro, Diogo and Fernandes, Marta and Marreiros, Goreti},
  journal={Neural Computing and Applications},
  volume={36},
  number={27},
  pages={16707--16725},
  year={2024},
  publisher={Springer}
}

@article{de2022survey,
  title={A survey on audio-video based defect detection through deep learning in railway maintenance},
  author={De Donato, Lorenzo and Flammini, Francesco and Marrone, Stefano and Mazzariello, Claudio and Nardone, Roberto and Sansone, Carlo and Vittorini, Valeria},
  journal={IEEE Access},
  volume={10},
  pages={65376--65400},
  year={2022},
  publisher={IEEE}
}

@article{di2023review,
  title={A review on deep learning techniques for railway infrastructure monitoring},
  author={Di Summa, Maria and Griseta, Maria Elena and Mosca, Nicola and Patruno, Cosimo and Nitti, Massimiliano and Ren{\`o}, Vito and Stella, Ettore},
  journal={IEEE Access},
  volume={11},
  pages={114638--114661},
  year={2023},
  publisher={IEEE}
}

@article{binder2023predictive,
  title={Predictive maintenance for railway domain: A systematic literature review},
  author={Binder, Mario and Mezhuyev, Vitaliy and Tschandl, Martin},
  journal={IEEE Engineering Management Review},
  volume={51},
  number={2},
  pages={120--140},
  year={2023},
  publisher={IEEE}
}

@article{chawla2002smote,
  title={SMOTE: synthetic minority over-sampling technique},
  author={Chawla, Nitesh V and Bowyer, Kevin W and Hall, Lawrence O and Kegelmeyer, W Philip},
  journal={Journal of artificial intelligence research},
  volume={16},
  pages={321--357},
  year={2002}
}

@inproceedings{he2008adasyn,
  title={ADASYN: Adaptive synthetic sampling approach for imbalanced learning},
  author={He, Haibo and Bai, Yang and Garcia, Edwardo A and Li, Shutao},
  booktitle={2008 IEEE international joint conference on neural networks (IEEE world congress on computational intelligence)},
  pages={1322--1328},
  year={2008},
  organization={Ieee}
}

@article{liu2023anomaly,
  title={Anomaly-GAN: A data augmentation method for train surface anomaly detection},
  author={Liu, Ruikang and Liu, Weiming and Zheng, Zhongxing and Wang, Liang and Mao, Liang and Qiu, Qisheng and Ling, Guangzheng},
  journal={Expert Systems with Applications},
  volume={228},
  pages={120284},
  year={2023},
  publisher={Elsevier}
}

@inproceedings{su2020adapting,
  title={Adapting object detectors with conditional domain normalization},
  author={Su, Peng and Wang, Kun and Zeng, Xingyu and Tang, Shixiang and Chen, Dapeng and Qiu, Di and Wang, Xiaogang},
  booktitle={Computer Vision--ECCV 2020: 16th European Conference, Glasgow, UK, August 23--28, 2020, Proceedings, Part XI 16},
  pages={403--419},
  year={2020},
  organization={Springer}
}

@article{shang2021vibration,
  title={Vibration-based damage detection for bridges by deep convolutional denoising autoencoder},
  author={Shang, Zhiqiang and Sun, Limin and Xia, Ye and Zhang, Wei},
  journal={Structural Health Monitoring},
  volume={20},
  number={4},
  pages={1880--1903},
  year={2021},
  publisher={SAGE Publications Sage UK: London, England}
}

@misc{openrail2025,
  author       = {OpenRailLab},
  title        = {OpenRailLab: Simulation Platform for Railway Systems},
  year         = {2025},
  url          = {https://www.openraillab.org},
  note         = {Accessed: 2025-07-03}
}

@phdthesis{huang2025ai,
  title={AI-Enabled Knowledge Transfer and Learning for Nondestructive Evaluation Toward Intelligent and Adaptive Systems},
  author={Huang, Xuhui},
  year={2025},
  school={Michigan State University}
}

@article{galvez2021fault,
  title={Fault detection and RUL estimation for railway HVAC systems using a hybrid model-based approach},
  author={G{\'a}lvez, Antonio and Diez-Olivan, Alberto and Seneviratne, Dammika and Galar, Diego},
  journal={Sustainability},
  volume={13},
  number={12},
  pages={6828},
  year={2021},
  publisher={MDPI}
}

@article{1,
  title={A comparative study of fault diagnosis for train door system: traditional versus deep learning approaches},
  author={Ham, Seokju and Han, Seok-Youn and Kim, Seokgoo and Park, Hyung Jun and Park, Kee-Jun and Choi, Joo-Ho},
  journal={Sensors},
  volume={19},
  number={23},
  pages={5160},
  year={2019},
  publisher={MDPI}
}

@article{cao2024railway,
  title={Railway intrusion detection based on machine vision: A survey, challenges, and perspectives},
  author={Cao, Zhiwei and Qin, Yong and Jia, Limin and Xie, Zhengyu and Gao, Yang and Wang, Yaguan and Li, Ping and Yu, Zujun},
  journal={IEEE Transactions on Intelligent Transportation Systems},
  volume={25},
  number={7},
  pages={6427--6448},
  year={2024},
  publisher={IEEE}
}

@article{bris2026survey,
  title={A Survey of AI-Enabled Predictive Maintenance for Railway Infrastructure: Models, Data Sources, and Research Challenges},
  author={Bris-Pe{\~n}alver, Francisco Javier and Verdecia-Pe{\~n}a, Randy and Alonso, Jos{\'e} I},
  journal={Sensors},
  volume={26},
  number={3},
  pages={906},
  year={2026},
  publisher={MDPI}
}

@article{gao2023railway,
  title={Railway fastener anomaly detection via multisensor fusion and self-driven loss reweighting},
  author={Gao, Yang and Cao, Zhiwei and Qin, Yong and Ge, Xuanyu and Lian, Lirong and Bai, Jie and Yu, Hang},
  journal={IEEE Sensors Journal},
  volume={24},
  number={2},
  pages={1812--1825},
  year={2023},
  publisher={IEEE}
}

@article{liu2020real,
  title={Real-time hierarchical neural network based fault detection and isolation for high-speed railway system under hybrid AC/DC grid},
  author={Liu, Qin and Liang, Tian and Dinavahi, Venkata},
  journal={IEEE Transactions on Power Delivery},
  volume={35},
  number={6},
  pages={2853--2864},
  year={2020},
  publisher={IEEE}
}

@article{sresakoolchai2022railway,
  title={Railway defect detection based on track geometry using supervised and unsupervised machine learning},
  author={Sresakoolchai, Jessada and Kaewunruen, Sakdirat},
  journal={Structural health monitoring},
  volume={21},
  number={4},
  pages={1757--1767},
  year={2022},
  publisher={SAGE Publications Sage UK: London, England}
}

@article{ham2019comparative,
  title={A comparative study of fault diagnosis for train door system: traditional versus deep learning approaches},
  author={Ham, Seokju and Han, Seok-Youn and Kim, Seokgoo and Park, Hyung Jun and Park, Kee-Jun and Choi, Joo-Ho},
  journal={Sensors},
  volume={19},
  number={23},
  pages={5160},
  year={2019},
  publisher={MDPI}
}

@article{siddiqui2022iot,
  title={IoT based railway track faults detection and localization using acoustic analysis},
  author={Siddiqui, Hafeez Ur Rehman and Saleem, Adil Ali and Raza, Muhammad Amjad and Zafar, Kainat and Munir, Kashif and Dudley, Sandra},
  journal={IEEE Access},
  volume={10},
  pages={106520--106533},
  year={2022},
  publisher={IEEE}
}

@article{sun2020fault,
  title={A fault diagnosis method for train plug doors via sound signals},
  author={Sun, Yongkui and Cao, Yuan and Ma, Lianchuan},
  journal={IEEE Intelligent Transportation Systems Magazine},
  volume={13},
  number={3},
  pages={107--117},
  year={2020},
  publisher={IEEE}
}

@article{tsiktsiris2024multimodal,
  title={Multimodal abnormal event detection in public transportation},
  author={Tsiktsiris, Dimitris and Lalas, Antonios and Dasygenis, Minas and Votis, Konstantinos},
  journal={IEEE Access},
  year={2024},
  publisher={IEEE}
}

@article{kaya2025detection,
  author={Kaya, Utku},

  journal={IEEE Access}, 

  title={Detection of Spark Erosion on Insulated Rail Joints by Deep Learning}, 

  year={2025},

  volume={13},

  number={},

  pages={79291-79303},

  doi={10.1109/ACCESS.2025.3567193}
}

@article{ning2024research,
  title={Research on the method of foreign object detection for railway tracks based on deep learning},
  author={Ning, Shanping and Ding, Feng and Chen, Bangbang},
  journal={Sensors},
  volume={24},
  number={14},
  pages={4483},
  year={2024},
  publisher={MDPI}
}

@article{wang2021efficient,
  title={An efficient anomaly detection for high-speed train braking system using broad learning system},
  author={Wang, Chong and Liu, Jie},
  journal={IEEE Access},
  volume={9},
  pages={63825--63832},
  year={2021},
  publisher={IEEE}
}

@inproceedings{yilmazer2022mask,
  title={Mask R-CNN architecture based railway fastener fault detection approach},
  author={Yilmazer, Merve and Karakose, Mehmet},
  booktitle={2022 International Conference on Decision Aid Sciences and Applications (DASA)},
  pages={1363--1366},
  year={2022},
  organization={IEEE}
}

@inproceedings{rakshit2022railway,
  title={Railway track fault detection using deep neural networks},
  author={Rakshit, Soumik and Sandeep, BS and others},
  booktitle={2022 IEEE 6th Conference on Information and Communication Technology (CICT)},
  pages={1--5},
  year={2022},
  organization={IEEE}
}

@article{xiong2023detection,
  title={Detection of rail defects using NDT methods},
  author={Xiong, Longhui and Jing, Guoqing and Wang, Jingru and Liu, Xiubo and Zhang, Yuhua},
  journal={Sensors},
  volume={23},
  number={10},
  pages={4627},
  year={2023},
  publisher={MDPI}
}

@article{shim2023methodology,
  title={A Methodology of Condition Monitoring System Utilizing Supervised and Semi-Supervised Learning in Railway},
  author={Shim, Jaeseok and Koo, Jeongseo and Park, Yongwoon},
  journal={Sensors},
  volume={23},
  number={22},
  pages={9075},
  year={2023},
  publisher={MDPI}
}

@article{wang2022improved,
  title={Improved LSTM-based time-series anomaly detection in rail transit operation environments},
  author={Wang, Yujie and Du, Xin and Lu, Zhihui and Duan, Qiang and Wu, Jie},
  journal={IEEE Transactions on Industrial Informatics},
  volume={18},
  number={12},
  pages={9027--9036},
  year={2022},
  publisher={IEEE}
}

@article{galvez2022hybrid,
  title={A hybrid model-based approach on prognostics for railway HVAC},
  author={G{\'a}lvez, Antonio and Galar, Diego and Seneviratne, Dammika},
  journal={IEEE Access},
  volume={10},
  pages={108117--108127},
  year={2022},
  publisher={IEEE}
}

@article{liu2021short,
  title={Short-term strong wind risk prediction for high-speed railway},
  author={Liu, Haoyu and Liu, Chen and He, Shibo and Chen, Jiming},
  journal={IEEE Transactions on Intelligent Transportation Systems},
  volume={22},
  number={7},
  pages={4243--4255},
  year={2021},
  publisher={IEEE}
}

@article{lee2021real,
  title={Real-time hybrid deep learning-based train running safety prediction framework of railway vehicle},
  author={Lee, Hyunsoo and Han, Seok-Youn and Park, Keejun and Lee, Hoyoung and Kwon, Taesoo},
  journal={Machines},
  volume={9},
  number={7},
  pages={130},
  year={2021},
  publisher={MDPI}
}

@article{hao2023track,
  title={Track geometry estimation from vehicle--body acceleration for high-speed railway using deep learning technique},
  author={Hao, Xiaoli and Yang, Jian and Yang, Fei and Sun, Xianfu and Hou, Yali and Wang, Jian},
  journal={Vehicle system dynamics},
  volume={61},
  number={1},
  pages={239--259},
  year={2023},
  publisher={Taylor \& Francis}
}

@inproceedings{davari2021predictive,
  title={Predictive maintenance based on anomaly detection using deep learning for air production unit in the railway industry},
  author={Davari, Narjes and Veloso, Bruno and Ribeiro, Rita P and Pereira, Pedro Mota and Gama, Jo{\~a}o},
  booktitle={2021 IEEE 8th International Conference on Data Science and Advanced Analytics (DSAA)},
  pages={1--10},
  year={2021},
  organization={IEEE}
}

@article{wang2025spatial,
  title={A spatial-temporal neural network based on ResNet-Transformer for predicting railroad broken rails},
  author={Wang, Xin and Dai, Junyan and Liu, Xiang},
  journal={Advanced Engineering Informatics},
  volume={65},
  pages={103126},
  year={2025},
  publisher={Elsevier}
}

@article{bernal2018onboard,
  title={Onboard condition monitoring sensors, systems and techniques for freight railway vehicles: a review},
  author={Bernal, Esteban and Spiryagin, Maksym and Cole, Colin},
  journal={IEEE sensors journal},
  volume={19},
  number={1},
  pages={4--24},
  year={2018},
  publisher={IEEE}
}

@article{castillo2020critical,
  title={A critical review of sensors for the continuous monitoring of smart and sustainable railway infrastructures},
  author={Castillo-Mingorance, Juan Manuel and Sol-S{\'a}nchez, Miguel and Moreno-Navarro, Fernando and Rubio-G{\'a}mez, Mar{\'\i}a Carmen},
  journal={Sustainability},
  volume={12},
  number={22},
  pages={9428},
  year={2020},
  publisher={Mdpi}
}

@article{xu2020train,
  title={Train--track coupled dynamics analysis: system spatial variation on geometry, physics and mechanics},
  author={Xu, Lei and Zhai, Wanming},
  journal={Railway Engineering Science},
  volume={28},
  number={1},
  pages={36--53},
  year={2020},
  publisher={Springer}
}

@article{shah2025recent,
  title={Recent Advances in Vibration Analysis for Predictive Maintenance of Modern Automotive Powertrains},
  author={Shah, Rajesh and Mittal, Vikram and Lotwin, Michael},
  journal={Vibration},
  volume={8},
  number={4},
  pages={68},
  year={2025},
  publisher={MDPI}
}

@article{niu2020unsupervised,
  title={Unsupervised saliency detection of rail surface defects using stereoscopic images},
  author={Niu, Menghui and Song, Kechen and Huang, Liming and Wang, Qi and Yan, Yunhui and Meng, Qinggang},
  journal={IEEE Transactions on Industrial Informatics},
  volume={17},
  number={3},
  pages={2271--2281},
  year={2020},
  publisher={IEEE}
}

@article{zhang2021rail5,
  title={Rail-5k: A real-world dataset for rail surface defects detection},
  author={Zhang, Zihao and Yu, Shaozuo and Yang, Siwei and Zhou, Yu and Zhao, Bingchen},
  journal={arXiv preprint arXiv:2106.14366},
  year={2021}
}

@article{sun2021survey,
  title={A survey on deep learning for data-driven soft sensors},
  author={Sun, Qingqiang and Ge, Zhiqiang},
  journal={IEEE Transactions on Industrial Informatics},
  volume={17},
  number={9},
  pages={5853--5866},
  year={2021},
  publisher={IEEE}
}

@article{wang2020deep,
  title={Deep learning for spatio-temporal data mining: A survey},
  author={Wang, Senzhang and Cao, Jiannong and Philip, S Yu},
  journal={IEEE transactions on knowledge and data engineering},
  volume={34},
  number={8},
  pages={3681--3700},
  year={2020},
  publisher={IEEE}
}

@article{he2025defect,
  title={Defect detection in railway catenary systems: a survey},
  author={He, Jin and He, Jinming and Gou, Zhijian and Lv, Fengmao},
  journal={Intelligent Transportation Infrastructure},
  pages={liaf027},
  year={2025},
  publisher={Oxford University Press}
}

@article{komorski2021advanced,
  title={Advanced acoustic signal analysis used for wheel-flat detection},
  author={Komorski, Pawel and Szymanski, Grzegorz M and Nowakowski, Tomasz and Orczyk, Malgorzata},
  journal={Latin American Journal of Solids and Structures},
  volume={18},
  number={01},
  pages={e338},
  year={2021},
  publisher={SciELO Brasil}
}

@article{page2021prisma,
  title={The PRISMA 2020 statement: an updated guideline for reporting systematic reviews},
  author={Page, Matthew J and McKenzie, Joanne E and Bossuyt, Patrick M and Boutron, Isabelle and Hoffmann, Tammy C and Mulrow, Cynthia D and Shamseer, Larissa and Tetzlaff, Jennifer M and Akl, Elie A and Brennan, Sue E and others},
  journal={bmj},
  volume={372},
  year={2021},
  publisher={British Medical Journal Publishing Group}
}

@article{chen2025large,
  title={Large Models for Machine Monitoring and Fault Diagnostics: Opportunities, Challenges, and Future Direction},
  author={Chen, Xuefeng and Lei, Yaguo and Li, Yanfu and Parkinson, Simon and Li, Xiang and Liu, Jinxin and Lu, Fan and Wang, Huan and Wang, Zisheng and Yang, Bin and others},
  journal={Journal of Dynamics, Monitoring and Diagnostics},
  volume={4},
  number={2},
  pages={76--90},
  year={2025}
}

@article{arain2024railway,
  title={Railway track surface faults dataset},
  author={Arain, Asfar and Mehran, Sanaullah and Shaikh, Muhammad Zakir and Kumar, Dileep and Chowdhry, Bhawani Shankar and Hussain, Tanweer},
  journal={Data in brief},
  volume={52},
  pages={110050},
  year={2024},
  publisher={Elsevier}
}

@article{shimizu2022real,
  title={A real-time fault detection framework based on unsupervised deep learning for prognostics and health management of railway assets},
  author={Shimizu, Minoru and Perinpanayagam, Suresh and Namoano, Bernadin},
  journal={IEEE Access},
  volume={10},
  pages={96442--96458},
  year={2022},
  publisher={IEEE}
}

@article{wang2022deep,
  title={A deep generative approach for rail foreign object detections via semisupervised learning},
  author={Wang, Tiange and Zhang, Zijun and Tsui, Kwok-Leung},
  journal={IEEE Transactions on Industrial Informatics},
  volume={19},
  number={1},
  pages={459--468},
  year={2022},
  publisher={IEEE}
}

@inproceedings{zendel2019railsem19,
  title={{RailSem19}: A Dataset for Semantic 
         Rail Scene Understanding},
  author={Zendel, Oliver and Murschitz, Markus 
          and Zeilinger, Marcel and Steininger, 
          Daniel and Abbasi, Sara and 
          Beleznai, Csaba},
  booktitle={Proceedings of the IEEE/CVF 
             Conference on Computer Vision 
             and Pattern Recognition Workshops},
  year={2019}
}

@inproceedings{tagiew2023osdar23,
  title={{OSDaR23}: Open Sensor Data for 
         Rail 2023},
  author={Tagiew, Rustam and K{\"o}ppel, Martin 
          and Schwalbe, Karsten and others},
  booktitle={International Conference on 
             Robotics and Automation Engineering},
  pages={270--276},
  year={2023}
}

@article{veloso2022metropt,
  title={The {MetroPT} dataset for predictive 
         maintenance},
  author={Veloso, Bráulio and Gama, João and 
          Ribeiro, Rita P. and Pereira, Pedro M.},
  journal={Scientific Data},
  volume={9},
  number={1},
  pages={764},
  year={2022},
  doi={10.1038/s41597-022-01877-3}
}

@article{pappaterra2021systematic,
  title={A Systematic Review of Artificial 
         Intelligence Public Datasets for 
         Railway Applications},
  author={Pappaterra, Mauro Jos{\'e} and 
          Flammini, Francesco and 
          Vittorini, Valeria and 
          Be{\v{s}}inovi{\'c}, Nikola},
  journal={Infrastructures},
  volume={6},
  number={10},
  pages={136},
  year={2021},
  doi={10.3390/infrastructures6100136}
}

@article{signorino2020dataset,
  title={Dataset of measured and commented 
         pantograph electric arcs in {DC} railways},
  author={Signorino, D. and Giordano, D. and 
          Mariscotti, A. and others},
  journal={Data in Brief},
  volume={31},
  pages={105978},
  year={2020},
  doi={10.1016/j.dib.2020.105978}
}

@article{mariscotti2020datasets,
  title={Data sets of measured pantograph 
         voltage and current of {European} 
         {AC} railways},
  author={Mariscotti, Andrea},
  journal={Data in Brief},
  volume={30},
  pages={105477},
  year={2020},
  doi={10.1016/j.dib.2020.105477}
}

@article{liu2018drtrain,
  title={The {DR-Train} dataset: dynamic responses, 
         {GPS} positions and environmental 
         conditions of two light rail vehicles 
         in {Pittsburgh}},
  author={Liu, Jingxiao and Chen, Shijia and 
          Lederman, Gil and others},
  journal={Scientific Data},
  volume={6},
  pages={146},
  year={2019},
  doi={10.1038/s41597-019-0148-8}
}

@article{khan2022transformers,
  title={Transformers in vision: A survey},
  author={Khan, Salman and Naseer, Muzammal and Hayat, Munawar and Zamir, Syed Waqas and Khan, Fahad Shahbaz and Shah, Mubarak},
  journal={ACM computing surveys (CSUR)},
  volume={54},
  number={10s},
  pages={1--41},
  year={2022},
  publisher={ACM New York, NY}
}

@incollection{koonce2021resnet,
  title={ResNet 50},
  author={Koonce, Brett},
  booktitle={Convolutional neural networks with swift for tensorflow: image recognition and dataset categorization},
  pages={63--72},
  year={2021},
  publisher={Springer}
}

@article{tammina2019transfer,
  title={Transfer learning using vgg-16 with deep convolutional neural network for classifying images},
  author={Tammina, Srikanth and others},
  journal={International Journal of Scientific and Research Publications (IJSRP)},
  volume={9},
  number={10},
  pages={143--150},
  year={2019}
}

@article{shafique2021acoustic,
  title={A Novel Approach to Railway Track 
         Faults Detection Using Acoustic Analysis},
  author={Shafique, Raheel and Siddiqui, 
          Hafeez Ur Rehman and Rustam, Furqan 
          and Ullah, Saleem and Siddique, 
          Muhammad Abubakar and Lee, Eunmok 
          and Ashraf, Imran and Dudley, Sandra},
  journal={Sensors},
  volume={21},
  number={18},
  pages={6221},
  year={2021},
  doi={10.3390/s21186221}
}

@article{shafique2023mfcc,
  title={Improved railway track faults detection 
         using {Mel}-frequency cepstral coefficient 
         and constant-{Q} transform features},
  author={Shafique, Raheel and Kanwal, Khadija 
          and Choi, Gyu Sang},
  journal={Scientific Reports},
  volume={15},
  year={2025},
  doi={10.1038/s41598-025-14763-w}
}

@article{bouguettaya2024ensemble,
  title={A reusable {AI}-enabled defect detection 
         system for railway using ensembled {CNN}},
  author={Bouguettaya, Abdelouahab and 
          Zarzour, Hafed and Kechida, Ahmed 
          and Taberkit, Abdelmadjid Mohammed},
  journal={Applied Intelligence},
  year={2024},
  doi={10.1007/s10489-024-05676-8}
}

@article{shafique2022wheelset,
  title={Deep learning-based fault detection 
         in railway wheelsets using time 
         series analysis},
  author={Shafique, Raheel and others},
  journal={Mehran University Research Journal 
          of Engineering and Technology},
  volume={42},
  number={3},
  year={2023}
}

@article{shaikh2025contribution,
  title={A contribution to reliable rail transport: AI-powered real-time wheel defect detection},
  author={Shaikh, Muhammad Zakir and Baro, Enrique Nava and Jatoi, Sahil},
  journal={Scientific Reports},
  year={2025},
  publisher={Nature Publishing Group UK London}
}

@article{chen2024lstm,
  title={A {Deep-LSTM}-Based Fault Detection 
         Method for Railway Vehicle Suspensions},
  author={Chen, Yuejian and Liu, Xuemei 
          and Fan, Wenkun and Duan, Ningyuan 
          and Zhou, Kai},
  journal={Machines},
  volume={12},
  number={2},
  pages={116},
  year={2024},
  doi={10.3390/machines12020116}
}

@article{autores2025yolo11,
  title={Automatic Detection of Railway Faults 
         Using Neural Networks: A Comparative 
         Study of Transfer Learning Models 
         and {YOLOv11}},
  author={Rodríguez-Alvarado, Omar and 
          others},
  journal={Infrastructures},
  volume={10},
  number={1},
  pages={3},
  year={2024},
  doi={10.3390/infrastructures10010003}
}

@article{brintha2024fodyolo,
  title={{FOD-YOLO} Net: Fasteners fault and 
         object detection in railway tracks 
         using deep {YOLO} network},
  author={Brintha, K. and Jawhar, S. Joseph},
  journal={Journal of Intelligent and 
          Fuzzy Systems},
  volume={46},
  year={2024},
  doi={10.3233/JIFS-236445}
}

@article{zhou2025rfddetr,
  title={Railway fastener defect detection 
         using {RFD-DETR}: A lightweight 
         real-time transformer-based approach},
  author={Zhou, Huixiang and Liu, Yuhao 
          and Wang, Jian and Fang, Aiqing 
          and Mohammadzadeh, Ardashir},
  journal={PLOS ONE},
  year={2025},
  doi={10.1371/journal.pone.0331513}
}

@inproceedings{padn2024pantograph,
  title={Pantograph unsupervised anomaly detection network based on reverse distillation},
  author={Yan, Hao and Lin, Chuan and Xu, Zhiyuan and Guo, Ningning and Qing, Anyong},
  booktitle={Proceedings of the 2024 8th International Conference on Advances in Image Processing (ICAIP)},
  pages={55--60},
  year={2024}
}

@article{railfdnet2025,
  title={RailFDNet: A hybrid supervision and feature discrepancy enhancement model for railway anomalous object detection},
  author={Sun, Tao and Guo, Baoqing and Ruan, Tao and Zhou, Xingfang and Bai, Dingyuan and Yu, Hang and Wang, Yu},
  journal={Expert Systems with Applications},
  volume={275},
  pages={127005},
  year={2025},
  publisher={Elsevier}
}

@article{edgeai2025survey,
  title={Edge AI in Practice: A Survey and Deployment Framework for Neural Networks on Embedded Systems},
  author={Cordova-Cardenas, Ruth and Amor, Daniel and Guti{\'e}rrez, {\'A}lvaro},
  journal={Electronics},
  volume={14},
  number={24},
  pages={4877},
  year={2025},
  publisher={MDPI}
}

@article{railpatchcore2025,
  title={Rail-PatchCore: unsupervised learning-based detection of visual anomalies in the railway-turnout environment: Y. Zhang et al.},
  author={Zhang, YuanHao and Yu, Zujun and Zhu, Liqiang and Guo, Baoqing and Wang, Yao},
  journal={Applied Intelligence},
  volume={55},
  number={6},
  pages={408},
  year={2025},
  publisher={Springer}
}

@article{jiang2022review,
  title={A Review of Yolo algorithm developments},
  author={Jiang, Peiyuan and Ergu, Daji and Liu, Fangyao and Cai, Ying and Ma, Bo},
  journal={Procedia computer science},
  volume={199},
  pages={1066--1073},
  year={2022},
  publisher={Elsevier}
}

@article{shaikh2025faultseg,
  title={{FaultSeg}: A dataset for train wheel 
         defect detection},
  author={Shaikh, Muhammad Zakir and Jatoi, Sahil 
          and Baro, Enrique Nava and Das, Bhagwan 
          and Hussain, Samreen and 
          Chowdhry, Bhawani Shankar},
  journal={Scientific Data},
  volume={12},
  number={1},
  pages={309},
  year={2025},
  doi={10.1038/s41597-025-04557-0}
}

@inproceedings{laurendin2021hazardous,
  title={Hazardous events detection in automatic train doors vicinity using deep neural networks},
  author={Laurendin, Olivier and Ambellouis, S{\'e}bastien and Fleury, Anthony and Mahtani, Ankur and Chafik, Sanaa and Strauss, Cl{\'e}ment},
  booktitle={2021 17th IEEE International Conference on Advanced Video and Signal Based Surveillance (AVSS)},
  pages={1--7},
  year={2021},
  organization={IEEE}
}

@article{yuan2021unsupervised,
  title={An unsupervised method based on convolutional variational auto-encoder and anomaly detection algorithms for light rail squat localization},
  author={Yuan, Zhandong and Zhu, Shengyang and Chang, Chao and Yuan, Xuancheng and Zhang, Qinglai and Zhai, Wanming},
  journal={Construction and Building Materials},
  volume={313},
  pages={125563},
  year={2021},
  publisher={Elsevier}
}

@misc{UIC2025,
  author    = "{International Union of Railways (UIC)}",
  title     = "{Railway Statistics Synopsis 2025 Edition}",
  year      = "2025",
  note      = "Available: https://uic-stats.uic.org",
  howpublished = "\url{https://uic-stats.uic.org}"
}

@article{javed2020anomaly,
  title={Anomaly detection in automated vehicles using multistage attention-based convolutional neural network},
  author={Javed, Abdul Rehman and Usman, Muhammad and Rehman, Saif Ur and Khan, Mohib Ullah and Haghighi, Mohammad Sayad},
  journal={IEEE Transactions on Intelligent Transportation Systems},
  volume={22},
  number={7},
  pages={4291--4300},
  year={2020},
  publisher={IEEE}
}

@article{baltruvsaitis2018multimodal,
  title={Multimodal machine learning: A survey and taxonomy},
  author={Baltru{\v{s}}aitis, Tadas and Ahuja, Chaitanya and Morency, Louis-Philippe},
  journal={IEEE transactions on pattern analysis and machine intelligence},
  volume={41},
  number={2},
  pages={423--443},
  year={2018},
  publisher={IEEE}
}

@article{nateghinia2025development,
  title={Development of an unsupervised 3D LiDAR-based methodology for automated safety monitoring of railway facilities},
  author={Nateghinia, Ehsan and Miranda-Moreno, Luis F},
  journal={International Journal of Transportation Science and Technology},
  year={2025},
  publisher={Elsevier}
}

@article{ferrari2018detecting,
  title={Detecting requirements defects with NLP patterns: an industrial experience in the railway domain},
  author={Ferrari, Alessio and Gori, Gloria and Rosadini, Benedetta and Trotta, Iacopo and Bacherini, Stefano and Fantechi, Alessandro and Gnesi, Stefania},
  journal={Empirical Software Engineering},
  volume={23},
  number={6},
  pages={3684--3733},
  year={2018},
  publisher={Springer}
}

@article{cui2019real,
  title={Real-time inspection system for ballast railway fasteners based on point cloud deep learning},
  author={Cui, Hao and Li, Jian and Hu, Qingwu and Mao, Qingzhou},
  journal={IEEE Access},
  volume={8},
  pages={61604--61614},
  year={2019},
  publisher={IEEE}
}

@article{gao2020automatic,
  title={Automatic detection and monitoring system of pantograph--catenary in China’s high-speed railways},
  author={Gao, Shibin},
  journal={IEEE Transactions on Instrumentation and Measurement},
  volume={70},
  pages={1--12},
  year={2020},
  publisher={IEEE}
}

@article{chandola2009anomaly,
  title={Anomaly detection: A survey},
  author={Chandola, Varun and Banerjee, Arindam and Kumar, Vipin},
  journal={ACM computing surveys (CSUR)},
  volume={41},
  number={3},
  pages={1--58},
  year={2009},
  publisher={ACM New York, NY, USA}
}

@inproceedings{malhotra2015long,
  title={Long short-term memory networks for anomaly detection in time series},
  author={Malhotra, Pankaj and Vig, Lovekesh and Shroff, Gautam and Agarwal, Puneet and others},
  booktitle={Proceedings},
  volume={89},
  number={9},
  pages={94},
  year={2015}
}

@inproceedings{gasparini2021anomaly,
  title={Anomaly detection, localization and classification for railway inspection},
  author={Gasparini, Riccardo and D'Eusanio, Andrea and Borghi, Guido and Pini, Stefano and Scaglione, Giuseppe and Calderara, Simone and Fedeli, Eugenio and Cucchiara, Rita},
  booktitle={2020 25th International Conference on Pattern Recognition (ICPR)},
  pages={3419--3426},
  year={2021},
  organization={IEEE}
}

@article{iqbal2024anomaly,
  title={Anomaly detection in multivariate time series data using deep ensemble models},
  author={Iqbal, Amjad and Amin, Rashid and Alsubaei, Faisal S and Alzahrani, Abdulrahman},
  journal={Plos one},
  volume={19},
  number={6},
  pages={e0303890},
  year={2024},
  publisher={Public Library of Science San Francisco, CA USA}
}

@article{di2019survey,
  title={A survey on gans for anomaly detection},
  author={Di Mattia, Federico and Galeone, Paolo and De Simoni, Michele and Ghelfi, Emanuele},
  journal={arXiv preprint arXiv:1906.11632},
  year={2019}
}

@article{vskvara2023auc,
  title={Is AUC the best measure for practical comparison of anomaly detectors?},
  author={{\v{S}}kv{\'a}ra, V{\'\i}t and Pevn{\`y}, Tom{\'a}{\v{s}} and {\v{S}}m{\'\i}dl, V{\'a}clav},
  journal={arXiv preprint arXiv:2305.04754},
  year={2023}
}

@article{berahmand2024autoencoders,
  title={Autoencoders and their applications in machine learning: a survey},
  author={Berahmand, Kamal and Daneshfar, Fatemeh and Salehi, Elaheh Sadat and Li, Yuefeng and Xu, Yue},
  journal={Artificial intelligence review},
  volume={57},
  number={2},
  pages={28},
  year={2024},
  publisher={Springer}
}

@article{choi2024deep,
  title={Deep learning (Fast R-CNN)-based evaluation of rail surface defects},
  author={Choi, Jung-Youl and Han, Jae-Min},
  journal={Applied Sciences},
  volume={14},
  number={5},
  pages={1874},
  year={2024},
  publisher={MDPI}
}

@article{pang2021deep,
  title={Deep learning for anomaly detection: A review},
  author={Pang, Guansong and Shen, Chunhua and Cao, Longbing and Hengel, Anton Van Den},
  journal={ACM computing surveys (CSUR)},
  volume={54},
  number={2},
  pages={1--38},
  year={2021},
  publisher={ACM New York, NY, USA}
}

@article{cai2018proxylessnas,
  title={Proxylessnas: Direct neural architecture search on target task and hardware},
  author={Cai, Han and Zhu, Ligeng and Han, Song},
  journal={arXiv preprint arXiv:1812.00332},
  year={2018}
}

\end{document}